\documentclass[smallextended]{svjour3}       

\usepackage[table,xcdraw]{xcolor}

\usepackage{appendix}
\usepackage{amsmath}
\usepackage{graphicx}
\usepackage{lineno}
\usepackage{array}
\usepackage{longtable}

\usepackage{multirow}
\usepackage{algorithm}
\usepackage[noend]{algpseudocode}

\usepackage{tikz,hyperref}

\definecolor{lime}{HTML}{A6CE39}
\DeclareRobustCommand{\orcidicon}{
	\begin{tikzpicture}
	\draw[lime, fill=lime] (0,0) 
	circle [radius=0.16] 
	node[white] {{\fontfamily{qag}\selectfont \tiny ID}};
	\draw[white, fill=white] (-0.0625,0.095) 
	circle [radius=0.007];
	\end{tikzpicture}
	\hspace{-2mm}
}
\foreach \x in {A, ..., Z}{\expandafter\xdef\csname orcid\x\endcsname{\noexpand\href{https://orcid.org/\csname orcidauthor\x\endcsname}
			{\noexpand\orcidicon}}
}

\newcommand*\patchAmsMathEnvironmentForLineno[1]{%
\expandafter\let\csname old#1\expandafter\endcsname\csname #1\endcsname
\expandafter\let\csname oldend#1\expandafter\endcsname\csname end#1\endcsname
\renewenvironment{#1}%
{\linenomath\csname old#1\endcsname}%
{\csname oldend#1\endcsname\endlinenomath}}%
\newcommand*\patchBothAmsMathEnvironmentsForLineno[1]{%
\patchAmsMathEnvironmentForLineno{#1}%
\patchAmsMathEnvironmentForLineno{#1*}}%
\AtBeginDocument{%
\patchBothAmsMathEnvironmentsForLineno{equation}%
\patchBothAmsMathEnvironmentsForLineno{align}%
\patchBothAmsMathEnvironmentsForLineno{flalign}%
\patchBothAmsMathEnvironmentsForLineno{alignat}%
\patchBothAmsMathEnvironmentsForLineno{gather}%
\patchBothAmsMathEnvironmentsForLineno{multline}%
}

\graphicspath{{figures/}}

\begin{document}

\title{Combining Lipschitz and RBF Surrogate Models for High-dimensional Computationally Expensive Problems
}

\titlerunning{Combining Lipschitz and RBF Surrogate Models for High-dimensional  ... }        

\author{Jakub K\r{u}dela\orcidA{}, Radomil Matoušek\orcidB{}
}


\institute{corresponding author: \\
              J. K\r{u}dela \at
              Institute of Automation and Computer Science \\
              Brno University of Technology \\
              Technicka 2, Brno \\
              Czech Republic \\
              Tel.: +420541143358 \\
              \email{Jakub.Kudela@vutbr.cz}       
}

\date{Received: DD Month YEAR / Accepted: DD Month YEAR}

\maketitle

\begin{abstract}
Standard evolutionary optimization algorithms assume that the evaluation of the objective and constraint functions is straightforward and computationally cheap. However, in many real-world optimization problems, these evaluations involve computationally expensive numerical simulations or physical experiments. Surrogate-assisted evolutionary algorithms (SAEAs) have recently gained increased attention for their performance in solving these types of problems. The main idea of SAEAs is the integration of an evolutionary algorithm with a selected surrogate model that approximates the computationally expensive function. In this paper, we propose a surrogate model based on a Lipschitz underestimation and use it to develop a differential evolution-based algorithm. The algorithm, called Lipschitz Surrogate-assisted Differential Evolution (LSADE), utilizes the Lipschitz-based surrogate model, along with a standard radial basis function surrogate model and a local search procedure. The experimental results on seven benchmark functions of dimensions 30, 50, 100, and 200 show that the proposed LSADE algorithm is competitive compared with the state-of-the-art algorithms under a limited computational budget, being especially effective for the very complicated benchmark functions in high dimensions.
\end{abstract}

\section{Introduction}
\label{intro}
Many real-world optimization problems involve expensive computations, such as computational fluid dynamics and finite element analysis, or executions of physical experiments. In such situations, the evaluation of objective functions or constraints can take an excessively long time, prohibiting the use of conventional optimization methods \cite{Jin09}. To mitigate the computational costs, surrogate models (sometimes called metamodels \cite{Emmerich02}) have been widely used in combination with evolutionary algorithms (EAs), which are known as surrogate-assisted EAs (SAEAs) \cite{Jin19}. 

SAEAs execute only a limited number of real objective function (or constraint) evaluations and use these evaluations to train surrogate models. The surrogate models then serve as approximations of the real functions \cite{Jin11}, and their evaluation should have negligible computational costs compared to evaluating the real functions. Many standard machine learning models, such as polynomial response surface \cite{Myers95}, Kriging (or Gaussian processes) \cite{Martin05}, artificial neural networks \cite{Jin04}, radial basis functions (RBFs) \cite{Dyn86} or support vector regression \cite{Gunn98} have been employed in SAEAs. The performance of different surrogate models under multiple criteria was investigated in \cite{Jin01}. 

EAs are effective metaheuristics used for global optimization, which are inspired by the processes of biological evolution, such as reproduction, mutation, and natural selection. The most widely known examples of these techniques are genetic algorithms (GA), differential evolution (DE), evolutionary strategy (ES), or particle swarm optimization (PSO). These methods were successfully used in the optimization of various complex problems such as the hyperparameter optimization in deep learning \cite{young2015optimizing}, difficult assignment problems \cite{matousek2017}, design of quantum operators \cite{vzufan2021advances}, dynamical systems prediction \cite{R1}, or solving boundary value problems \cite{R3}.

Surrogate models are being employed in a variety of real-world problems, including protein structure prediction \cite{Rakh19}, elastic actuator design \cite{Dong21}, structural optimization design of truss topology \cite{Wang14} or robust optimization of large scale networks \cite{Wang20}. A review of recent advances and applications of surrogate models for finite element method computations can be found in \cite{SC}.

Based on the current surrogate model, the SAEAs typically choose two types of solutions for real function evaluation: promising samples around the optimum of the surrogate model, and uncertain samples with a large expected approximation error. For example, in \cite{Regis14} the authors designed multiple trial positions for each particle and then used an RBF model to select a position with the minimum predicted fitness value. A global and a local surrogate-assisted PSO algorithm for computationally expensive problems was developed in \cite{Sun14}. Here, the particle with a smaller predicted fitness value than its personal historical best was exactly evaluated. The uncertain samples were used to guide the search into some sparse and not yet well-explored areas, while the promising samples were used to guide a local search in the most promising areas. Many combinations of the two types are used to keep a good balance of global exploration and local exploitation. For instance, \cite{Liu14} developed a dimension reduction method to construct a Kriging surrogate model in a lower-dimensional space and chose the offspring with better lower confidence bound (LCB) values for real function evaluation. The LBC values were also used in \cite{Li19}, where the authors employed two different surrogate models. Here, the weight coefficient of the two models was changed to control the evolutionary progress. Another approach utilizing a trust region method for the interleaved use of exact models with computationally inexpensive RBF surrogates during a local search was developed in \cite{Ong03}. 

Surrogate models can guide the search of EAs to promising directions by using optima of these models, as was demonstrated in \cite{Regis14}, \cite{SACOSO17}, and many others. It has also been shown that evaluating the uncertain samples can strengthen the exploration capabilities of SAEAs and effectively improve the approximation accuracy of the surrogate \cite{Emmerich02}, \cite{Jin11}, and different methods for estimating the degree of uncertainty in function prediction have been proposed \cite{Wang16}.

In recent years, there has been a multitude of SAEAs proposed in the literature. These algorithms usually employ a metaheuristic algorithm to be the primary optimization framework and use the surrogates as additional tools to accelerate the convergence of the underlying metaheuristic algorithm. In general, it is difficult for EAs to search for global optima in high-dimensional spaces because of the curse of dimensionality. SAEAs also encounter the same challenge when the dimension of a problem is high. Although current SAEAs can handle high-dimensional expensive problems relatively well, most of these algorithms still need many function evaluations (usually more than several thousands) to obtain good optimization results. Also, these algorithms are developed for optimizing problems whose dimensions are usually less than 30. For instance, the generalized surrogate single-objective memetic algorithm proposed in \cite{Lim10} needs 8000 function evaluations for 30D problems. The surrogate-assisted DE algorithm introduced \cite{Mallipeddi15} needs more than 10000 function evaluations for 30D problems. A similarly high number of required function evaluations were utilized by Lipschitz-based algorithm in \cite{R2R1}. A framework combining particle swarm optimization and RBF global surrogate was developed in \cite{Regis14}, where the proposed method first generates multiple candidate solutions for each particle in each generation, and then the surrogate is employed to select the promising positions to form the new population. The Gaussian process model was utilized in \cite{Liu14} with the lower confidence bound to prescreen solutions in a differential evolution (DE) algorithm and a dimensional reduction technique was used to enhance the accuracy of the model. The maximum dimension of the test problems used in \cite{Liu14} was 50 and the dimension was reduced to 4 before the surrogate was constructed. An alternative approach for this issue is the use of multiple swarms, that can enhance population diversity, explore different search spaces simultaneously to efficiently find promising areas, and combine the advantage of different swarms if heterogeneous swarms are used. For computationally expensive problems, multiple swarms were used in the surrogate-assisted multiswarm optimization (SAMSO) algorithm \cite{Li21}. The SAMSO algorithm takes advantage of the good global searchability of the teaching learning-based optimization algorithm and the fast convergence ability of the PSO algorithm.

Multiple surrogates have been shown to perform better than single ones in assisting EAs, typically utilizing a global surrogate model to smooth out the local optima, and local surrogate models to capture the local details of the fitness function around the neighborhood of the current best individuals. In \cite{Wans17} an ensemble surrogate-based model management method for surrogate-assisted PSO was proposed. This method searches for the promising and most uncertain candidate solutions to be evaluated using the expensive fitness function. Their results were outstanding on medium-scale test functions with a limited number of function evaluations. Surrogate-assisted cooperative swarm optimization (SA-COSO) for high-dimensional expensive problems, developed in \cite{SACOSO17}, combined two PSO methods to solve problems with dimension up to 200. Another algorithm for high dimensional expensive problems, called evolutionary sampling assisted optimization (ESAO), which utilized a global RBF model and a local optimizer, was developed in \cite{ESAO19}.

A generalized surrogate-assisted evolutionary algorithm (GSGA) based on the optimization framework of the genetic algorithm was proposed in \cite{Cai20}. This algorithm uses a surrogate-based trust region local search method, a surrogate-guided GA updating mechanism with a neighbor region partition strategy, and a prescreening strategy based on the expected improvement infilling criterion of a simplified Kriging in the optimization process. A multi-objective infill criterion for a Gaussian process assisted social learning particle swarm optimization (MGP-SLPSO) algorithm was proposed in \cite{Tian19}. The multi-objective infill criterion considers the approximated fitness and the approximation uncertainty as two objectives and uses non-dominated sorting for model management. Surrogate-assisted grey wolf optimization (SAGWO) algorithm was introduced in \cite{Dong20}, where RBF is employed as the surrogate model. SAGWO conducts the search in three phases, initial exploration, RBF-assisted meta-heuristic exploration, and knowledge mining on RBF.

In this paper, we propose a novel Lipschitz-based surrogate model, that is designed to increase the exploration capabilities of SAEAs. We also develop a new Lipschitz surrogate-assisted differential evolution (LSADE) algorithm that uses the Lipschitz-based surrogate in combination with a standard RBF surrogate and a local optimization procedure. The rest of this paper is organized as follows. Section \ref{S:2} briefly introduces the related techniques, including surrogate models, Lipschitz-based underestimation, and DE. Section \ref{S:3} describes the proposed LSADE algorithm in detail. In Section \ref{S:4}, we provide a computational analysis of the individual components of the LSADE algorithm, the frequency of the utilization of said components, the choice of an RBF, and a comparison with other state-of-the-art SAEAs, namely with SA-COSO, ESAO, SAMSO, GSGA, MGP-SLPSO, and SAGWO. The conclusions and future research directions are described in Section \ref{S:5}.

\section{Related Techniques} \label{S:2}

\subsection{Surrogate Models}
Kriging models and RBFs are the most widely applied methods for generating surrogate models \cite{Forrester09}. It has been shown that the Kriging model outperforms other surrogate models in solving low-dimensional optimization problems, and RBF is the most efficient method among surrogates for solving high-dimensional optimization problems \cite{Diaz16}. A disadvantage of Kriging is that the training of the model is time-consuming when the number of samples is large. Since this paper focuses on high-dimensional problems, we will adopt the RBF methodology for building the surrogate model, which has been successfully used in several other SAEAs \cite{SACOSO17}.

RBFs compute a weighted sum of prespecified simple functions to approximate complex design landscape. Given $t$ different sample points $X_1,\dots,X_t$, the RBF surrogates are written as \cite{SC}

$$
f_\text{RBF} (x) = \sum_{i=1}^t w_i \psi(||x - X_i ||_2),
$$
where $w_i$ denotes the weight which is computed using the method of least squares, and $\psi$ is the chosen basis function. There are several (symmetric) radial functions that can serve as a basis function, such as Gaussian function, thin-plate splines, linear splines, cubic splines, and multiquadrics splines \cite{SC}.

\subsection{Lipschitz-based Underestimation}
The use of a Lipschitz constant in optimization was first proposed in \cite{Piy72} and \cite{Shu72} and initiated a line of research within global optimization that is active to this day \cite{LIPO17}. We assume that the unknown or expensive to compute objective function $f$ has a finite Lipschitz constant $k$, i.e. 

$$
\exists k \geq 0 \,\, \textnormal{s.t. }\,\,|f(x) - f(x')| \leq k ||x-x'||_2 \,\, \forall (x,x') \in \mathcal{X}^2,
$$
which is among the weakest regularity assumptions we can ask for. Based on a sample of $t$ evaluations of the function $f$ at points $X_1,\dots,X_t$, we can construct a global underestimator $f_L$ of $f$ by using the following expression \cite{LIPO17}
\begin{equation}
    f_L(x) = \max_{i=1,\dots,t} f(X_i) - k || x-X_i||_2. \label{eq1}
\end{equation}

\begin{figure}
    \centering
    \includegraphics[width = 1\linewidth]{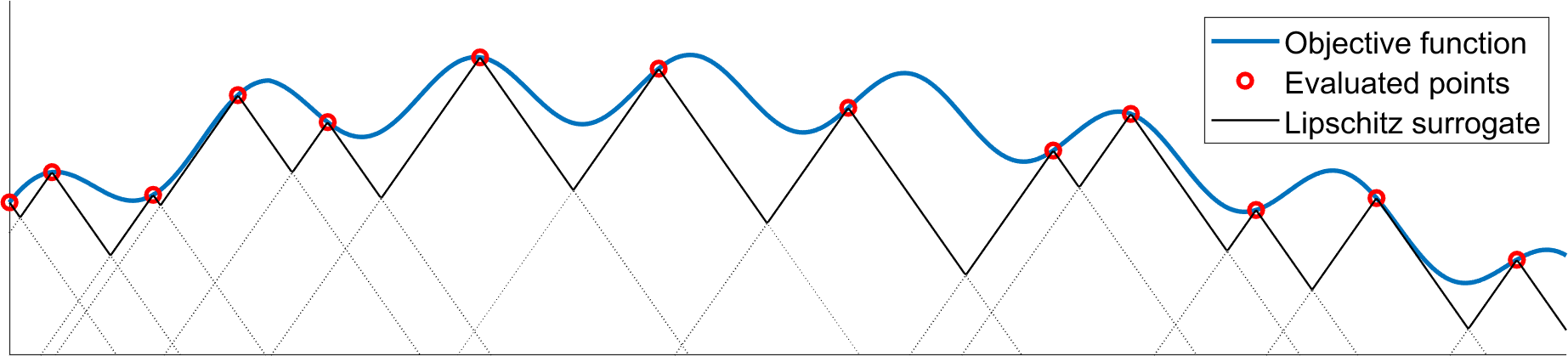}
    \caption{Visual representation of the Lipschitz-based surrogate in 1D.}
    \label{fig:lipo}
\end{figure}

\begin{figure}
    \centering
    \includegraphics[width = 1\linewidth]{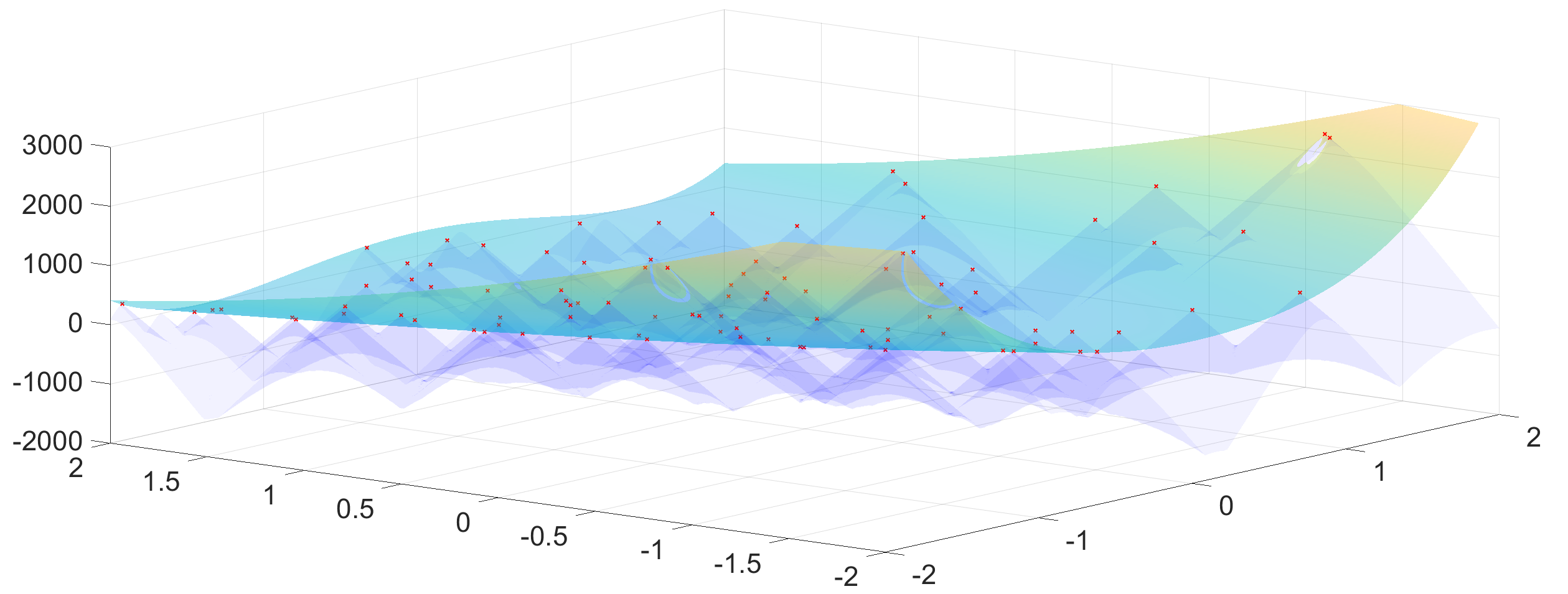}
    \caption{Visual representation of the Lipschitz-based surrogate on the Rosenbrock function in 2D. Sampled points are highlighted in red and the Lipschitz-based surrogate in light blue.}
    \label{fig:2D}
\end{figure}

A visual representation of this Lipschitz-based surrogate function in 1D is depicted in Figure \ref{fig:lipo}, where each already evaluated point has two lines (one to the left and the other to the right) emanating from it under an angle that depends on the Lipschitz constant $k$. Then the surrogate is constructed as the pointwise maximum of the individual lines. A 2D visualization is shown in Figure \ref{fig:2D}. This surrogate has two important properties -- it assigns low values to points that are far from previously evaluated points and combines it with the information (objective value and ``global'' Lipschitz constant) from the closest evaluated point. Therefore, it can serve as a good ``uncertainty measure'' of prospective points for evaluation, as points with low values of $f_L$ are either far from any other evaluated solution, or relatively close to a good one. 

Naturally, since we do not know the objective function $f$ itself, we can hardly expect to know the Lipschitz constant $k$. We will approach this issue by estimating $k$ from the previously evaluated points. We will use the approach described in \cite{LIPO17}, which utilizes a nondecreasing sequence of Lipschitz constants $k_{i \in \mathbf{Z}}$ that defines a meshgrid on $\mathbf{R}^+$. The estimate $\hat{k}_t$ of the Lipschitz constant is then computed as
\begin{equation} \label{eq2a}
    \hat{k}_t = \inf \left\{ k_{i\in \mathbf{Z}}: \max_{l \neq j} \frac{|f(X_j) - f(X_l)|}{||X_j - X_l ||_2} \leq k_i \right\}.
\end{equation}

Sequences of different shapes could be considered -- we utilize a sequence $k_i = (1+\alpha)^i$ that uses a parameter $\alpha > 0$. For this sequence, the computation (\ref{eq2a}) of the estimate is simplifies into $\hat{k}_t = (1+\alpha)^{i_t}$, where
\begin{equation}
    i_t = \left \lceil{ \ln (\max_{l \neq j} \frac{|f(X_j) - f(X_l)|}{||X_j - X_l ||_2} )/ \ln(1+\alpha)   }\right \rceil. \label{eq2}
\end{equation}

\subsection{Differential Evolution}
EAs are powerful methods for solving complex engineering optimization problems, that are difficult to approach with standard optimization methods. In this work, DE is employed as the optimization solver due to its straightforward structure and its global optimization capabilities. Several variants of DE have been developed to improve its performance \cite{Das11}. In general, there are four stages of DE: initialization, mutation, crossover, and selection. We assume we have a population at the current generation, $x = [x_1,\dots,x_t]$, where each individual has dimension $D$, $x_i = (x^1_i,\dots,x^D_i)$. In this work, we utilize the DE/best/1 strategy for the mutation process of DE which, can be expressed as
\begin{equation}
    v_i = x_b + F\cdot (x_{i_1} - x_{i_2}), \label{eq:DE1}
\end{equation}
where $x_b$ is the current best solution, $x_{i_1}$ and $x_{i_2}$ are different randomly selected individuals from the population, and $F$ is a scalar number typically within the interval [0.4, 1] \cite{Das11}. The crossover stage of DE is conducted after mutation and has the following form:
\begin{equation} \label{eq:DE2}
    u_i^j = \begin{cases}
v_i^j, \quad \textnormal{if } (U_j(0,1) \leq C_r \, | \, j = j_{rand}), \\ 
x_i^j, \quad \textnormal{otherwise,}
\end{cases}
\end{equation}
where $u_i^j$ the $j$th component of $i$th offspring, $x_j^i$ and $v_j^i$ are the $j$th component of $i$th parent individual and the mutated individual, respectively. The crossover constant $C_r$ is between 0 and 1, $U_j(0,1)$ indicates a uniformly distributed random number, and $j_{rand} \in [1,\dots,D]$ is a randomly chosen index that ensures $u_i$ has at least one component of $v_i$. The interested reader can find more information about the intricacies of DE in \cite{Das11}.

\section{Proposed LSADE Method} \label{S:3}
The proposed LSADE method has four distinct parts: 1) the DE-based generation of prospective points, 2) the global RBF evaluation of the prospective points, 3) the Lipschitz surrogate evaluation of the prospective points, and 4) the local optimization within a close range of the best solution found so far. The execution of parts 2) -- 4) of the algorithm can be controlled based on chosen conditions, i.e., we may sometimes skip RBF surrogate evaluation, Lipschitz surrogate evaluation, or local optimization, if deemed advantageous.

At the beginning of the process, Latin hypercube sampling \cite{Forrester09} is used to generate the initial population of $t$ individuals, whose objective function is evaluated \cite{HTS}. The best individual is found, a parent population of size $p$ is randomly selected from the evaluated points and a new population is constructed based on the DE rules (\ref{eq:DE1}) and (\ref{eq:DE2}). If the \textit{RBF evaluation condition} is true, the new population is evaluated based on the RBF surrogate model. Then the best individual based on the RBF model has its objective function evaluated and is added to the whole population. This step constitutes a global search strategy.

\begin{algorithm}[t!]
\caption{Pseudocode of the LSADE.}\label{A:1}
\begin{algorithmic}[1]
\State Generate an initial population of $t$ points $X_1,\dots,X_t$ and evaluate their objective function values. Denode the best solution as $X_b$.
\State Set $iter = 0$ (iteration counter), $NFE$ = $t$ (number of function evaluations).
\State Use the evaluated points so far to estimate $k$ by (\ref{eq2}) and to construct the RBF surrogate.
\State Sample $p$ points from the population as parents for DE.
\State Based on the DE rules (\ref{eq:DE1}) and (\ref{eq:DE2}), generate children.
\State Increase $iter$ by 1.
\If {\textit{RBF condition}}
\State Evaluate the children on the RBF surrogate.
\State Find the child with the minimum RFB surrogate value, and add it to the population and evaluate its objective function value. Increase $NFE$ by 1.
\EndIf
\If {\textit{Lipschitz condition}}
\State Evaluate the children on the Lipschitz surrogate (\ref{eq1}).
\State Find the child with the minimum Lipschitz surrogate value, and add it to the population and evaluate its objective function value. Increase $NFE$ by 1.
\EndIf
\If {\textit{Local Optimization condition}}
\State Construct a RBF local surrogate model using the best $c$ solutions found so far.
\State Find the bounds in each dimension for the local optimization (\ref{eq5}).
\State Minimize the local RBF surrogate model within the bounds. Denote the minimum as $\hat{X}_m$ and, if it is not already in the population, add it to the population and evaluate its objective function value. Increase $NFE$ by 1.
\EndIf
\State Find the best solution so far and denote it as $X_b$.
\If {$NFE < NFE_{max}$}
\State \textbf{goto} \emph{3}.
\Else
\State \textbf{terminate}.
\EndIf
\end{algorithmic}
\end{algorithm}

If the \textit{Lipschitz evaluation condition} is true, the Lipschitz constant $k$ is estimated based on (\ref{eq2}) and the new population is evaluated on the Lipschitz surrogate model (\ref{eq1}). The best individual based on the Lipschitz surrogate model has its objective function evaluated and is added to the whole population. 

If the \textit{Local optimization condition} is true, we construct a local RBF surrogate model using the best $c$ solutions found so far, which we denote by $\hat{X}_1,\dots,\hat{X}_c$. Additionally, we find the bounds for the local optimization procedure within those $c$ points:
\begin{align} \label{eq5}
\begin{split}
    lb(i) = \min_{j=1,\dots,c}{\hat{X}_j(i)}, \quad i=1,\dots,D,\\ 
    ub(i) = \max_{j=1,\dots,c}{\hat{X}_j(i)}, \quad i=1,\dots,D,
    \end{split}
\end{align}
and perform a local optimization of the local RBF model within the bounds $[lb,ub]$. For local optimization we adapt a sequential quadratic programming strategy, which was also used by the winner of the 2020 CEC Single Objective Bound Constrained Competition \cite{IMODE20}. We find the local optimum and check, if it is not already in the population, before evaluating it and adding it to the population. 

The evaluation of points based on the Lipschitz-based surrogate model can be thought of as an exploration step in the algorithm (and should increase our ability to find the regions of good solutions), whereas the evaluation of points based on the local optimization procedure can be thought of as an exploitation step of the algorithm (and should give us the means to improve the best solutions we have found so far).

The cycle of generating new population, evaluating it on the RBF and Lipschitz surrogate models and conducting the local optimization is carried out until a maximum number of objective function evaluations is reached. The pseudocode\footnote{The MATLAB code can be found at the authors github: \\  \url{https://github.com/JakubKudela89/LSADE}} for the LSADE method is described in Algorithm \ref{A:1}. 

\section{Results and Discussion} \label{S:4}
To examine the effectiveness of the proposed method, we compare it with six other state-of-the-art algorithms on a testbed of standard benchmark functions \cite{Suga05} that are summarized in Table \ref{tab:1}. Although there are more recent benchmark sets, such as \cite{ZZ2}, these were not yet used for benchmarking SAEAs. The dimensions for the comparison are $D=30,50,100,200$ for all of the benchmark functions. We also investigate the advantages of the individual components of the LSADE method, the choice of the conditions for using the different components, and the choice of basis functions for the RBF surrogates. The algorithm is implemented in MATLAB R2020b and runs on an Intel(R) Core(TM) i5-4460 CPU @ 3.20 GHz desktop PC.

\begin{table}[]
    \centering
    \caption{Benchmark functions used for the comparison}
    \begin{tabular}{llll}
         Problem & Description & Property & Optimum\\ \hline 
         F1 & Ellipsoid & Unimodal & 0 \\
         F2 & Rosenbrock & Multimodal with narrow valley & 0 \\
        F3 & Ackley & Multimodal & 0 \\
        F4 & Griewank&  Multimodal & 0 \\ 
        F5 & F10 in \cite{Suga05} & Very complicated multimodal & -330 \\
        F6 & F16 in \cite{Suga05} & Very complicated multimodal & 120 \\
        F7 & F19 in \cite{Suga05} &  Very complicated multimodal & 10
    \end{tabular}
    \label{tab:1}
\end{table}

\begin{table}[h!]
\centering
\caption{Comparison of the individual components of LSADE on $D=[30,50]$. } \label{t:2}
\bgroup
\def\arraystretch{1.4}
\setlength{\tabcolsep}{1.7pt}
\begin{tabular}{ll|r|rrr|rrrr} 
 $D$&  F  & {R0$\,|\,$Li0$\,|\,$Lo0} & {R0$\,|\,$Li0$\,|\,$Lo1}    & {R0$\,|\,$Li1$\,|\,$Lo0}   & {R0$\,|\,$Li1$\,|\,$Lo1}    & {R1$\,|\,$Li0$\,|\,$Lo0}   & {R1$\,|\,$Li0$\,|\,$Lo1}   &{R1$\,|\,$Li1$\,|\,$Lo0}    &{R1$\,|\,$Li1$\,|\,$Lo1}  \\ \hline \hline
                         & F1 & \cellcolor[HTML]{EFEFEF}1898                         & \cellcolor[HTML]{F2B8B3}124.3  & \cellcolor[HTML]{E67C73}222.1 & \cellcolor[HTML]{FFFFFF}7.787       & \cellcolor[HTML]{ABDDC5}3.660   & \cellcolor[HTML]{57BB8A}0.0041   & \cellcolor[HTML]{FFFFFF}7.237       & \cellcolor[HTML]{57BB8A}0.010  \\
                         & F2 & \cellcolor[HTML]{EFEFEF}4641                         & \cellcolor[HTML]{F4C6C2}193.2  & \cellcolor[HTML]{E67C73}379.9  & \cellcolor[HTML]{FEFAF9}60.79   & \cellcolor[HTML]{B0DFC8}38.55   & \cellcolor[HTML]{5CBD8D}30.32  & \cellcolor[HTML]{FFFFFF}46.32      & \cellcolor[HTML]{57BB8A}29.79   \\
                         & F3 & \cellcolor[HTML]{EFEFEF}20.35                        & \cellcolor[HTML]{E77E76}16.94   & \cellcolor[HTML]{FEF9F9}13.55  & \cellcolor[HTML]{F6FBF8}12.96   & \cellcolor[HTML]{EFF8F4}12.63  & \cellcolor[HTML]{E67C73}16.99  & \cellcolor[HTML]{57BB8A}5.399    & \cellcolor[HTML]{FFFFFF}13.37    \\
                         & F4 & \cellcolor[HTML]{EFEFEF}467.0                       & \cellcolor[HTML]{E67C73}109.1  & \cellcolor[HTML]{F4C6C2}53.63  & \cellcolor[HTML]{FFFFFF}9.595       & \cellcolor[HTML]{65C094}1.234   & \cellcolor[HTML]{FFFEFE}10.69  & \cellcolor[HTML]{74C69E}2.030    & \cellcolor[HTML]{57BB8A}0.431  \\
                         & F5 & \cellcolor[HTML]{EFEFEF}434.7                       & \cellcolor[HTML]{FFFBFA}-126.9 & \cellcolor[HTML]{E67C73}33.95  & \cellcolor[HTML]{FDF1F0}-114.6 & \cellcolor[HTML]{FFFFFF}-133.2   & \cellcolor[HTML]{D5EEE2}-153.7 & \cellcolor[HTML]{FEFEFE}-133.4 & \cellcolor[HTML]{57BB8A}-216.9 \\
                         & F6 & \cellcolor[HTML]{EFEFEF}1154                         & \cellcolor[HTML]{E67C73}814.7  & \cellcolor[HTML]{FFFFFF}587.6    & \cellcolor[HTML]{8DD1AF}488.4  & \cellcolor[HTML]{FDF3F2}608.8 & \cellcolor[HTML]{FEF6F6}603.3 & \cellcolor[HTML]{90D2B1}490.8  & \cellcolor[HTML]{57BB8A}440.7 \\
\multirow{-7}{*}{30}    & F7 & \cellcolor[HTML]{EFEFEF}1348                         & \cellcolor[HTML]{E67C73}1194    & \cellcolor[HTML]{FFFFFF}987.9    & \cellcolor[HTML]{57BB8A}959.2  & \cellcolor[HTML]{F7D0CD}1062   & \cellcolor[HTML]{F4C1BD}1086   & \cellcolor[HTML]{BDE4D1}976.7   & \cellcolor[HTML]{A7DBC2}973.0 \\ \hline \hline
                         & F1 & \cellcolor[HTML]{EFEFEF}6365                         & \cellcolor[HTML]{FEF6F5}148.5 & \cellcolor[HTML]{E67C73}1131   & \cellcolor[HTML]{61BF91}6.645    & \cellcolor[HTML]{FAE5E3}285.5  & \cellcolor[HTML]{5ABC8C}3.727   & \cellcolor[HTML]{FFFFFF}69.96     & \cellcolor[HTML]{57BB8A}2.352    \\
                         & F2 & \cellcolor[HTML]{EFEFEF}10279                          & \cellcolor[HTML]{FCEEED}283.5 & \cellcolor[HTML]{E67C73}1070   & \cellcolor[HTML]{70C59B}79.83   & \cellcolor[HTML]{FEF8F7}214.4 & \cellcolor[HTML]{57BB8A}65.41  & \cellcolor[HTML]{FFFFFF}161.8    & \cellcolor[HTML]{57BB8A}65.12   \\
                         & F3 & \cellcolor[HTML]{EFEFEF}20.59                        & \cellcolor[HTML]{EFA7A1}17.45  & \cellcolor[HTML]{FCFDFD}15.48  & \cellcolor[HTML]{B5E1CB}13.38   & \cellcolor[HTML]{E67C73}18.36  & \cellcolor[HTML]{EA8E86}17.98  & \cellcolor[HTML]{57BB8A}10.58  & \cellcolor[HTML]{FFFFFF}15.56      \\
                         & F4 & \cellcolor[HTML]{EFEFEF}926.6                       & \cellcolor[HTML]{E67C73}307.0 & \cellcolor[HTML]{F6D0CD}162.5 & \cellcolor[HTML]{7AC9A2}21.87   & \cellcolor[HTML]{FFFFFF}79.97     & \cellcolor[HTML]{F3BFBA}191.9 & \cellcolor[HTML]{5DBD8E}9.117   & \cellcolor[HTML]{57BB8A}6.463    \\
                         & F5 & \cellcolor[HTML]{EFEFEF}1185                         & \cellcolor[HTML]{FFFFFF}30.64     & \cellcolor[HTML]{E67C73}396.1 & \cellcolor[HTML]{66C194}-122.9& \cellcolor[HTML]{EFA9A3}272.9 & \cellcolor[HTML]{F4FAF7}20.02  & \cellcolor[HTML]{F7D0CD}161.9 & \cellcolor[HTML]{57BB8A}-138.0 \\
                         & F6 & \cellcolor[HTML]{EFEFEF}1276                         & \cellcolor[HTML]{E67C73}880.6 & \cellcolor[HTML]{FFFFFF}679.3    & \cellcolor[HTML]{57BB8A}368.9  & \cellcolor[HTML]{F2B9B4}787.4 & \cellcolor[HTML]{F6D0CC}752.7  & \cellcolor[HTML]{C2E6D4}567.8 & \cellcolor[HTML]{6DC499}410.4  \\
\multirow{-7}{*}{50}    & F7 & \cellcolor[HTML]{EFEFEF}1460                       & \cellcolor[HTML]{E67C73}1296   & \cellcolor[HTML]{FFFFFF}1086      & \cellcolor[HTML]{57BB8A}1019    & \cellcolor[HTML]{EEA6A0}1229   & \cellcolor[HTML]{EDA19A}1238   & \cellcolor[HTML]{9ED7BB}1047   & \cellcolor[HTML]{E7F5EE}1077 \\ \hline \hline
\end{tabular}
\egroup
\end{table}

\subsection{Experiment Setting}
For constructing both the local and the global RBF surrogate models we used the SURROGATES toolbox \cite{Viana11} with default settings (multiquadric RBF with parameter $c = 1$). The DE coefficients were set to $F = 0.5$ and $C_r = 0.5$ \cite{kaz}. The number of initial points were set to 100 for $D=[30,50]$ and 200 for $D=[100,200]$. The number of children was set to $D$. The local optimization uses the best $c=3\cdot D$ points found so far (or less if there are not enough points yet evaluated), and utilizes the sequential quadratic programming algorithm implemented in the FMINCON function with default parameters. The Lipschitz approximation parameter was set to $\alpha = 0.01$. The maximum number of function evaluations was set to 1000 for all problems. For all benchmark functions, 20 independent runs are conducted to get statistical results. Finally, some of the more in-depth results regarding the sensitivity of the parameters of the LSADE algorithm are studied in the Appendix.

\subsection{Comparison of Individual Components}
Firstly, we assess the effectiveness of the individual components of the LSADE: the RBF surrogate, the Lipschitz surrogate, the local optimization procedure, and their combinations. This corresponds to setting the \textit{RBF condition, Lipschitz condition}, and \textit{Local Optimization condition} to true or false (1 or 0) for every iteration of the algorithm. We denote the 8 possible variations as a triplet (R -- RBF, Li -- Lipschitz, Lo -- Local Optimization) R\#$\,|\,$Li\#$\,|\,$Lo\#, where the ``\#'' indicates if the condition was true or false. The R0$\,|\,$Li0$\,|\,$Lo0 variation does not use any optimization (as there is no rule to add points for evaluation) and instead just evaluates 1000 randomly selected points, using the entire computational budget. The results (mean of the best-found objective function values over the 20 runs) for the different variations in dimensions $D=[30,50]$ are reported in Table \ref{t:2}. Not surprisingly, the R0$\,|\,$Li0$\,|\,$Lo0 variant comes out being substantially worse than the other ones and is the only one that has its cells in the table colored in grey. The remaining variations are color-coded in the following way: the variant with the best (lowest) mean objective function value for a given problem instance has the corresponding cell in the table colored in a dark shade of green, the one with the worst (highest) mean objective function value has a dark red color, and the ones in between are ordered from green (better) to red (worse). This paradigm is also used in the subsequent tables for making straightforward comparisons. From Table \ref{t:2} we can see that the ``usefulness'' of the individual components of LSADE is very problem-dependent, as there are instances, where adding either component may be beneficial or detrimental. However, based on the results, it seems advantageous to have the \textit{RFB condition} be true, as the majority of the best results (11 of the 14 instances) were achieved by the R1 variants. As for the other two components, the situation is more nuanced -- it is clear that they are both beneficial (the best results are always in a variant with either Li1 or Lo1), but the trade-off between adding one or the other needs to be be investigated in more detail.

\begin{table}[]
\centering
\caption{Comparison of the static rules for the Lipschitz and Local Optimization conditions, $D=[30,50].$} \label{t:3}
\bgroup
\def\arraystretch{1.4}
\setlength{\tabcolsep}{1.5pt}
\begin{tabular}{ccc}
\begin{tabular}{r|rrrrr}
\multicolumn{6}{c}{F1 [0.0036,\,0.3671]} \\ 
Li$\,|\,$Lo & \multicolumn{1}{c}{1}                               & \multicolumn{1}{c}{2}                                & \multicolumn{1}{c}{4}                                & \multicolumn{1}{c}{8}                                & \multicolumn{1}{c}{0}                              \\ \hline
1                                        & \cellcolor[HTML]{F3FAF6}0.0102 & \cellcolor[HTML]{FFFFFF}0.0342    & \cellcolor[HTML]{FFFDFD}0.1328 & \cellcolor[HTML]{FEF9F9}0.3671 & \cellcolor[HTML]{E67C73}7.2373 \\
2                                        & \cellcolor[HTML]{96D4B6}0.0063 & \cellcolor[HTML]{CAE9DA}0.0085 & \cellcolor[HTML]{FFFFFF}0.0223    & \cellcolor[HTML]{FFFFFF}0.041     & \cellcolor[HTML]{FAE3E1}1.5932 \\
4                                        & \cellcolor[HTML]{62BF92}0.0041 & \cellcolor[HTML]{73C69D}0.0048 & \cellcolor[HTML]{CFEBDE}0.0087 & \cellcolor[HTML]{FFFFFF}0.0129    & \cellcolor[HTML]{FDF3F2}0.6838 \\
8                                        & \cellcolor[HTML]{71C59C}0.0047 & \cellcolor[HTML]{57BB8A}0.0036 & \cellcolor[HTML]{94D3B4}0.0062 & \cellcolor[HTML]{FFFFFF}0.0107    & \cellcolor[HTML]{FFFAFA}0.2917 \\
0                                        & \cellcolor[HTML]{62BF92}0.0041 & \cellcolor[HTML]{6EC49A}0.0046 & \cellcolor[HTML]{FFFFFF}0.0179    & \cellcolor[HTML]{96D4B6}0.0063 & \cellcolor[HTML]{F3BDB9}3.6604
\end{tabular} & 
\begin{tabular}{r|rrrrr}
\multicolumn{6}{c}{F2 [25.90,\,32.59]} \\ 
Li$\,|\,$Lo & \multicolumn{1}{c}{1}                               & \multicolumn{1}{c}{2}                                & \multicolumn{1}{c}{4}                                & \multicolumn{1}{c}{8}                                & \multicolumn{1}{c}{0}                              \\ \hline
1                    & \cellcolor[HTML]{FFFFFF}29.79      & \cellcolor[HTML]{FFFFFF}29.82    & \cellcolor[HTML]{ABDDC4}27.85 & \cellcolor[HTML]{A8DBC2}27.78  & \cellcolor[HTML]{E67C73}46.32 \\
2                    & \cellcolor[HTML]{DFF2E8}29.05 & \cellcolor[HTML]{FBE9E8}32.59 & \cellcolor[HTML]{F7FCF9}29.61 & \cellcolor[HTML]{CFEBDD}28.69 & \cellcolor[HTML]{EA8E86}44.16 \\
4                    & \cellcolor[HTML]{FEF6F6}30.93 & \cellcolor[HTML]{EFF8F4}29.42 & \cellcolor[HTML]{DDF1E7}29.02 & \cellcolor[HTML]{7DCAA4}26.79 & \cellcolor[HTML]{FAE0DE}33.70 \\
8                    & \cellcolor[HTML]{FFFCFC}30.21  & \cellcolor[HTML]{57BB8A}25.90 & \cellcolor[HTML]{FDF0EF}31.71 & \cellcolor[HTML]{6BC398}26.38  & \cellcolor[HTML]{F6CDCA}36.15 \\
0                    & \cellcolor[HTML]{FFFBFB}30.32 & \cellcolor[HTML]{A9DCC3}27.80 & \cellcolor[HTML]{C9E9D9}28.55 & \cellcolor[HTML]{FDF4F3}31.25 & \cellcolor[HTML]{F2BAB5}38.55 
\end{tabular} & 
\begin{tabular}{r|rrrrr}
\multicolumn{6}{c}{F3 [1.67,\,15.78]} \\ 
Li$\,|\,$Lo & \multicolumn{1}{c}{1}                               & \multicolumn{1}{c}{2}                                & \multicolumn{1}{c}{4}                                & \multicolumn{1}{c}{8}                                & \multicolumn{1}{c}{0}                              \\ \hline
1                    & \cellcolor[HTML]{F4C2BE}13.37 & \cellcolor[HTML]{D0ECDE}7.85  & \cellcolor[HTML]{63C092}2.30  & \cellcolor[HTML]{57BB8A}1.67   & \cellcolor[HTML]{A0D8BD}5.39  \\
2                    & \cellcolor[HTML]{F2B7B2}13.94 & \cellcolor[HTML]{FDF1F0}10.95 & \cellcolor[HTML]{B1DFC9}6.28  & \cellcolor[HTML]{7AC9A2}3.48  & \cellcolor[HTML]{93D3B4}4.74  \\
4                    & \cellcolor[HTML]{ED9F98}15.22 & \cellcolor[HTML]{F2BAB5}13.79 & \cellcolor[HTML]{ECF7F1}9.24  & \cellcolor[HTML]{B4E0CB}6.42  & \cellcolor[HTML]{C8E8D8}7.40  \\
8                    & \cellcolor[HTML]{EB948C}15.78 & \cellcolor[HTML]{EFAAA4}14.62 & \cellcolor[HTML]{F7FBF9}9.80  & \cellcolor[HTML]{FFFFFF}10.19    & \cellcolor[HTML]{F5FAF8}9.69  \\
0                    & \cellcolor[HTML]{E67C73}16.99 & \cellcolor[HTML]{EA8C84}16.17 & \cellcolor[HTML]{EEA39D}14.99 & \cellcolor[HTML]{F4C2BE}13.39 & \cellcolor[HTML]{F7D1CD}12.63
\end{tabular}  \vspace{3mm}
\\
\begin{tabular}{r|rrrrr}
\multicolumn{6}{c}{F4 [0.0035,\,1.107]} \\ 
Li$\,|\,$Lo & \multicolumn{1}{c}{1}           & \multicolumn{1}{c}{2}          & \multicolumn{1}{c}{4}          & \multicolumn{1}{c}{8}          & \multicolumn{1}{c}{0}          \\ \hline
1                    & \cellcolor[HTML]{E3F3EB}0.431  & \cellcolor[HTML]{B1DFC9}0.290 & \cellcolor[HTML]{FFFFFF}0.508    & \cellcolor[HTML]{FFFEFE}0.595 & \cellcolor[HTML]{FCECEB}2.030 \\
2                    & \cellcolor[HTML]{FFFEFE}0.586  & \cellcolor[HTML]{7DCAA4}0.144  & \cellcolor[HTML]{71C59C}0.109 & \cellcolor[HTML]{90D2B2}0.197 & \cellcolor[HTML]{FEF6F5}1.253 \\
4                    & \cellcolor[HTML]{FFFDFD}0.702  & \cellcolor[HTML]{75C79F}0.120 & \cellcolor[HTML]{65C093}0.075  & \cellcolor[HTML]{66C194}0.078 & \cellcolor[HTML]{FEF8F8}1.082\\
8                    & \cellcolor[HTML]{FEF8F7}1.107  & \cellcolor[HTML]{83CDA9}0.160 & \cellcolor[HTML]{57BB8A}0.035 & \cellcolor[HTML]{60BE90}0.061 & \cellcolor[HTML]{FEF9F8}1.025 \\
0                    & \cellcolor[HTML]{E67C73}10.69 & \cellcolor[HTML]{FBEAE8}2.193 & \cellcolor[HTML]{FFFEFE}0.615 & \cellcolor[HTML]{7BC9A3}0.139 & \cellcolor[HTML]{FEF6F6}1.234
\end{tabular} & 
\begin{tabular}{r|rrrrr}
\multicolumn{6}{c}{F5 [-222.1,\,-168.5]} \\ 
Li$\,|\,$Lo & \multicolumn{1}{c}{1}             & \multicolumn{1}{c}{2}             & \multicolumn{1}{c}{4}             & \multicolumn{1}{c}{8}             & \multicolumn{1}{c}{0}             \\ \hline
1                    & \cellcolor[HTML]{6AC297}-216.9   & \cellcolor[HTML]{57BB8A}-222.1 & \cellcolor[HTML]{7CCAA4}-212.1 & \cellcolor[HTML]{68C296}-217.3& \cellcolor[HTML]{E77D74}-133.4 \\
2                    & \cellcolor[HTML]{C6E8D7}-192.1 & \cellcolor[HTML]{90D2B1}-206.8 & \cellcolor[HTML]{C8E9D9}-191.6 & \cellcolor[HTML]{FBE6E4}-168.5 & \cellcolor[HTML]{E98981}-137.5 \\
4                    & \cellcolor[HTML]{D1ECDF}-189.3 & \cellcolor[HTML]{E9F6F0}-182.8 & \cellcolor[HTML]{E0F2E9}-185.3 & \cellcolor[HTML]{FFFFFF}-177.0    & \cellcolor[HTML]{EB928B}-140.4 \\
8                    & \cellcolor[HTML]{FDF4F3}-173.0 & \cellcolor[HTML]{F9FCFB}-178.5 & \cellcolor[HTML]{E4F4EC}-184.1 & \cellcolor[HTML]{FFFFFF}-177.0    & \cellcolor[HTML]{EA9189}-140.0 \\
0                    & \cellcolor[HTML]{F2BAB5}-153.7  & \cellcolor[HTML]{F4C6C2}-157.6 & \cellcolor[HTML]{FAE4E2}-167.8 & \cellcolor[HTML]{FAE2DF}-167.0 & \cellcolor[HTML]{E67C73}-133.2
\end{tabular} &
\begin{tabular}{r|rrrrr}
\multicolumn{6}{c}{F6 [418.7,\,558.4]} \\ 
Li$\,|\,$Lo & \multicolumn{1}{c}{1}          & \multicolumn{1}{c}{2}          & \multicolumn{1}{c}{4}          & \multicolumn{1}{c}{8}          & \multicolumn{1}{c}{0}          \\ \hline
1                    & \cellcolor[HTML]{89CFAD}440.7 & \cellcolor[HTML]{62BF92}423.8 & \cellcolor[HTML]{83CCA8}438.0 & \cellcolor[HTML]{81CCA7}437.4 & \cellcolor[HTML]{FCFDFC}490.8 \\
2                    & \cellcolor[HTML]{DAF0E5}476.1 & \cellcolor[HTML]{88CFAC}440.5 & \cellcolor[HTML]{BCE3D0}462.8& \cellcolor[HTML]{57BB8A}418.7 & \cellcolor[HTML]{FFFEFE}493.7 \\
4                    & \cellcolor[HTML]{FFFFFF}492.1    & \cellcolor[HTML]{C4E7D5}466.4 & \cellcolor[HTML]{D0ECDE}471.9 & \cellcolor[HTML]{BEE4D2}463.9 & \cellcolor[HTML]{FBEAE8}511.6 \\
8                    & \cellcolor[HTML]{F1B5B0}558.4 & \cellcolor[HTML]{F8D6D3}529.0 & \cellcolor[HTML]{FCECEB}509.1 & \cellcolor[HTML]{FFFCFC}495.0 & \cellcolor[HTML]{F4C5C1}543.8 \\
0                    & \cellcolor[HTML]{E8837A}603.3 & \cellcolor[HTML]{EA8E87}592.8 & \cellcolor[HTML]{E98981}597.7 & \cellcolor[HTML]{EB958E}587.1 & \cellcolor[HTML]{E67C73}608.9
\end{tabular} \vspace{3mm}
\\

\begin{tabular}{r|rrrrr}
\multicolumn{6}{c}{F7 [958.1,\,1036.4]} \\ 
Li$\,|\,$Lo & \multicolumn{1}{c}{1}          & \multicolumn{1}{c}{2}          & \multicolumn{1}{c}{4}          & \multicolumn{1}{c}{8}          & \multicolumn{1}{c}{0}          \\ \hline 
1                    & \cellcolor[HTML]{9CD7BA}973.1  & \cellcolor[HTML]{DCF0E6}986.9  & \cellcolor[HTML]{88CFAC}968.9  & \cellcolor[HTML]{60BE90}960.1  & \cellcolor[HTML]{ADDDC6}976.7  \\
2                    & \cellcolor[HTML]{94D3B4}971.4  & \cellcolor[HTML]{87CEAB}968.5  & \cellcolor[HTML]{A2D9BE}974.3  & \cellcolor[HTML]{57BB8A}958.1  & \cellcolor[HTML]{9BD6B9}972.8  \\
4                    & \cellcolor[HTML]{FBE9E7}1010 & \cellcolor[HTML]{FEF9F9}998.7  & \cellcolor[HTML]{FFFFFF}994.3     & \cellcolor[HTML]{C5E7D6}981.9  & \cellcolor[HTML]{FCF0EE}1005 \\
8                    & \cellcolor[HTML]{F4C4C0}1036.4 & \cellcolor[HTML]{FAE4E2}1013 & \cellcolor[HTML]{DDF1E7}987.1  & \cellcolor[HTML]{FAE4E2}1014 & \cellcolor[HTML]{F6CDCA}1029 \\
0                    & \cellcolor[HTML]{E67C73}1086.8 & \cellcolor[HTML]{EB948C}1070 & \cellcolor[HTML]{F3BEB9}1040 & \cellcolor[HTML]{F5C8C4}1033 & \cellcolor[HTML]{ED9F99}1062
\end{tabular}
& 
\multicolumn{2}{c}{
\begin{tabular}{r|rrrrrrr}
    \multicolumn{8}{c}{Min and max mean values from the static rules for $D=50$}\\ \multicolumn{8}{c}{(disregarding rules with Li0 or Lo0)} \\ 
     & \quad\quad F1 & \quad\quad F2 &\quad\quad F3 &\quad\quad F4 &\quad\quad F5 &\quad\quad F6 &\quad\quad F7 \\ \hline 
    min & 0.445 & 47.45 & 6.459 & 1.010 & -138.0 & 363.2 & 1019 \\     
    max &  6.003 & 65.37 & 16.78 & 71.78 & 0.750 & 615.8 & 1195 \\
\end{tabular}
} 

\end{tabular}
\egroup
\end{table}

\begin{table}[]
\centering
\caption{Comparison of the dynamic rules for the Lipschitz and Local Optimization conditions, $D=[30,50,100].$} \label{t:4}
\bgroup
\def\arraystretch{1.4}
\setlength{\tabcolsep}{3.5pt}
\begin{tabular}{rl|rrr|rrr|rrr}
$D$ & F \,\,\,[min,max] & 1-4$\,|\,$8-1            & 1-6$\,|\,$8-1            & 1-8$\,|\,$8-1            & 1-4$\,|\,$6-1            & 1-6$\,|\,$6-1            & 1-8$\,|\,$6-1            & 1-4$\,|\,$4-1           & 1-6$\,|\,$4-1            & 1-8$\,|\,$4-1            \\ \hline \hline 
                                             & F1 [0.0061,\,0.0113]                   & \cellcolor[HTML]{E67C73}0.0113    & \cellcolor[HTML]{FBEAE9}0.0087    & \cellcolor[HTML]{96D4B6}0.0069    & \cellcolor[HTML]{EFA7A1}0.0103    & \cellcolor[HTML]{EB968F}0.0107    & \cellcolor[HTML]{E7F5EE}0.0079    & \cellcolor[HTML]{FFFFFF}0.0082      & \cellcolor[HTML]{7ECBA5}0.0066    & \cellcolor[HTML]{57BB8A}0.0061    \\
                                             & F2 [26.63,\,27.06]                  & \cellcolor[HTML]{E67C73}27.06   & \cellcolor[HTML]{7ECBA5}26.69   & \cellcolor[HTML]{E9F6F0}26.83   & \cellcolor[HTML]{EA8E86}27.04   & \cellcolor[HTML]{EEA19B}27.01   & \cellcolor[HTML]{57BB8A}26.63   & \cellcolor[HTML]{F4C4C0}26.95   & \cellcolor[HTML]{FFFFFF}26.86      & \cellcolor[HTML]{A9DCC3}26.75   \\
                                             & F3 [1.122,\,3.496]                  & \cellcolor[HTML]{FFFFFF}1.308       & \cellcolor[HTML]{72C69D}1.152    & \cellcolor[HTML]{FEF5F4}1.480    & \cellcolor[HTML]{BCE4D0}1.235    & \cellcolor[HTML]{E4F4EC}1.279    & \cellcolor[HTML]{57BB8A}1.122    & \cellcolor[HTML]{F1B5B0}2.546   & \cellcolor[HTML]{E67C73}3.496    & \cellcolor[HTML]{E8877E}3.327    \\
                                             & F4 [0.013,\,0.051]                  & \cellcolor[HTML]{E67C73}0.051    & \cellcolor[HTML]{FFFFFF}0.033       & \cellcolor[HTML]{57BB8A}0.012    & \cellcolor[HTML]{FBE7E5}0.037     & \cellcolor[HTML]{F6D0CD}0.040    & \cellcolor[HTML]{8CD0AF}0.019    & \cellcolor[HTML]{F6CECB}0.040   & \cellcolor[HTML]{5CBD8D}0.013    & \cellcolor[HTML]{61BF91}0.014    \\
                                             & F5 [-218.7,\,-196.3]                  & \cellcolor[HTML]{57BB8A}-218.7 & \cellcolor[HTML]{F4FAF7}-213.4 & \cellcolor[HTML]{D4EDE1}-214.5 & \cellcolor[HTML]{E67C73}-196.3 & \cellcolor[HTML]{E88279}-197.0 & \cellcolor[HTML]{EA8D86}-198.5 & \cellcolor[HTML]{FFFFFF}-213.1   & \cellcolor[HTML]{FDF2F1}-211.3 & \cellcolor[HTML]{B5E1CC}-215.5 \\
                                             & F6 [402.8,\,439.6]                  & \cellcolor[HTML]{FFFFFF}433.7     & \cellcolor[HTML]{E67C73}439.6  & \cellcolor[HTML]{F4C2BE}436.5  & \cellcolor[HTML]{57BB8A}402.8  & \cellcolor[HTML]{6AC297}406.3    & \cellcolor[HTML]{89CFAD}412.2  & \cellcolor[HTML]{D0ECDE}425.1 & \cellcolor[HTML]{FCECEA}434.6  & \cellcolor[HTML]{FCECEA}434.6  \\
\multirow{-7}{*}{30}                        & F7 [964.8,\,978.9]                  & \cellcolor[HTML]{72C69D}965.7  & \cellcolor[HTML]{A6DBC1}967.5  & \cellcolor[HTML]{F2BBB6}975.0  & \cellcolor[HTML]{57BB8A}964.8  & \cellcolor[HTML]{CFEBDD}969.0  & \cellcolor[HTML]{FFFFFF}970.6     & \cellcolor[HTML]{E67C73}978.9 & \cellcolor[HTML]{E9877F}978.3  & \cellcolor[HTML]{F3BDB9}974.8 \\ \hline \hline 
                      & F1 [0.839,\,1.686] & \cellcolor[HTML]{FEF8F8}1.358   & \cellcolor[HTML]{E67C73}1.686   & \cellcolor[HTML]{B7E2CD}1.126   & \cellcolor[HTML]{FBE8E7}1.400     & \cellcolor[HTML]{FFFFFF}1.339        & \cellcolor[HTML]{57BB8A}0.839    & \cellcolor[HTML]{F4C3BF}1.499    & \cellcolor[HTML]{E0F2E9}1.248   & \cellcolor[HTML]{BDE4D1}1.144    \\
                      & F2 [47.65,\,58.89] & \cellcolor[HTML]{57BB8A}47.65  & \cellcolor[HTML]{5CBD8D}47.73  & \cellcolor[HTML]{DCF0E6}49.71  & \cellcolor[HTML]{F6FBF8}50.12 & \cellcolor[HTML]{FFFFFF}50.25       & \cellcolor[HTML]{FBE9E8}51.67   & \cellcolor[HTML]{E8857D}58.15   & \cellcolor[HTML]{E67C73}58.69  & \cellcolor[HTML]{E98880}57.93   \\
                      & F3 [6.876,\,12.42] & \cellcolor[HTML]{57BB8A}6.876   & \cellcolor[HTML]{86CEAA}7.469   & \cellcolor[HTML]{D5EEE1}8.467   & \cellcolor[HTML]{FFFFFF}8.995     & \cellcolor[HTML]{F4FAF7}8.858    & \cellcolor[HTML]{FDF2F1}9.344     & \cellcolor[HTML]{F1B2AD}11.02   & \cellcolor[HTML]{E98B83}12.03   & \cellcolor[HTML]{E67C73}12.42   \\
                      & F4 [0.749,\,1.097] & \cellcolor[HTML]{ABDDC5}0.819   & \cellcolor[HTML]{87CEAB}0.789    & \cellcolor[HTML]{57BB8A}0.749   & \cellcolor[HTML]{FFFFFF}0.887     & \cellcolor[HTML]{FEF9F8}0.898    & \cellcolor[HTML]{F5FBF8}0.879   & \cellcolor[HTML]{EEA69F}1.031   & \cellcolor[HTML]{ED9D96}1.045   & \cellcolor[HTML]{E67C73}1.097     \\
                      & F5 [-136.4,\,-97.26] & \cellcolor[HTML]{EA8E86}-98.78 & \cellcolor[HTML]{E67C73}-97.26 & \cellcolor[HTML]{EC9B95}-99.96 & \cellcolor[HTML]{FFFFFF}-108.7  & \cellcolor[HTML]{FDF2F1}-107.5 & \cellcolor[HTML]{A2D9BE}-123.9 & \cellcolor[HTML]{57BB8A}-136.4 & \cellcolor[HTML]{70C59B}-132.3 & \cellcolor[HTML]{77C8A0}-131.1 \\
                      & F6 [367.2,\,405.2] & \cellcolor[HTML]{7DCAA5}370.3  & \cellcolor[HTML]{ECF7F2}379.2  & \cellcolor[HTML]{E67C73}405.2 & \cellcolor[HTML]{FBE9E8}384.8 & \cellcolor[HTML]{B9E2CE}375.1  & \cellcolor[HTML]{F7D4D1}388.8  & \cellcolor[HTML]{57BB8A}367.2  & \cellcolor[HTML]{FCEDEC}384.1 & \cellcolor[HTML]{FFFFFF}380.7     \\
\multirow{-7}{*}{50} & F7 [1015,\,1068] & \cellcolor[HTML]{60BE90}1016     & \cellcolor[HTML]{CBEADB}1027   & \cellcolor[HTML]{F1B3AE}1053   & \cellcolor[HTML]{57BB8A}1015    & \cellcolor[HTML]{B5E1CB}1025    & \cellcolor[HTML]{FFFFFF}1033       & \cellcolor[HTML]{FDF0EF}1037    & \cellcolor[HTML]{F3BBB7}1051   & \cellcolor[HTML]{E67C73}1068   \\ \hline \hline 
                       & F1 [88.13,\,125.3] & \cellcolor[HTML]{F7D3D0}112.8 & \cellcolor[HTML]{F4FAF7}105.2 & \cellcolor[HTML]{92D3B3}94.59   & \cellcolor[HTML]{B1DFC9}97.99  & 106.3                         & \cellcolor[HTML]{57BB8A}88.13  & \cellcolor[HTML]{E67C73}125.3 & \cellcolor[HTML]{FAE4E3}110.3 & \cellcolor[HTML]{F8D9D6}112.0 \\
                       & F2 [123.6,\,147.5] & \cellcolor[HTML]{F9DBD9}140.6 & \cellcolor[HTML]{E3F3EB}135.7 & \cellcolor[HTML]{C2E6D4}132.8  & \cellcolor[HTML]{FFFFFF}138.0     & \cellcolor[HTML]{F7D5D2}141.1 & \cellcolor[HTML]{57BB8A}123.6 & \cellcolor[HTML]{E67C73}147.5 & \cellcolor[HTML]{95D4B5}129.0 & \cellcolor[HTML]{FEFAFA}138.4 \\
                       & F3 [12.05,\,15.11] & \cellcolor[HTML]{57BB8A}12.05                          & \cellcolor[HTML]{ABDDC5}12.77  & \cellcolor[HTML]{F6FBF9}13.41  & \cellcolor[HTML]{E4F4EC}13.25  & 13.48                           & \cellcolor[HTML]{F7D1CE}14.06  & \cellcolor[HTML]{EC9790}14.78  & \cellcolor[HTML]{EA8D86}14.90  & \cellcolor[HTML]{E67C73}15.11  \\
                       & F4 [6.517,\,18.74] & \cellcolor[HTML]{57BB8A}6.517                           & \cellcolor[HTML]{7CCAA4}7.434   & \cellcolor[HTML]{FAE2E0}12.47  & \cellcolor[HTML]{82CCA7}7.574   & \cellcolor[HTML]{A8DCC2}8.522   & \cellcolor[HTML]{FDF4F3}11.37  & \cellcolor[HTML]{FFFFFF}10.64     & \cellcolor[HTML]{F3BDB9}14.74  & \cellcolor[HTML]{E67C73}18.74  \\
                       & F5 [34.52,\,117.6] & \cellcolor[HTML]{B1DFC9}60.28  & \cellcolor[HTML]{E67C73}117.6 & \cellcolor[HTML]{FFFFFF}82.19     & \cellcolor[HTML]{F8D9D6}92.60   & \cellcolor[HTML]{F5C9C5}96.94  & \cellcolor[HTML]{F6D0CD}94.94  & \cellcolor[HTML]{57BB8A}34.52  & \cellcolor[HTML]{7BC9A3}44.79  & \cellcolor[HTML]{FEFEFE}82.18  \\
                       & F6 [332.7,\,363.0] & \cellcolor[HTML]{57BB8A}332.7                         & \cellcolor[HTML]{EBF7F1}343.6 & \cellcolor[HTML]{EB928B}360.0 & \cellcolor[HTML]{E8F5EF}343.4 & 345.0                         & \cellcolor[HTML]{F3BDB9}354.1 & \cellcolor[HTML]{64C093}333.7 & \cellcolor[HTML]{F6D0CD}351.5 & \cellcolor[HTML]{E67C73}363.0 \\
\multirow{-7}{*}{100} & F7 [1144,\,1193] & \cellcolor[HTML]{57BB8A}1144                           & \cellcolor[HTML]{B4E0CB}1162   & \cellcolor[HTML]{F2B9B4}1185   & \cellcolor[HTML]{B5E1CB}1162   & 1176                           & \cellcolor[HTML]{E67C73}1193   & \cellcolor[HTML]{A7DBC2}1160     & \cellcolor[HTML]{F4C2BE}1184   & \cellcolor[HTML]{E8857D}1192  \\ \hline \hline 

\end{tabular}
\egroup
\end{table}

\subsection{Tuning the Lipschitz and Local Optimization Conditions}
As LSADE allows controlling the addition of points for evaluation for the individual surrogates, we use it for tuning the balance between the exploration via the \textit{Lipschitz condition} and the exploitation via the \textit{Local Optimization condition} (from this point onward, the \textit{RBF condition} is always true). We start by using static rules for both conditions to be true, which will be based on the current iteration number. We consider 5 possibilities: 1 -- iteration number divisible by 1 (i.e., every iteration); 2 -- iteration number divisible by 2 (every other iteration); 4 -- iteration number divisible by 4; 8 -- iteration number divisible by 8; 0 -- never. For example, Li2$\,|\,$Lo0 means that points for real function evaluation based on the \textit{Lipschitz condition} are added every two iterations and the \textit{Local Optimization} is not used at all. In this setting, there were 25 variations in total. The results of the computations (mean over the 20 independent runs) for all 25 variations of the considered static rules for $D=30$ are reported in Table \ref{t:3}. In the table, next to the benchmark function identifier is the best and worst results in square brackets, where we chose to omit the rules with Li0 or Lo0 (as these were often quite a lot worse than the other ones). Also in Table \ref{t:3} are the aggregate results for $D=50$, while the detailed results can be found in the Appendix. These results suggest that using both the Lipschitz surrogate and the local optimization procedure is beneficial for every benchmark problem. The Lispchitz surrogate is especially well suited for problems F3 and F5-F7 (which are the ones with the complicated multimodal structure). However, none of the variations performed very well for all the considered problems, and the difference between the best and the worst variation for a given problem (even with disregarding rules with Li0 or Lo0) was quite high.

Since the Lipschitz surrogate should serve as an exploration-enhancing part of the algorithm, it is only natural that the frequency of its use should diminish as the iterations progress, to make space for the parts of the algorithm that focus on the exploitation of prospective areas. Hence, we devised several dynamic rules that decrease the frequency of using the Lipschitz surrogate, and increase the frequency of the local optimization, both in a linear manner. For instance, the variant {Li1-4$\,|\,$Lo8-1} starts with the Lipschitz surrogate being used every iteration and the local optimization procedure being used every 8 iterations, and ends with the Lipschitz surrogate being used every 4 iterations and the location optimization procedure being used every iteration. The individual conditions for the 9 considered variations can be found in the Appendix. The results of the computations with the dynamic rules for $D=[30,50,100]$ are summarized in Table \ref{t:4}. When comparing the results from the dynamic and the static rules, two important observations can be made. First, the dynamic rules have a much smaller interval between the best and the worst variation for the given problem instance, while the values of the best instances remain comparable. Second, there is one variation that stands out as having good results across many problem instances, particularly in higher dimensions. 

The Li1-4$\,|\,$Lo8-1 variant of the algorithm was selected as the best-performing one and will be used as the default variation for the subsequent modifications. It would probably be advantageous to devise a scheme that automatically decides on the frequency of using the Lipschitz surrogate or the local optimization procedure based on the past improvements and to tailor it for each problem separately. This is a research topic we plan to investigate in the future. 

\subsection{Comparison of Different RBFs}
Next, we investigate the effect of using different basis functions for the two RBF surrogate models (one global and one local). We use the Li1-4$\,|\,$Lo8-1 rule for the \textit{Lipschitz} and \textit{Local Optimization conditions} that was tuned for the multiquadratic (MQ) basis function and run the algorithm with cubic, thin plate spline (TPS), linear, and Gaussian basis function for the two RBFs instead. The results of the computations can be found in Table \ref{t:5}. From these results, it is apparent that the choice of the basis function has a substantial effect on the performance of the algorithm. Both the multiquadratic and the cubic basis functions performed very well on most of the problem instances, the TPS function was consistently mediocre, the Gaussian function performer mostly poorly (apart from the F1 problem) and the linear function performed the worst. The convergence histories of these variations can be found in the Appendix. Once again, it would very likely be beneficial to devise a scheme that would automatically choose the ``appropriate'' basis function for each problem separately. In the same vein, using different RBFs for the local and global models could also improve the performance of the algorithm.

\begin{table}[]
\centering
\caption{Comparison of different basis functions, $D = [30,50,100]$.} \label{t:5}
\bgroup
\def\arraystretch{1.4}
\setlength{\tabcolsep}{5.7pt}
\begin{tabular}{rl|rrrrr}
$D$                    & F  & MQ      & Cubic   & TPS     & Linear  & Gaussian \\ \hline \hline
                      & F1 & \cellcolor[HTML]{FAFCFB}0.011   & \cellcolor[HTML]{FFFFFF}0.011      & \cellcolor[HTML]{FEF5F4}0.509  & \cellcolor[HTML]{E67C73}6.276   & \cellcolor[HTML]{57BB8A}0.003   \\
                      & F2 & \cellcolor[HTML]{57BB8A}27.06   & \cellcolor[HTML]{C9E9D9}27.77   & \cellcolor[HTML]{FEF8F7}31.86  & \cellcolor[HTML]{E67C73}93.31   & \cellcolor[HTML]{FFFFFF}28.10      \\
                      & F3 & \cellcolor[HTML]{FFFFFF}1.308      & \cellcolor[HTML]{57BB8A}0.256   & \cellcolor[HTML]{70C59B}0.418  & \cellcolor[HTML]{E67C73}4.946   & \cellcolor[HTML]{EC9992}4.164   \\
                      & F4 & \cellcolor[HTML]{57BB8A}0.051   & \cellcolor[HTML]{7FCBA5}0.176   & \cellcolor[HTML]{FFFFFF}0.577     & \cellcolor[HTML]{E67C73}1.883   & \cellcolor[HTML]{F8DBD8}0.944   \\
                      & F5 & \cellcolor[HTML]{57BB8A}-218.7 & \cellcolor[HTML]{D2ECDF}-172.6 & \cellcolor[HTML]{FFFFFF}-155.9   & \cellcolor[HTML]{FDF5F4}-143.6& \cellcolor[HTML]{E67C73}-9.30   \\
                      & F6 & \cellcolor[HTML]{C9E9D9}433.7  & \cellcolor[HTML]{57BB8A}426.2  & \cellcolor[HTML]{FFFFFF}437.2    & \cellcolor[HTML]{FCEFEE}448.1  & \cellcolor[HTML]{E67C73}526.5  \\
\multirow{-7}{*}{30}  & F7 & \cellcolor[HTML]{E77D74}965.7   & \cellcolor[HTML]{57BB8A}938.8   & \cellcolor[HTML]{9FD8BC}944.4  & \cellcolor[HTML]{E67C73}965.8   & \cellcolor[HTML]{FFFFFF}951.7      \\  \hline \hline 
                      & F1 & \cellcolor[HTML]{FFFFFF}1.358      & \cellcolor[HTML]{79C9A2}0.434   & \cellcolor[HTML]{FDF0EF}7.556  & \cellcolor[HTML]{E67C73}54.78  & \cellcolor[HTML]{57BB8A}0.191   \\
                      & F2 & \cellcolor[HTML]{57BB8A}47.65   & \cellcolor[HTML]{FFFFFF}47.98      & \cellcolor[HTML]{FDF5F4}62.20  & \cellcolor[HTML]{E67C73}221.8  & \cellcolor[HTML]{73C69E}47.71   \\
                      & F3 & \cellcolor[HTML]{F8D6D3}6.876   & \cellcolor[HTML]{57BB8A}0.695   & \cellcolor[HTML]{81CCA7}1.822  & \cellcolor[HTML]{E67C73}10.56  & \cellcolor[HTML]{FFFFFF}5.161      \\
                      & F4 & \cellcolor[HTML]{FFFFFF}0.819      & \cellcolor[HTML]{57BB8A}0.380   & \cellcolor[HTML]{F8FCFA}0.801  & \cellcolor[HTML]{E67C73}5.668   & \cellcolor[HTML]{FFFCFC}0.930   \\
                      & F5 & \cellcolor[HTML]{57BB8A}-98.78  & \cellcolor[HTML]{EAF6F0}-10.03  & \cellcolor[HTML]{FFFFFF}2.45      & \cellcolor[HTML]{F8D9D6}82.64   & \cellcolor[HTML]{E67C73}274.9  \\
                      & F6 & \cellcolor[HTML]{57BB8A}370.3  & \cellcolor[HTML]{FFFFFF}481.6     & \cellcolor[HTML]{E5F4ED}464.5 & \cellcolor[HTML]{F6CDCA}521.6  & \cellcolor[HTML]{E67C73}585.6  \\ 
\multirow{-7}{*}{50}  & F7 & \cellcolor[HTML]{F5C6C2}1016  & \cellcolor[HTML]{57BB8A}976.3   & \cellcolor[HTML]{91D2B2}979.6  & \cellcolor[HTML]{E67C73}1054 & \cellcolor[HTML]{FFFFFF}985.7      \\ \hline \hline 
                      & F1 & \cellcolor[HTML]{FFFFFF}112.8     & \cellcolor[HTML]{6AC297}30.94   & \cellcolor[HTML]{F9DEDC}279.5 & \cellcolor[HTML]{E67C73}766.6  & \cellcolor[HTML]{57BB8A}20.39   \\
                      & F2 & \cellcolor[HTML]{B8E2CE}140.6  & \cellcolor[HTML]{57BB8A}106.4  & \cellcolor[HTML]{F8D8D5}331.7 & \cellcolor[HTML]{E67C73}714.5  & \cellcolor[HTML]{FFFFFF}165.1     \\
                      & F3 & \cellcolor[HTML]{F6CCC9}12.05  & \cellcolor[HTML]{57BB8A}4.622   & \cellcolor[HTML]{FFFFFF}9.089     & \cellcolor[HTML]{E67C73}16.65  & \cellcolor[HTML]{FAFDFB}8.965   \\
                      & F4 & \cellcolor[HTML]{FEF7F7}6.517   & \cellcolor[HTML]{57BB8A}0.816   & \cellcolor[HTML]{FFFFFF}2.190     & \cellcolor[HTML]{E67C73}69.61  & \cellcolor[HTML]{66C195}0.946   \\
                      & F5 & \cellcolor[HTML]{57BB8A}60.28   & \cellcolor[HTML]{FFFFFF}646.8     & \cellcolor[HTML]{DCF1E7}527.1 & \cellcolor[HTML]{FCECEB}701.2  & \cellcolor[HTML]{E67C73}1012 \\
                      & F6 & \cellcolor[HTML]{57BB8A}332.7  & \cellcolor[HTML]{FFFFFF}550.4     & \cellcolor[HTML]{E9F6F0}522.9 & \cellcolor[HTML]{F3C0BC}572.5  & \cellcolor[HTML]{E67C73}596.3  \\
\multirow{-7}{*}{100} & F7 & \cellcolor[HTML]{FFFFFF}1144     & \cellcolor[HTML]{57BB8A}1056  & \cellcolor[HTML]{FFFDFD}1146 & \cellcolor[HTML]{E67C73}1248  & \cellcolor[HTML]{C1E5D3}1112 \\
\hline \hline 
\end{tabular}
\egroup
\end{table}

\begin{table}[]
\centering
\caption{Comparison with other algorithms, average objective function value.} \label{t:6}
\bgroup
\def\arraystretch{1.4}
\setlength{\tabcolsep}{2.7pt}
\begin{tabular}{rl|rrrrrr|rr}
$D$                     & F  & \,\,\,\,\,SAMSO\,\,\,\,\,                          & MGP-SLPSO                       & \,\,\,\,\,\,\,\,GSGA\,\,\,\,\,\,\,\,                           & \,\,\,\,\,SAGWO \,\,\,\,\,                          & \,\,\,\,\,\,\,\,ESAO\,\,\,\,\,\,\,\,                          & \,\,SA-COSO\,\,                        & \,LSADE-MQ\,                     & \,\,LSADE-C\,\,                  \\ \hline \hline 
                      & F1 & \cellcolor[HTML]{A5DAC0}0.0053 & \cellcolor[HTML]{57BB8A}0       & \cellcolor[HTML]{FFFDFD}0.073  & \cellcolor[HTML]{57BB8A}0.00007 & \cellcolor[HTML]{FFFFFF}0.027    & \cellcolor[HTML]{E67C73}3.85   & \cellcolor[HTML]{FDFEFD}0.0113 & \cellcolor[HTML]{FFFFFF}0.0115    \\
                      & F2 & \cellcolor[HTML]{FFFFFF}28.3      & \cellcolor[HTML]{E67C73}100     & \cellcolor[HTML]{E6F5ED}27.60  & \cellcolor[HTML]{FFFFFF}28.30      & \cellcolor[HTML]{57BB8A}25.04 & \cellcolor[HTML]{F4C5C2}59.9   & \cellcolor[HTML]{C8E8D8}27.06  & \cellcolor[HTML]{F0F9F4}27.77  \\
                      & F3 & \cellcolor[HTML]{C3E7D5}0.628  & \cellcolor[HTML]{E67C73}6.58    & \cellcolor[HTML]{5BBC8C}0.023  & \cellcolor[HTML]{57BB8A}0       & \cellcolor[HTML]{F9DBD9}2.521 & \cellcolor[HTML]{EDA19B}5.01   & \cellcolor[HTML]{FEF8F7}1.308  & \cellcolor[HTML]{83CCA8}0.256  \\
                      & F4 & \cellcolor[HTML]{F9DCDA}0.538  & \cellcolor[HTML]{57BB8A}0.013   & \cellcolor[HTML]{FFFDFD}0.228  & \cellcolor[HTML]{58BB8B}0.015   & \cellcolor[HTML]{F0B0AB}0.953 & \cellcolor[HTML]{E67C73}1.44   & \cellcolor[HTML]{78C8A1}0.051  & \cellcolor[HTML]{E7F5EE}0.176  \\
                      & F5 & \cellcolor[HTML]{57BB8A}-239   & \cellcolor[HTML]{95D4B5}-220    & \cellcolor[HTML]{CDEADC}-203.0 & \cellcolor[HTML]{F8D8D5}-128.8  & \cellcolor[HTML]{E67C73}6.325 & \cellcolor[HTML]{EFA8A1}-57.4  & \cellcolor[HTML]{99D5B8}-218.7 & \cellcolor[HTML]{FEF5F5}-172.6 \\
                      & F6 & \cellcolor[HTML]{57BB8A}372    & \cellcolor[HTML]{EFEFEF}N/A      & \cellcolor[HTML]{EFF8F4}424.7  & \cellcolor[HTML]{F0B0AA}489.8   & \cellcolor[HTML]{F3F3F3}N/A    & \cellcolor[HTML]{E67C73}528    & \cellcolor[HTML]{FFFAFA}433.7  & \cellcolor[HTML]{F4FAF7}426.2  \\
\multirow{-7}{*}{30}  & F7 & \cellcolor[HTML]{57BB8A}922    & \cellcolor[HTML]{FAE0DE}952     & \cellcolor[HTML]{7CCAA4}927.2  & \cellcolor[HTML]{E67C73}973.2   & \cellcolor[HTML]{9BD6BA}931.6 & \cellcolor[HTML]{EA9089}969    & \cellcolor[HTML]{EDA099}965.7  & \cellcolor[HTML]{CFEBDD}938.8  \\ \hline \hline
                      & F1 & \cellcolor[HTML]{EEF8F3}0.513  & \cellcolor[HTML]{57BB8A}0       & \cellcolor[HTML]{FFFFFF}0.621     & \cellcolor[HTML]{58BB8A}0.004   & \cellcolor[HTML]{FFFFFF}0.740    & \cellcolor[HTML]{E67C73}46.6 & \cellcolor[HTML]{FFFDFD}1.358  & \cellcolor[HTML]{D7EFE3}0.434  \\
                      & F2 & \cellcolor[HTML]{FFFFFE}50.1   & \cellcolor[HTML]{F7D2CF}120     & \cellcolor[HTML]{C6E7D7}48.21  & \cellcolor[HTML]{FFFFFF}49.06      & \cellcolor[HTML]{57BB8A}47.39 & \cellcolor[HTML]{E67C73}253    & \cellcolor[HTML]{79C9A2}47.65  & \cellcolor[HTML]{A6DBC1}47.98  \\
                      & F3 & \cellcolor[HTML]{FFFFFF}1.53      & \cellcolor[HTML]{E67C73}9.31    & \cellcolor[HTML]{59BB8B}0.022  & \cellcolor[HTML]{57BB8A}0       & \cellcolor[HTML]{F9FCFB}1.431 & \cellcolor[HTML]{E8847C}8.86   & \cellcolor[HTML]{EEA59F}6.876  & \cellcolor[HTML]{A5DAC0}0.695  \\
                      & F4 & \cellcolor[HTML]{FFFCFC}0.666  & \cellcolor[HTML]{82CCA8}0.154   & \cellcolor[HTML]{C3E6D5}0.346  & \cellcolor[HTML]{57BB8A}0.025   & \cellcolor[HTML]{FDF5F4}0.94  & \cellcolor[HTML]{E67C73}5.63   & \cellcolor[HTML]{FEF8F7}0.819  & \cellcolor[HTML]{CEEBDD}0.380  \\
                      & F5 & \cellcolor[HTML]{57BB8A}-169   & \cellcolor[HTML]{FDF3F2}33      & \cellcolor[HTML]{ADDEC6}-75.82 & \cellcolor[HTML]{F6CDC9}98.39   & \cellcolor[HTML]{EB928A}198.6 & \cellcolor[HTML]{E67C73}235    & \cellcolor[HTML]{98D5B7}-98.78 & \cellcolor[HTML]{EAF6F1}-10.03 \\
                      & F6 & \cellcolor[HTML]{57BB8A}326    & \cellcolor[HTML]{F3F3F3}N/A      & \cellcolor[HTML]{C6E8D7}403.3  & \cellcolor[HTML]{F7D2CF}502.0   & \cellcolor[HTML]{F3F3F3}N/A    & \cellcolor[HTML]{E67C73}613    & \cellcolor[HTML]{96D4B6}370.3  & \cellcolor[HTML]{FAE1DF}481.6  \\
\multirow{-7}{*}{50}  & F7 & \cellcolor[HTML]{57BB8A}970    & \cellcolor[HTML]{EC9C95}1060    & \cellcolor[HTML]{5BBC8D}970.7  & \cellcolor[HTML]{F1B5AF}1044.1  & \cellcolor[HTML]{79C8A1}975.3 & \cellcolor[HTML]{E67C73}1080   & \cellcolor[HTML]{FAE0DE}1016   & \cellcolor[HTML]{7FCBA6}976.3  \\ \hline \hline
                      & F1 & \cellcolor[HTML]{FFFDFD}72.1   & \cellcolor[HTML]{57BB8A}0.00005 & \cellcolor[HTML]{7FCBA5}12.33  & \cellcolor[HTML]{57BB8A}0.139   & \cellcolor[HTML]{E67C73}1283  & \cellcolor[HTML]{ED9C95}985    & \cellcolor[HTML]{FEF9F9}112.8  & \cellcolor[HTML]{BBE3D0}30.94  \\
                      & F2 & \cellcolor[HTML]{FFFBFB}286    & \cellcolor[HTML]{FBE9E7}612     & \cellcolor[HTML]{5BBC8C}109.1  & \cellcolor[HTML]{71C59C}123.4   & \cellcolor[HTML]{FCEBE9}578.8 & \cellcolor[HTML]{E67C73}2500   & \cellcolor[HTML]{8CD0AF}140.6  & \cellcolor[HTML]{57BB8A}106.4  \\
                      & F3 & \cellcolor[HTML]{D3EDE0}6.12   & \cellcolor[HTML]{EC9891}14.3    & \cellcolor[HTML]{71C59C}1.31   & \cellcolor[HTML]{57BB8A}0       & \cellcolor[HTML]{F9DBD9}10.36 & \cellcolor[HTML]{E67C73}15.9   & \cellcolor[HTML]{F3BEBA}12.05  & \cellcolor[HTML]{B5E1CB}4.622  \\
                      & F4 & \cellcolor[HTML]{FFFFFF}1.06      & \cellcolor[HTML]{D6EEE2}0.715   & \cellcolor[HTML]{D4EDE1}0.706  & \cellcolor[HTML]{57BB8A}0.023   & \cellcolor[HTML]{E98981}57.34 & \cellcolor[HTML]{E67C73}63.5   & \cellcolor[HTML]{FDF4F3}6.517  & \cellcolor[HTML]{E8F5EF}0.816  \\
                      & F5 & \cellcolor[HTML]{FFFDFD}737    & \cellcolor[HTML]{FAE1DF}885     & \cellcolor[HTML]{F1F9F5}672.5  & \cellcolor[HTML]{FDF1F0}800.1   & \cellcolor[HTML]{FCFDFC}713.4 & \cellcolor[HTML]{E67C73}1420   & \cellcolor[HTML]{57BB8A}60.28  & \cellcolor[HTML]{EBF6F1}646.8  \\
                      & F6 & \cellcolor[HTML]{FCFDFD}513    & \cellcolor[HTML]{F3F3F3}N/A      & \cellcolor[HTML]{C0E5D3}447.2  & \cellcolor[HTML]{FFFEFE}518.6   & \cellcolor[HTML]{F3F3F3}N/A    & \cellcolor[HTML]{E67C73}807    & \cellcolor[HTML]{57BB8A}332.7  & \cellcolor[HTML]{FDF0EF}550.4  \\
\multirow{-7}{*}{100} & F7 & \cellcolor[HTML]{EBF7F1}1290   & \cellcolor[HTML]{EC9A93}1390    & \cellcolor[HTML]{D6EEE2}1256   & \cellcolor[HTML]{F7D4D1}1350    & \cellcolor[HTML]{F1B4AF}1372  & \cellcolor[HTML]{E67C73}1410   & \cellcolor[HTML]{8FD1B1}1144   & \cellcolor[HTML]{57BB8A}1056   \\ \hline \hline
                      & F1 & \cellcolor[HTML]{7DCAA4}1520   & \cellcolor[HTML]{EFEFEF}N/A      & \cellcolor[HTML]{EFEFEF}N/A     & \cellcolor[HTML]{EFEFEF}N/A      & \cellcolor[HTML]{E67C73}17616 & \cellcolor[HTML]{E98880}16382  & \cellcolor[HTML]{FFFFFF}3959      & \cellcolor[HTML]{57BB8A}793.6  \\
                      & F2 & \cellcolor[HTML]{FFFFFF}1150      & \cellcolor[HTML]{EFEFEF}N/A      & \cellcolor[HTML]{EFEFEF}N/A     & \cellcolor[HTML]{EFEFEF}N/A      & \cellcolor[HTML]{FAE4E2}4318  & \cellcolor[HTML]{E67C73}16411  & \cellcolor[HTML]{BDE4D1}927.2  & \cellcolor[HTML]{57BB8A}576.3  \\
                      & F3 & \cellcolor[HTML]{57BB8A}12     & \cellcolor[HTML]{EFEFEF}N/A      & \cellcolor[HTML]{EFEFEF}N/A     & \cellcolor[HTML]{EFEFEF}N/A      & \cellcolor[HTML]{FFFFFF}14.69    & \cellcolor[HTML]{E67C73}17.86  & \cellcolor[HTML]{FBEAE9}15.20  & \cellcolor[HTML]{F8FCFA}14.58  \\
                      & F4 & \cellcolor[HTML]{5EBE8F}9.03   & \cellcolor[HTML]{EFEFEF}N/A      & \cellcolor[HTML]{EFEFEF}N/A     & \cellcolor[HTML]{EFEFEF}N/A      & \cellcolor[HTML]{E77E75}572.9 & \cellcolor[HTML]{E67C73}577.7  & \cellcolor[HTML]{FFFFFF}135.6     & \cellcolor[HTML]{57BB8A}2.892  \\
                      & F5 & \cellcolor[HTML]{EEA39D}4960   & \cellcolor[HTML]{EFEFEF}N/A      & \cellcolor[HTML]{EFEFEF}N/A     & \cellcolor[HTML]{EFEFEF}N/A      & \cellcolor[HTML]{E67C73}5389  & \cellcolor[HTML]{FFFFFF}3927      & \cellcolor[HTML]{57BB8A}1416   & \cellcolor[HTML]{92D3B3}2305   \\
                      & F6 & \cellcolor[HTML]{FFFFFF}684       & \cellcolor[HTML]{EFEFEF}N/A      & \cellcolor[HTML]{EFEFEF}N/A     & \cellcolor[HTML]{EFEFEF}N/A      & \cellcolor[HTML]{F3F3F3}N/A    & \cellcolor[HTML]{F3F3F3}N/A     & \cellcolor[HTML]{57BB8A}578.7  & \cellcolor[HTML]{E67C73}722.7  \\
\multirow{-7}{*}{200} & F7 & \cellcolor[HTML]{FFFFFF}1340      & \cellcolor[HTML]{EFEFEF}N/A      & \cellcolor[HTML]{EFEFEF}N/A     & \cellcolor[HTML]{EFEFEF}N/A      & \cellcolor[HTML]{E67C73}1456  & \cellcolor[HTML]{FEF8F7}1347   & \cellcolor[HTML]{A3DABF}1276   & \cellcolor[HTML]{57BB8A}1222  \\ \hline \hline
\end{tabular}
\egroup
\end{table}

\subsection{Comparison with Other Algorithms}
The proposed LSADE method is compared with six SAEAs, namely, SA-COSO \cite{SACOSO17}, ESAO \cite{ESAO19}, SAGWO \cite{Dong20}, GSGA \cite{Cai20}, MGP-SLPSO \cite{Tian19}, and SAMSO \cite{Li21}, which are all methods for high-dimensional expensive problems that can be compared on the same testbed (although some of the problems have not been evaluated by some of the algorithms). SA-COSO is a surrogate-assisted cooperative swarm optimization algorithm, in which a surrogate-assisted particle swarm optimization algorithm and a surrogate-assisted social learning based particle swarm optimization algorithm cooperatively
search for the global optimum. ESAO is an evolutionary sampling-assisted optimization method that combines global and local search to balance exploration and exploitation, and employs DE as the optimization method. SAGWO utilizes the grey wolf optimization algorithm and conducts the search in three phases, initial exploration, RBF-assisted meta-heuristic exploration, and knowledge mining on RBF. GSGA uses a surrogate-based trust region local search method, a surrogate-guided GA updating mechanism with a neighbor region partition strategy, and a prescreening strategy based on the expected improvement infilling criterion of a simplified
Kriging in the optimization process. MGP-SLPSO employs a multi-objective
infill criterion that considers the approximated fitness and the approximation uncertainty as two objectives for a Gaussian process assisted social learning particle swarm optimization algorithm. SAMSO is a a surrogate-assisted multiswarm optimization algorithm for high-dimensional problems, which includes two swarms: the first one uses the learner phase of teaching-learning-based optimization to enhance exploration and the second one uses the particle swarm
optimization for faster convergence. The data for the comparison were obtained from the corresponding papers, with the exception of the data for SA-COSO and ESAO, which were obtained from \cite{Li21}. 

The average objective function value for the considered algorithms and for the LSADE algorithm with multiquadratic and cubic RBFs are reported in Table \ref{t:6}. More detailed results, including the best results, worst results, and standard deviations of the independent runs for all the considered algorithms can be found in the Appendix. Looking at $D=30$ first, we can see that there is no one algorithm that is strictly better than all the others on all the benchmark functions. The less complicated functions F1-F4 are dominated by MGP-SLPSO, GSGA, SAGWO, and EASO, while for the more complicated functions F5-F7 SAMSO seems to be the best. Both of the LSADE variants come out somewhere in the middle for all problems. In a direct comparison with LSADE, the best ones are SAMSO (better in 5/7 than LSADE-MQ) and GSGA (better in 5/7 than LSADE-C). For $D=50$ the situation is quite similar: the best algorithms for the less complicated problems are MGP-SLPSO, SAGWO, and ESAO, while SAMSO dominates the more complicated problems again. Both of the LSADE variants are, once again, somewhere in the middle. In a direct comparison with LSADE, the SAMSO is the best (better in 6/7 than LSADE-MQ). However, the situation changes substantially for higher dimensions. For $D=100$, MGP-SLPSO, LSADE-C, and SAGWO dominate the less complicated functions, while LSADE-MQ and LSADE-C have the best results for the more complicated function. In direct comparison with LSADE, the best ones are GSGA and SAGWO (both 4/7 for both variants). For $D=200$, only three of the six considered algorithms reported results (possibly because of prohibitively large computational times as will be investigated in the following section). In these largest instances, LSADE-MQ and LSADE-C were the best choices for all problems with the exception of F3 for which SAMSO was the best. 

\begin{figure*}[ht!]
\centering
\setlength{\tabcolsep}{5pt}
\bgroup
\def\arraystretch{5}
\begin{tabular}{ccc}
\includegraphics[height = 0.28\textwidth]{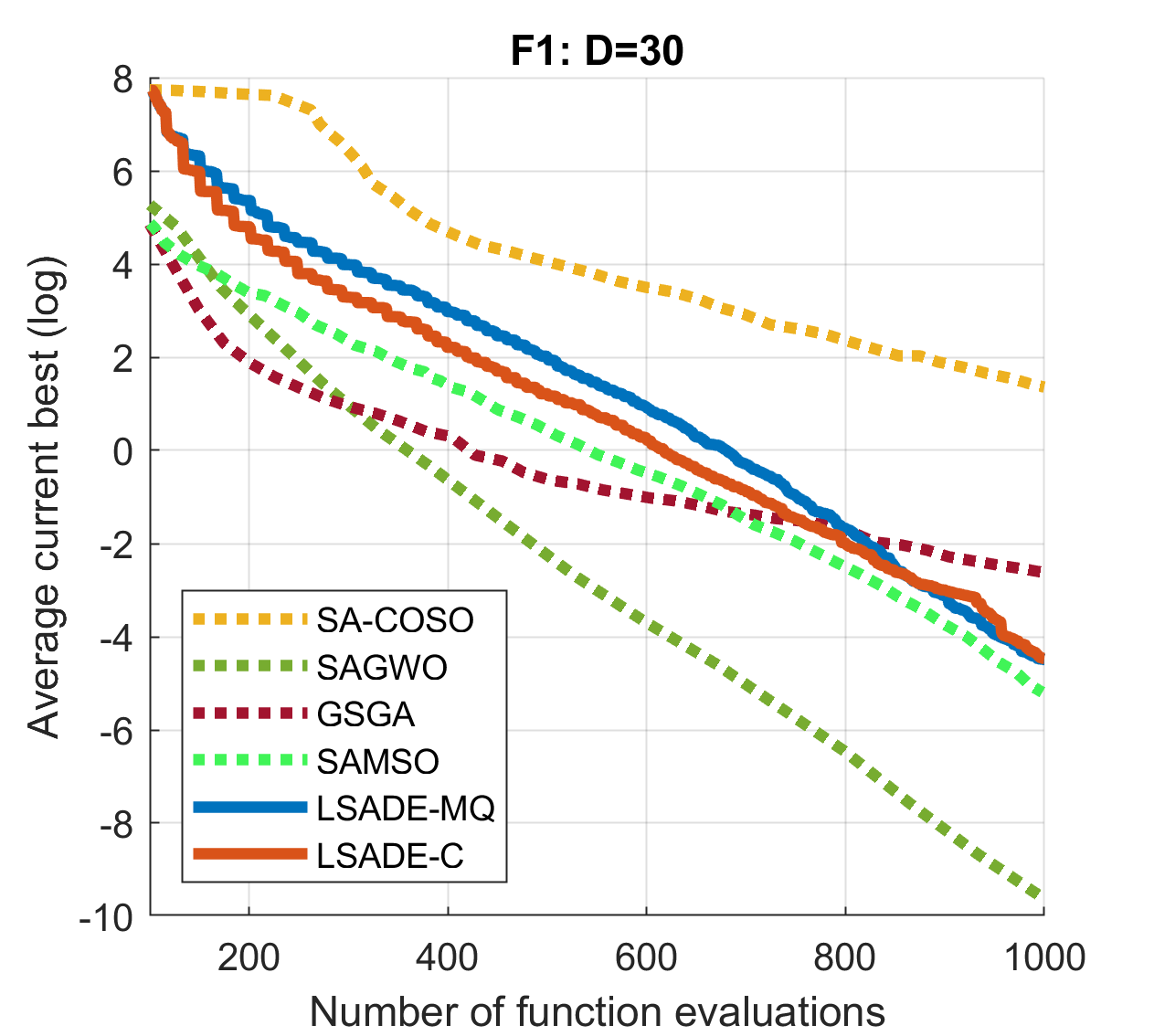} & \includegraphics[height = 0.28\textwidth]{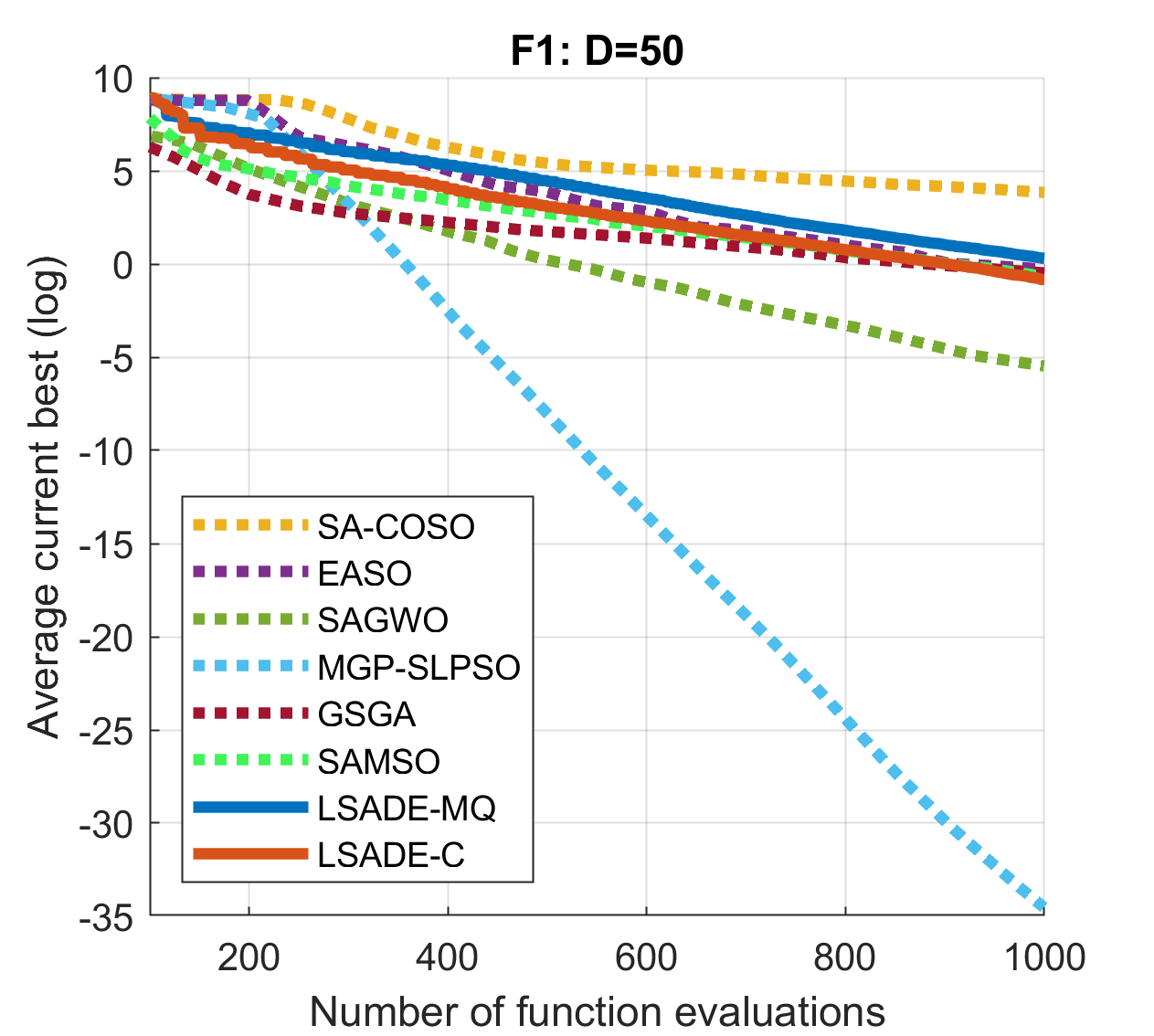} & \includegraphics[height = 0.28\textwidth]{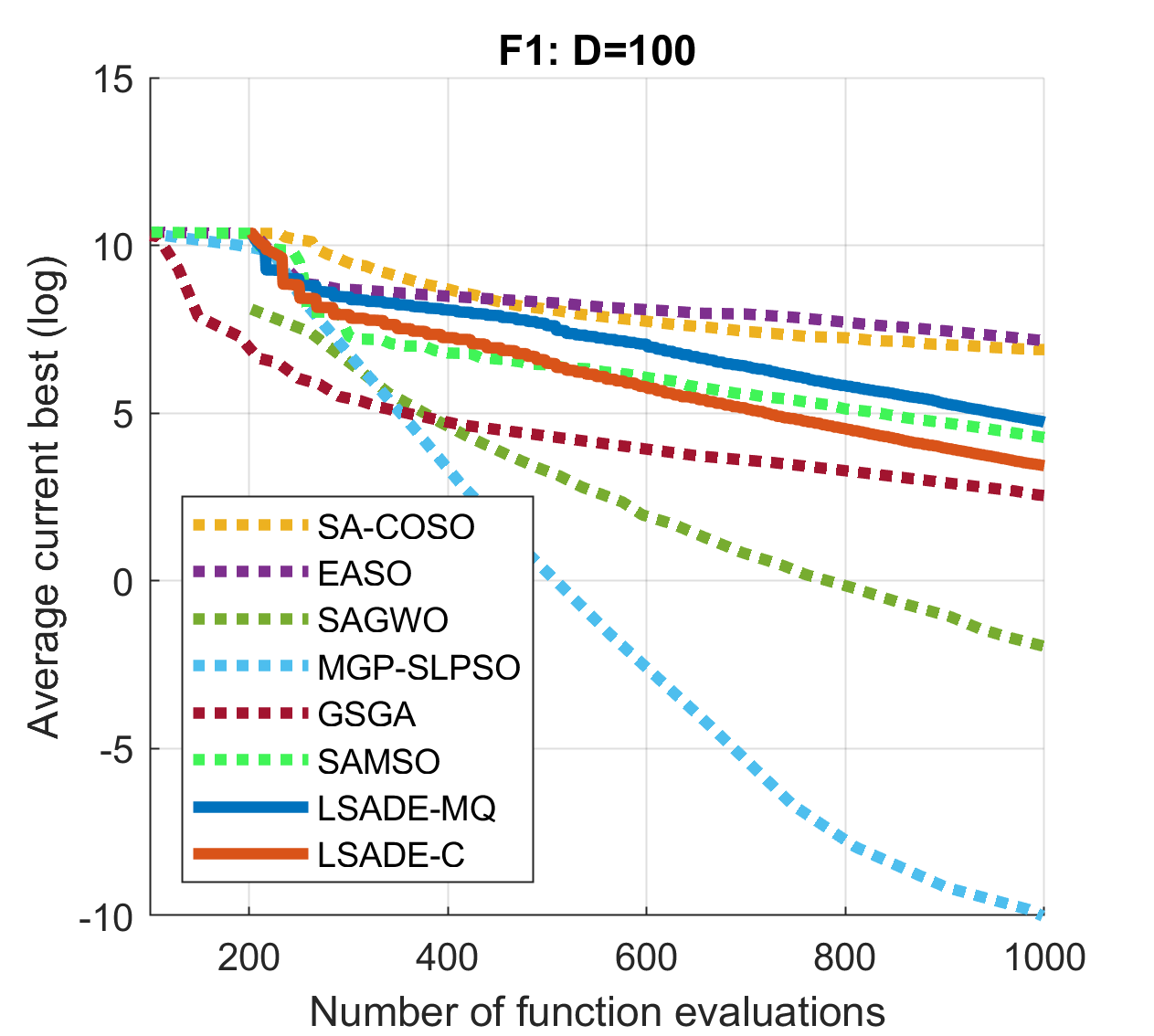} \\
\includegraphics[height = 0.28\textwidth]{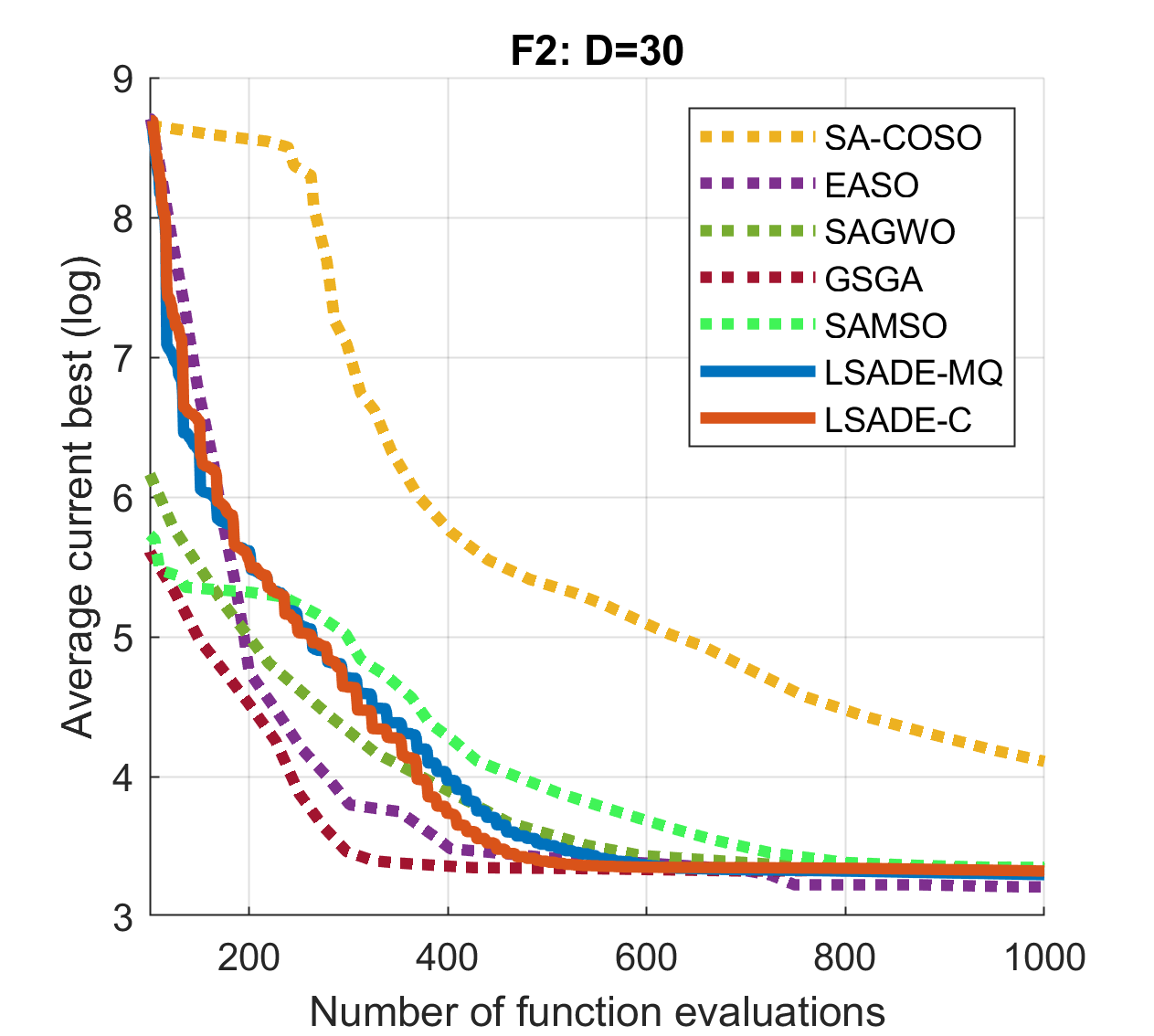} & \includegraphics[height = 0.28\textwidth]{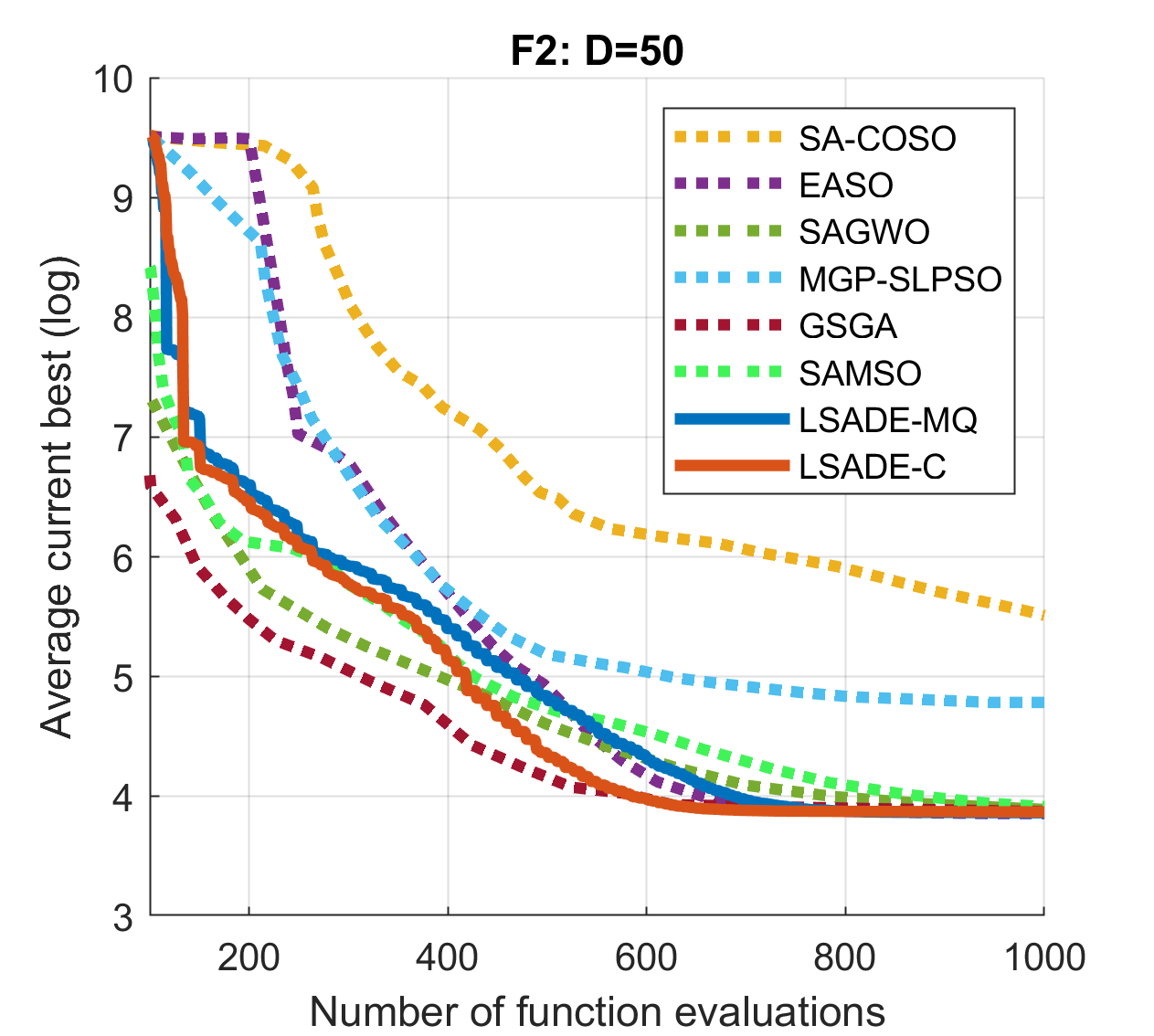} & \includegraphics[height = 0.28\textwidth]{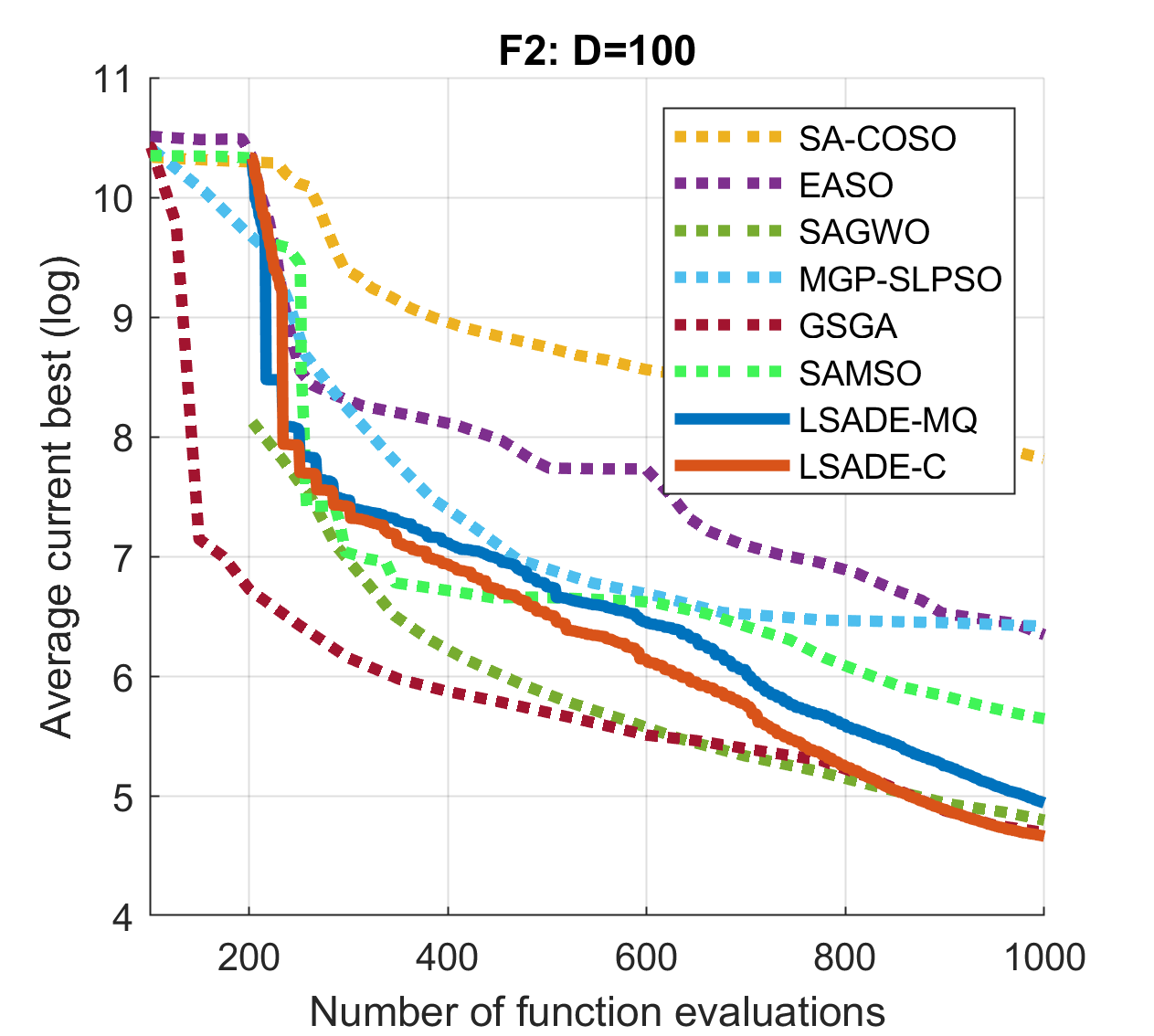} \\
\includegraphics[height = 0.28\textwidth]{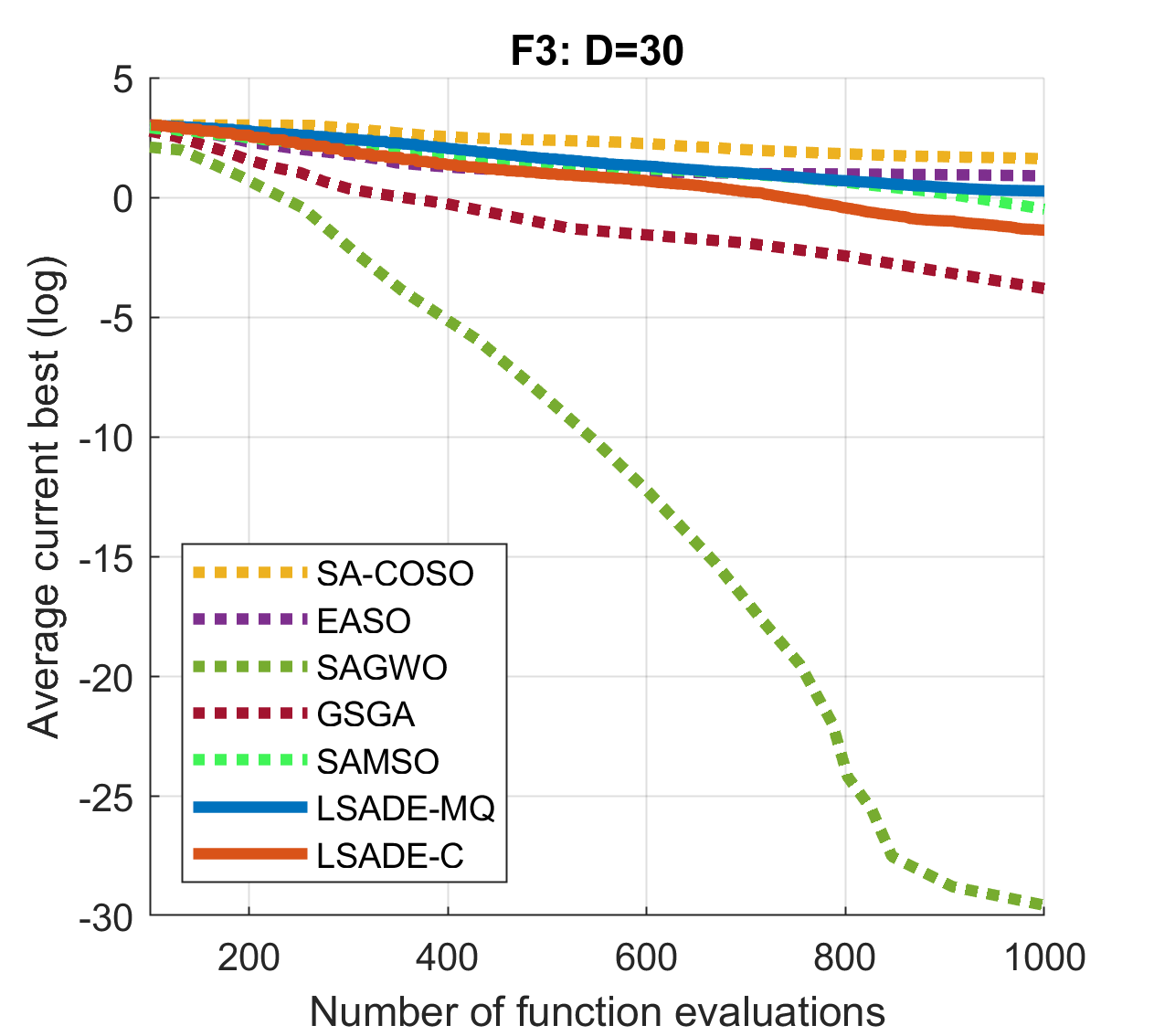} & \includegraphics[height = 0.28\textwidth]{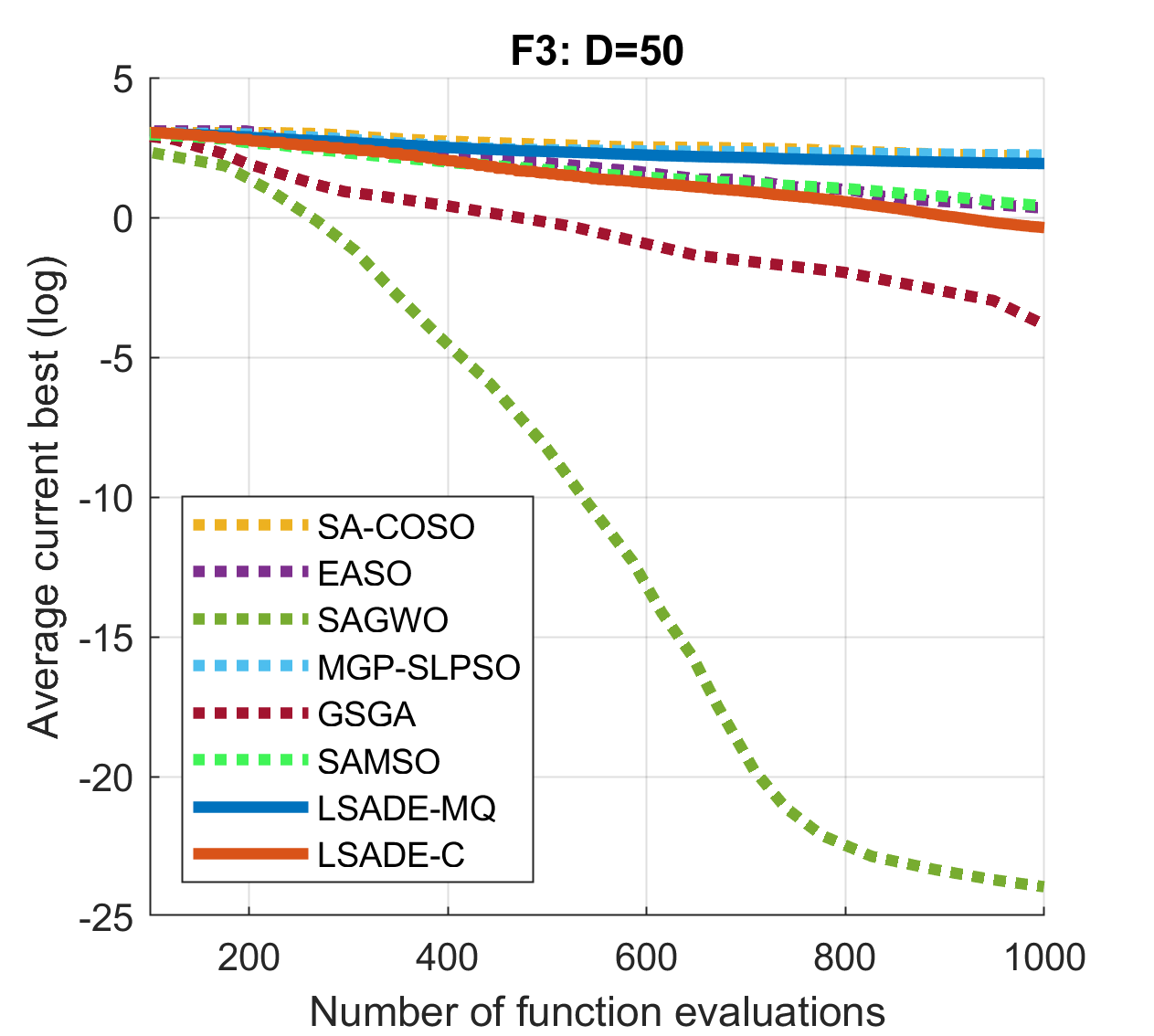} & \includegraphics[height = 0.28\textwidth]{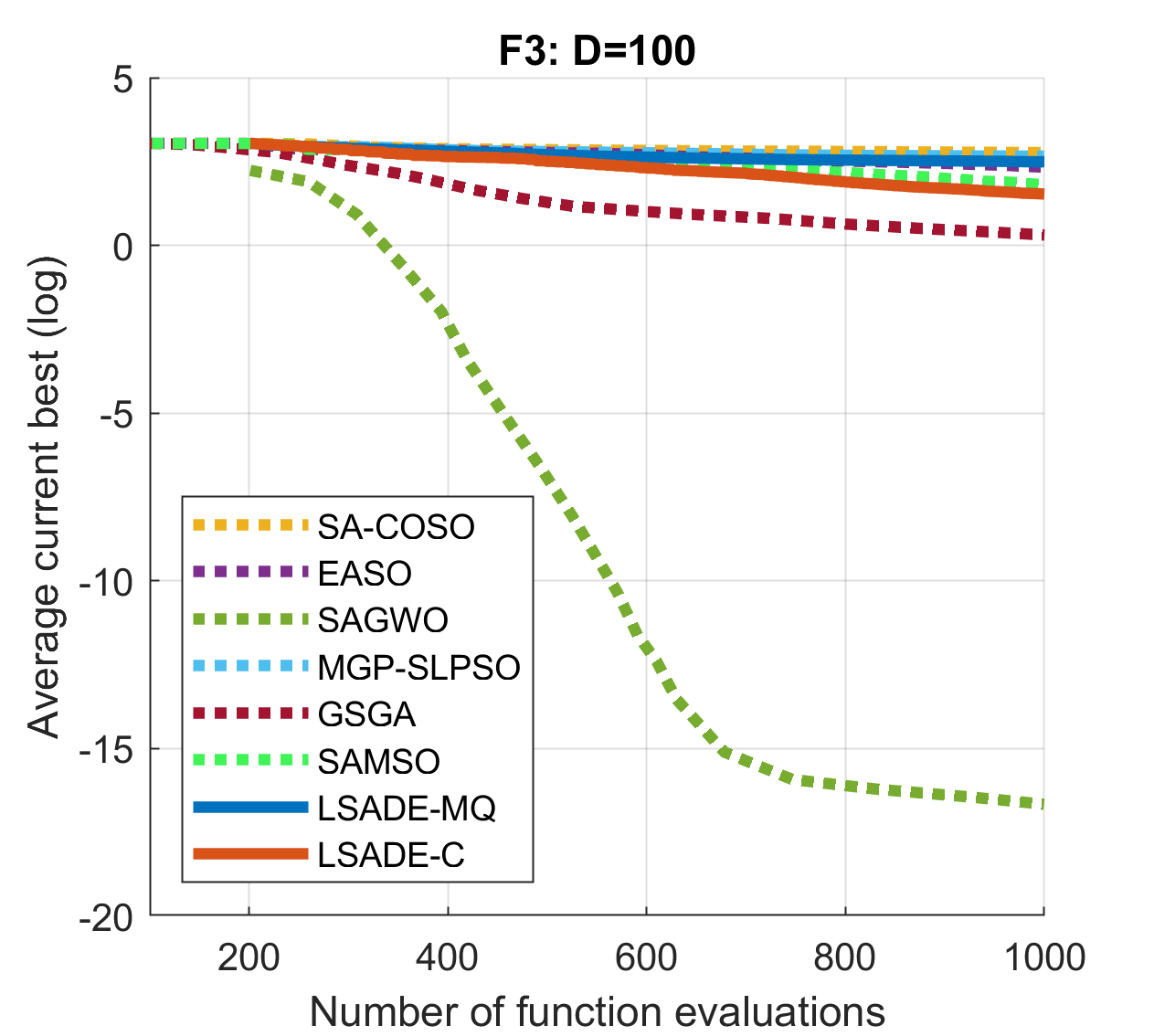} \\
\includegraphics[height = 0.28\textwidth]{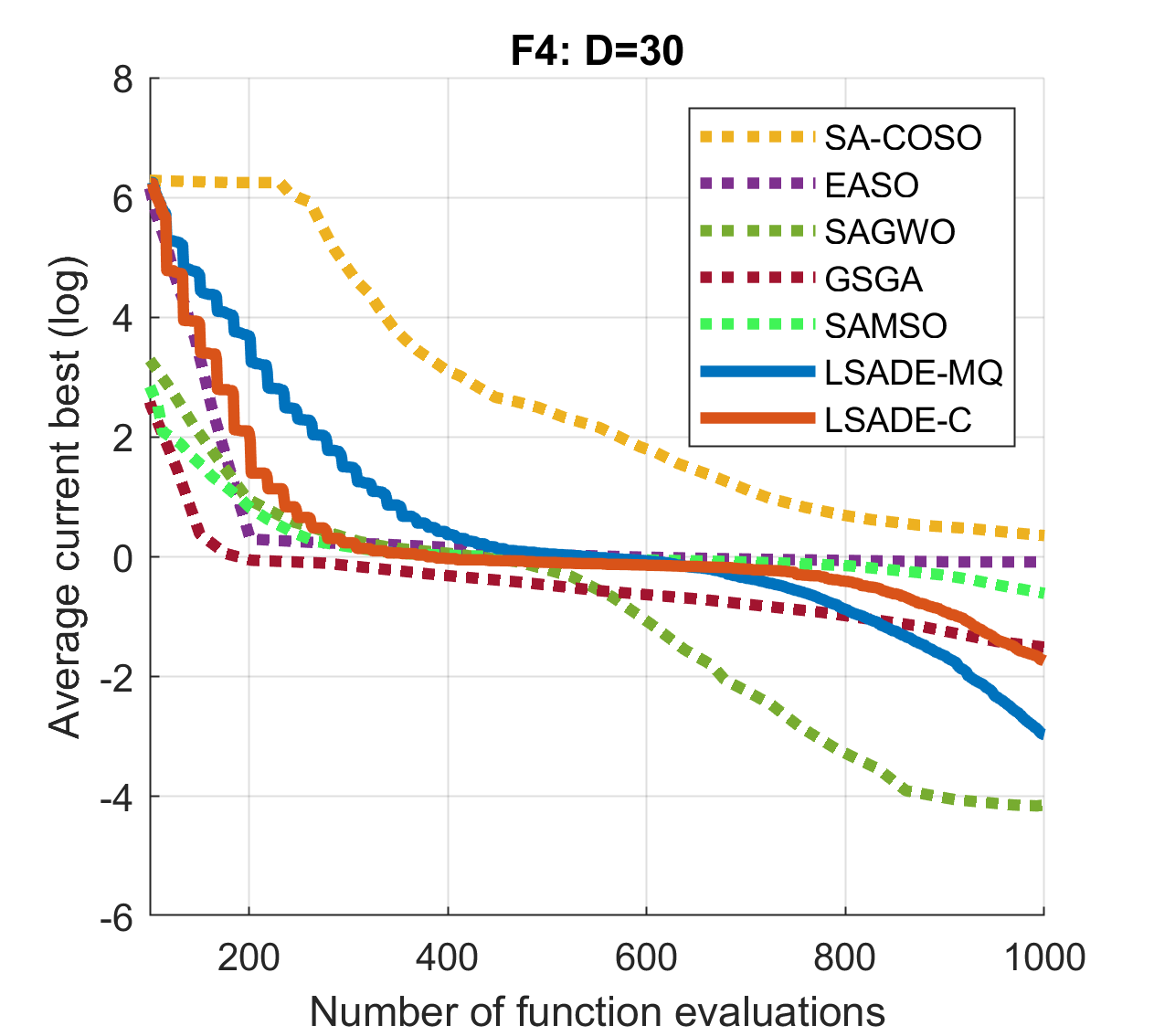} & \includegraphics[height = 0.28\textwidth]{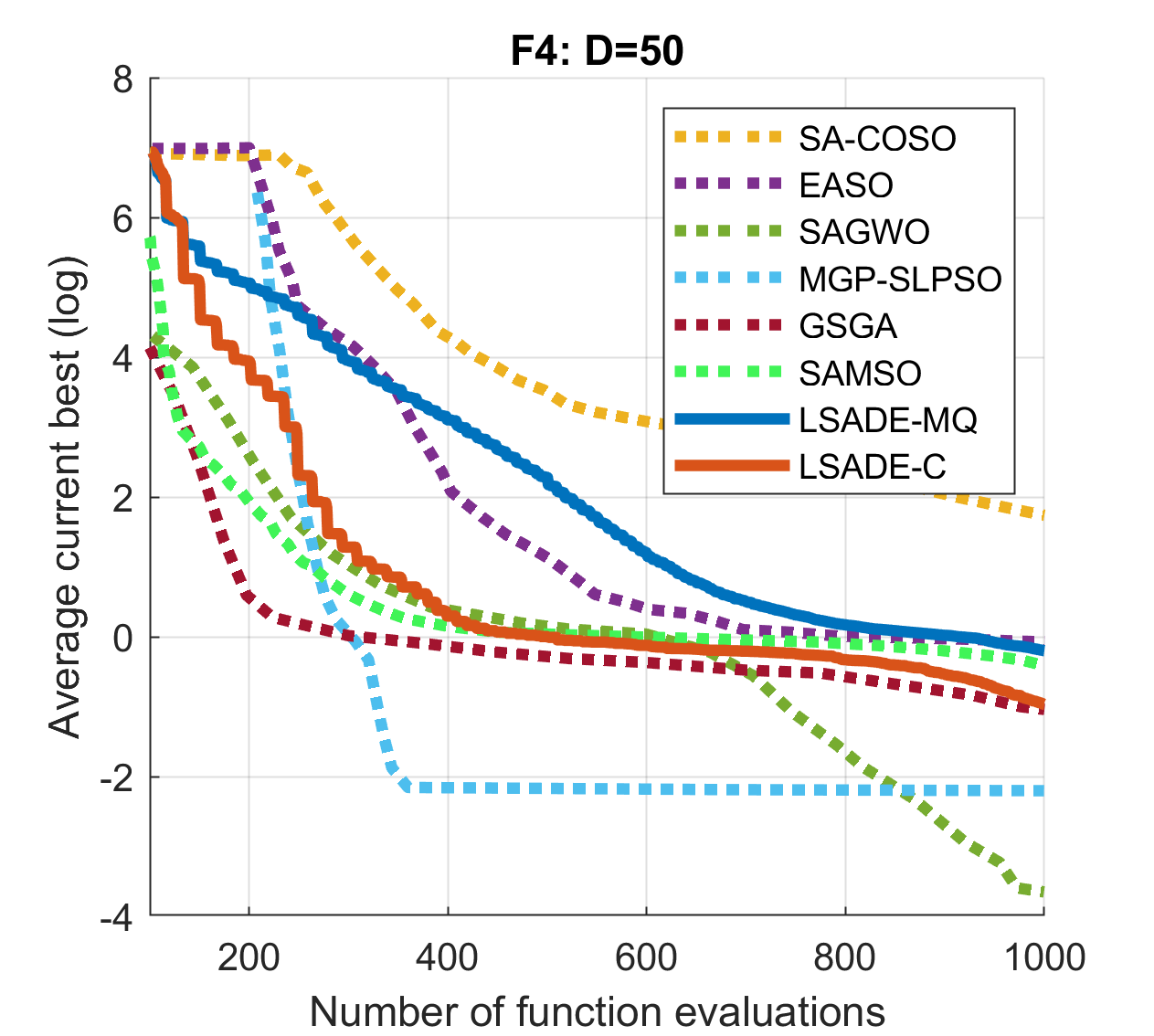} & \includegraphics[height = 0.28\textwidth]{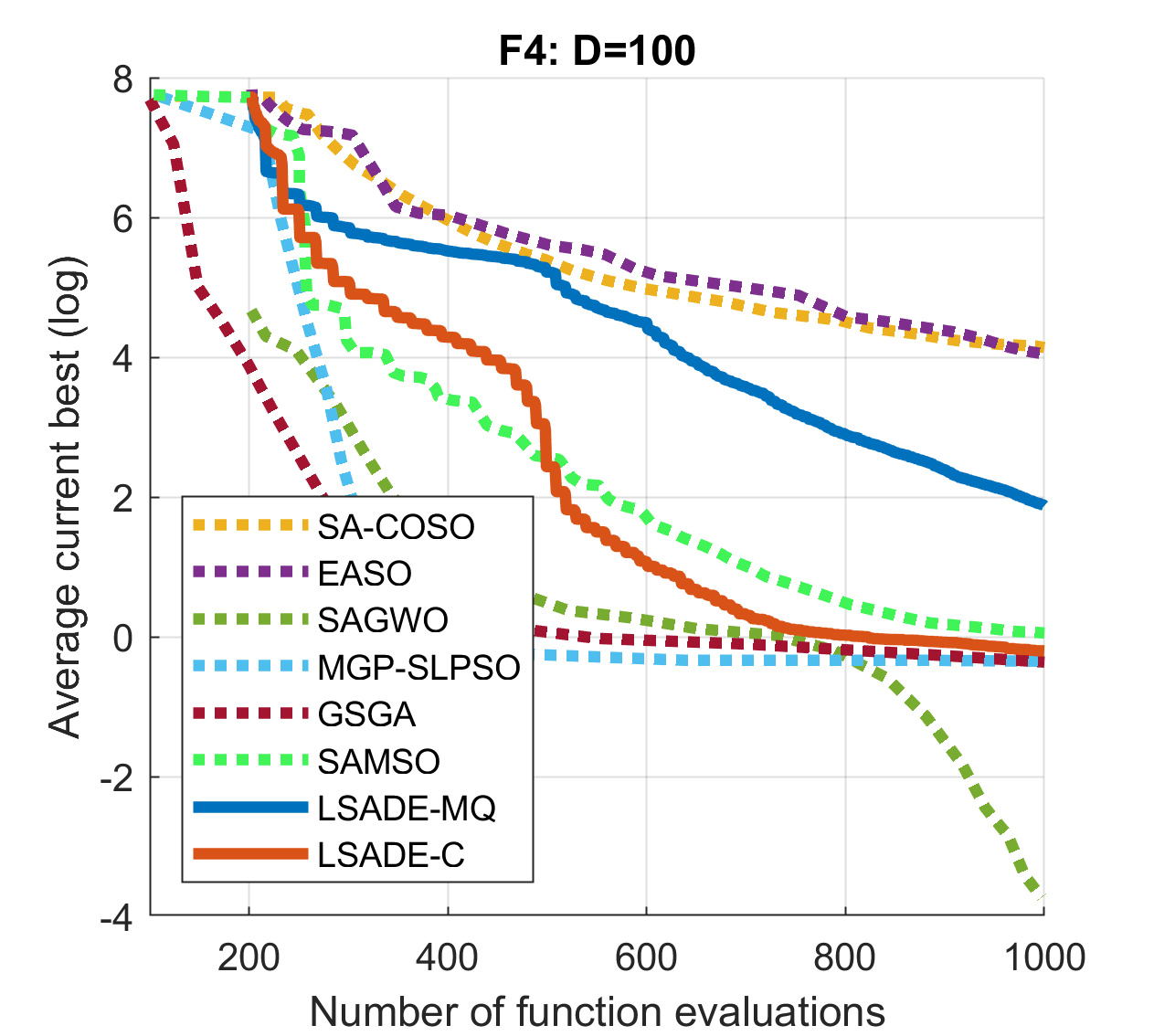} \\
\end{tabular}
\egroup
\caption{Convergence history of the considered algorithms on the benchmark functions F1--F4 in dimensions $D = [30,50,100]$.}
 \label{f:1a}
\end{figure*}

\begin{figure*}[ht!]
\centering
\setlength{\tabcolsep}{5pt}
\bgroup
\def\arraystretch{5}
\begin{tabular}{ccc}
\includegraphics[height = 0.28\textwidth]{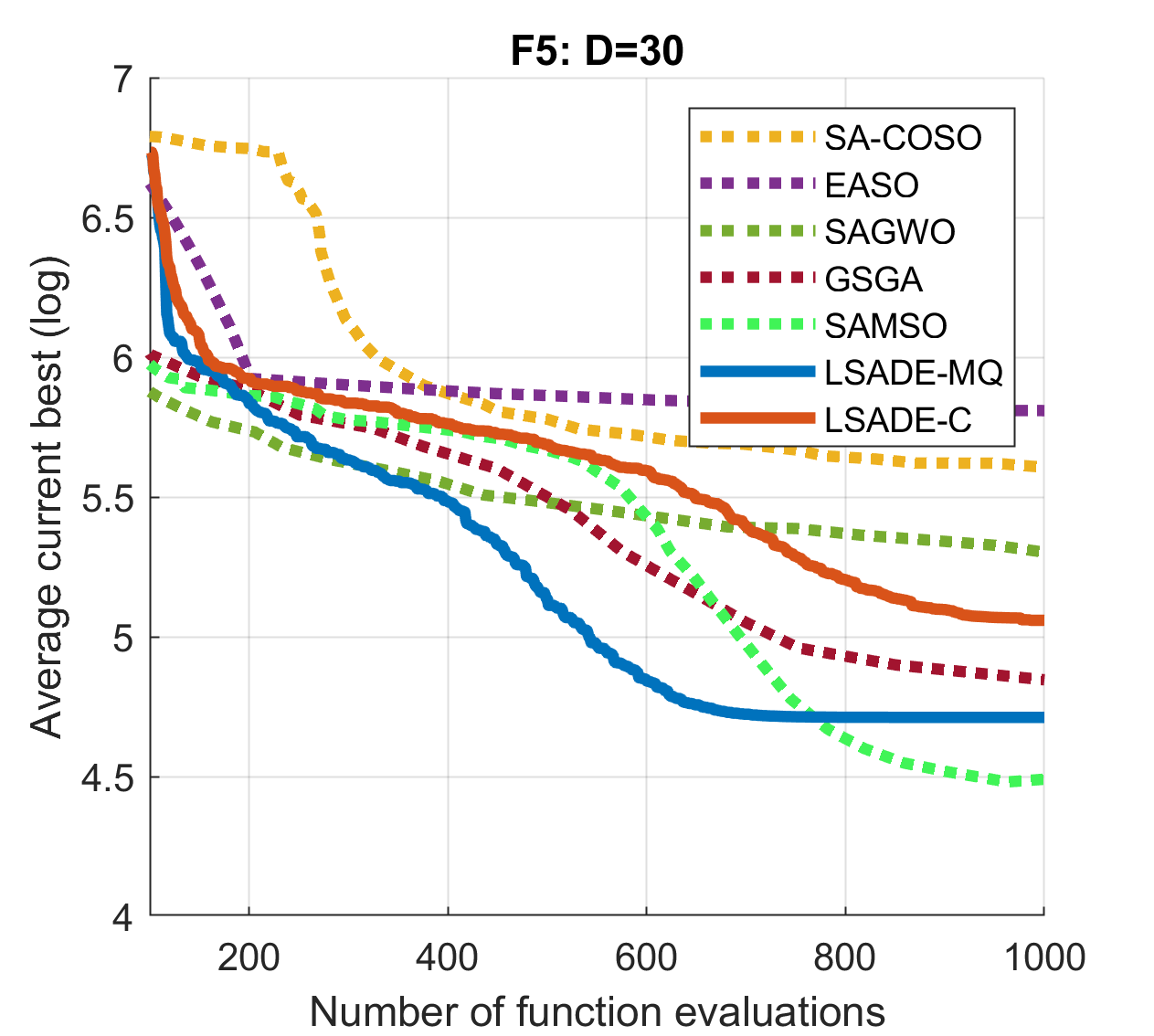} & \includegraphics[height = 0.28\textwidth]{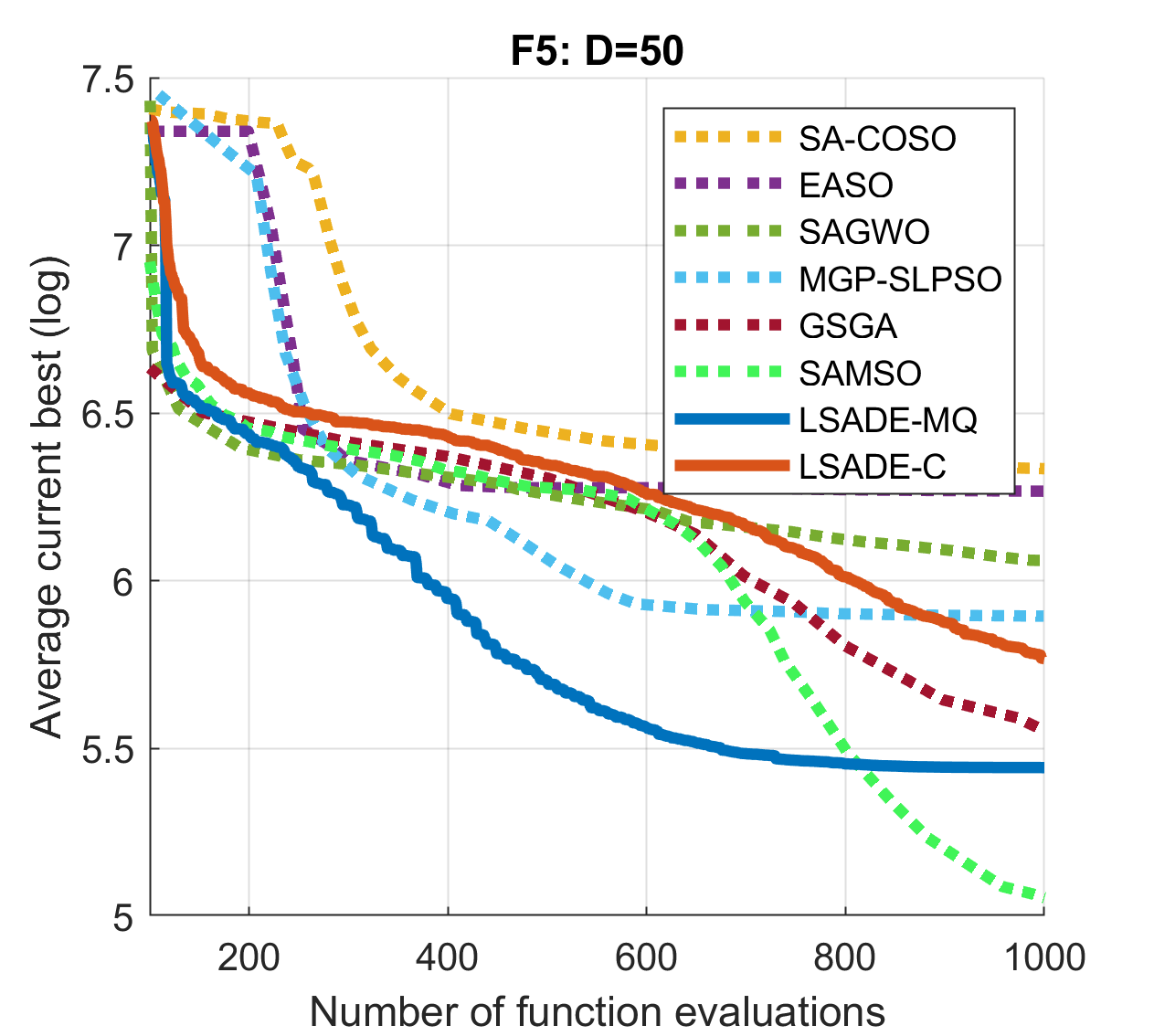} & \includegraphics[height = 0.28\textwidth]{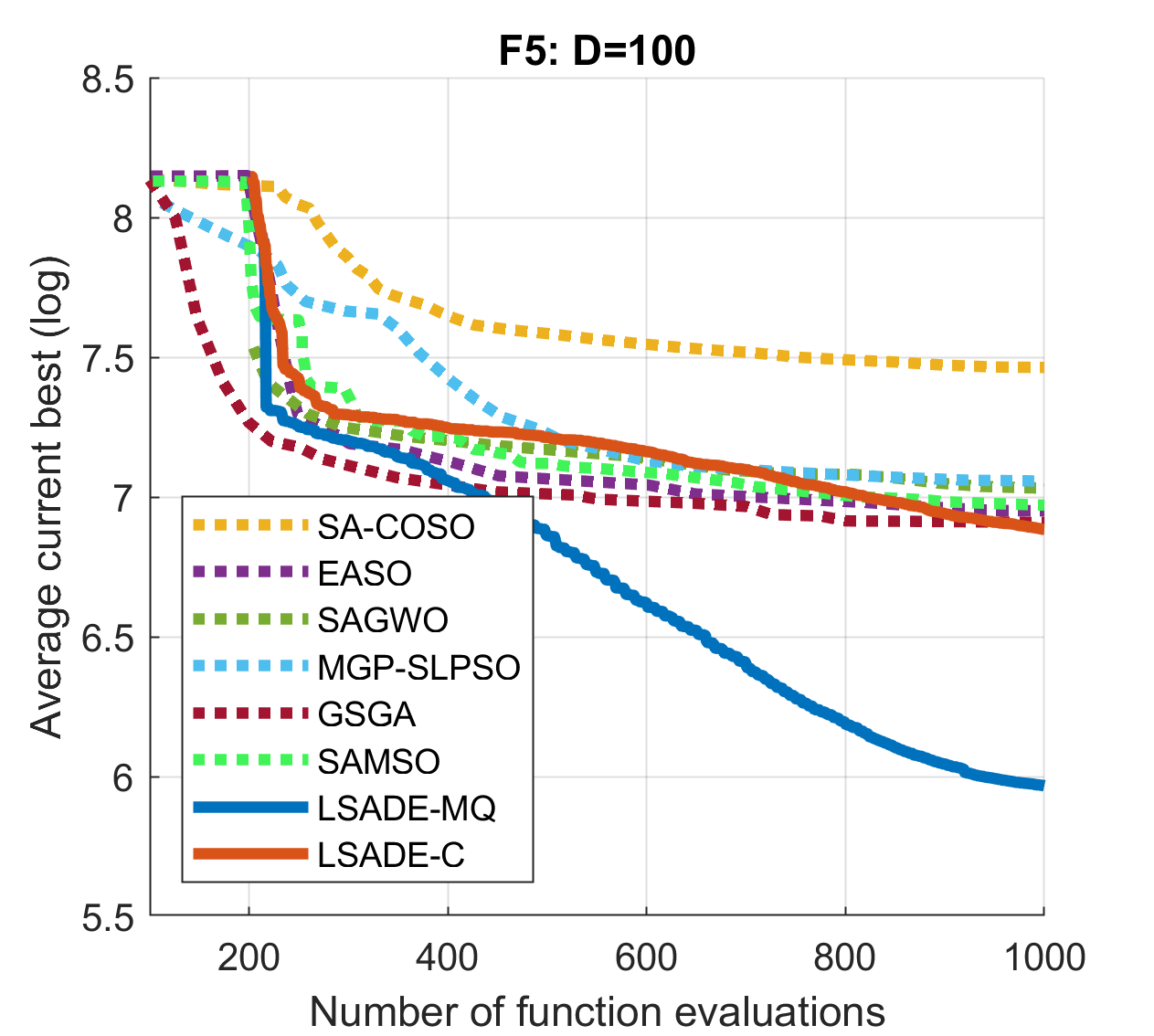} \\
\includegraphics[height = 0.28\textwidth]{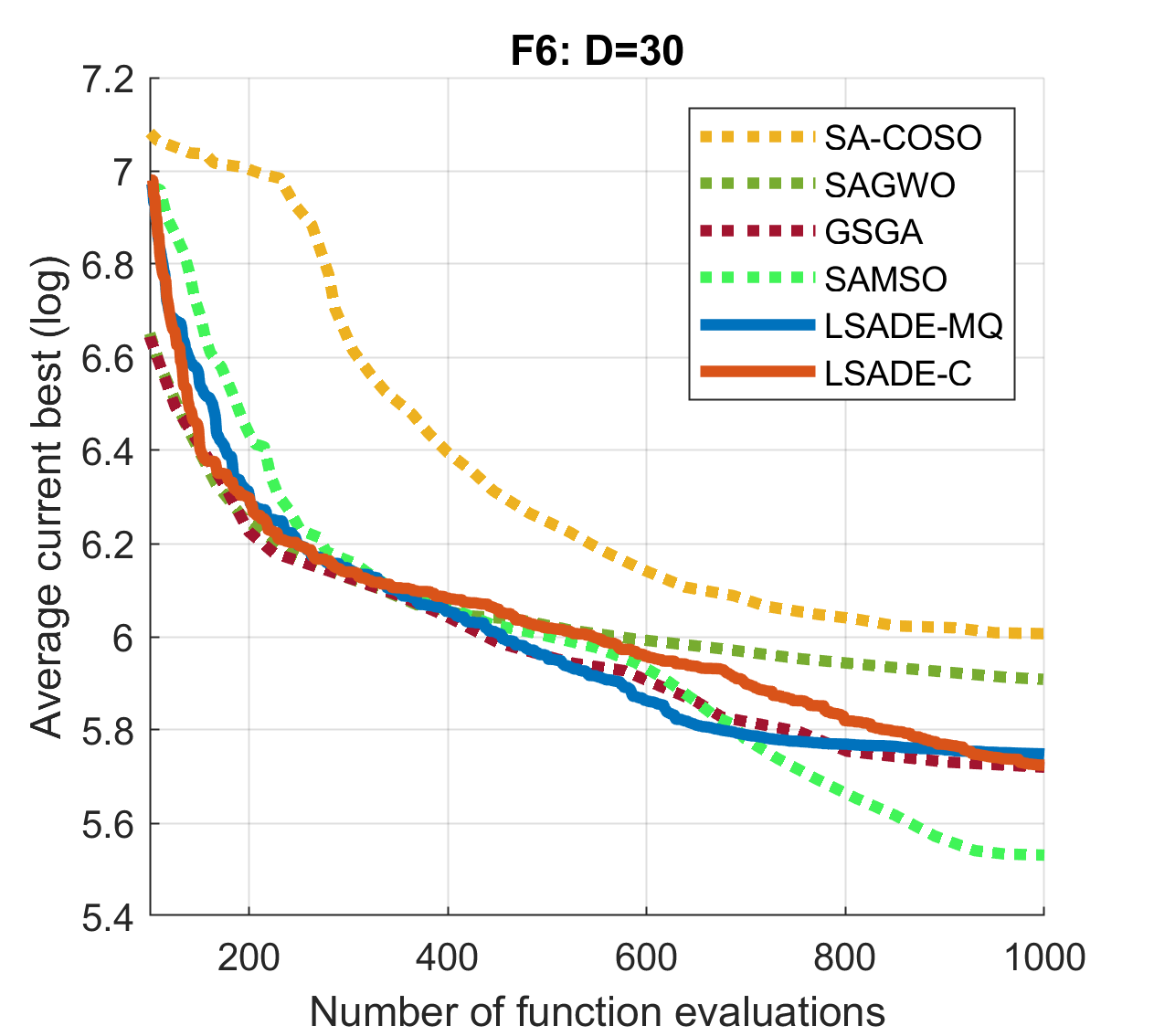} & \includegraphics[height = 0.28\textwidth]{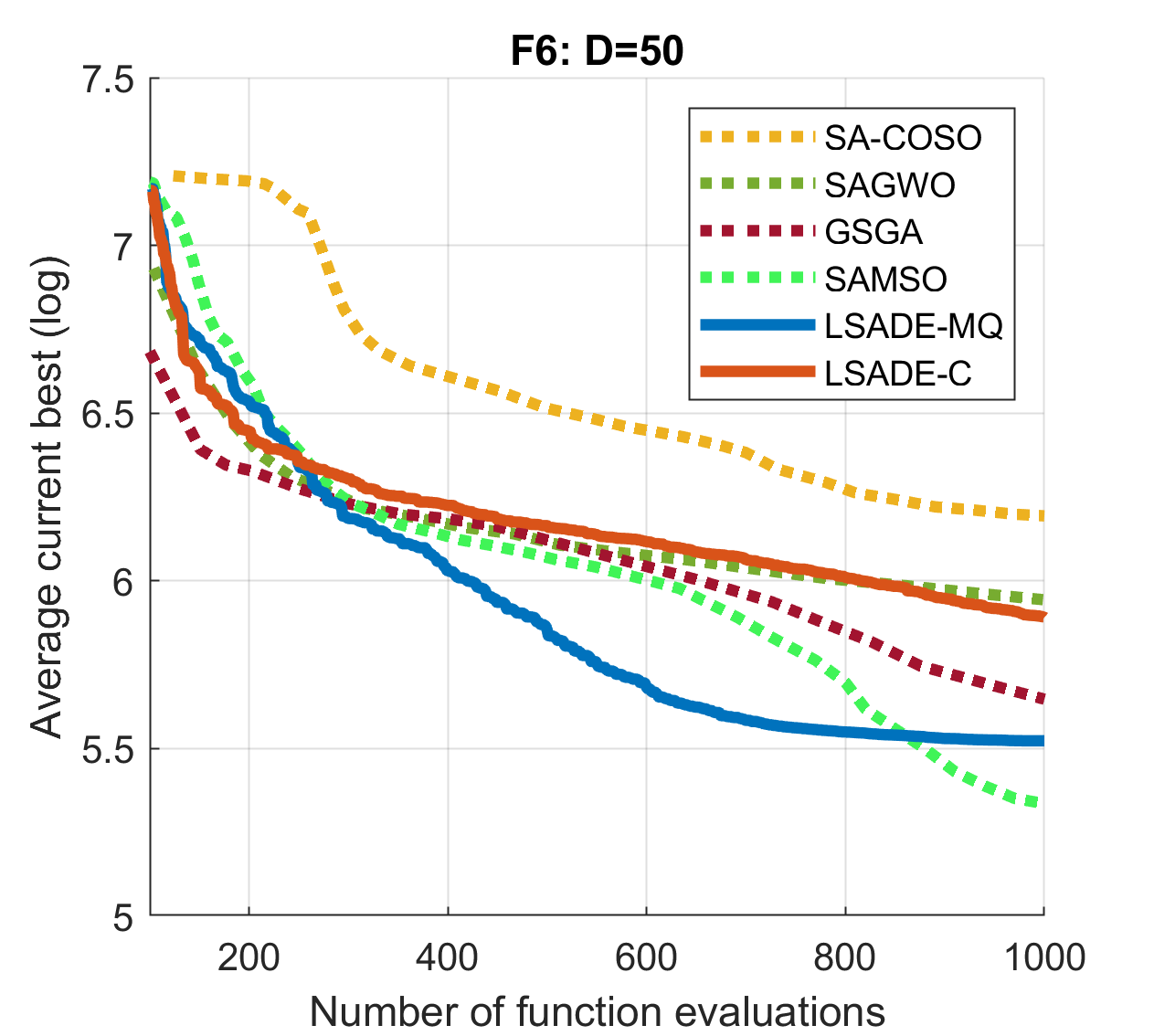} & \includegraphics[height = 0.28\textwidth]{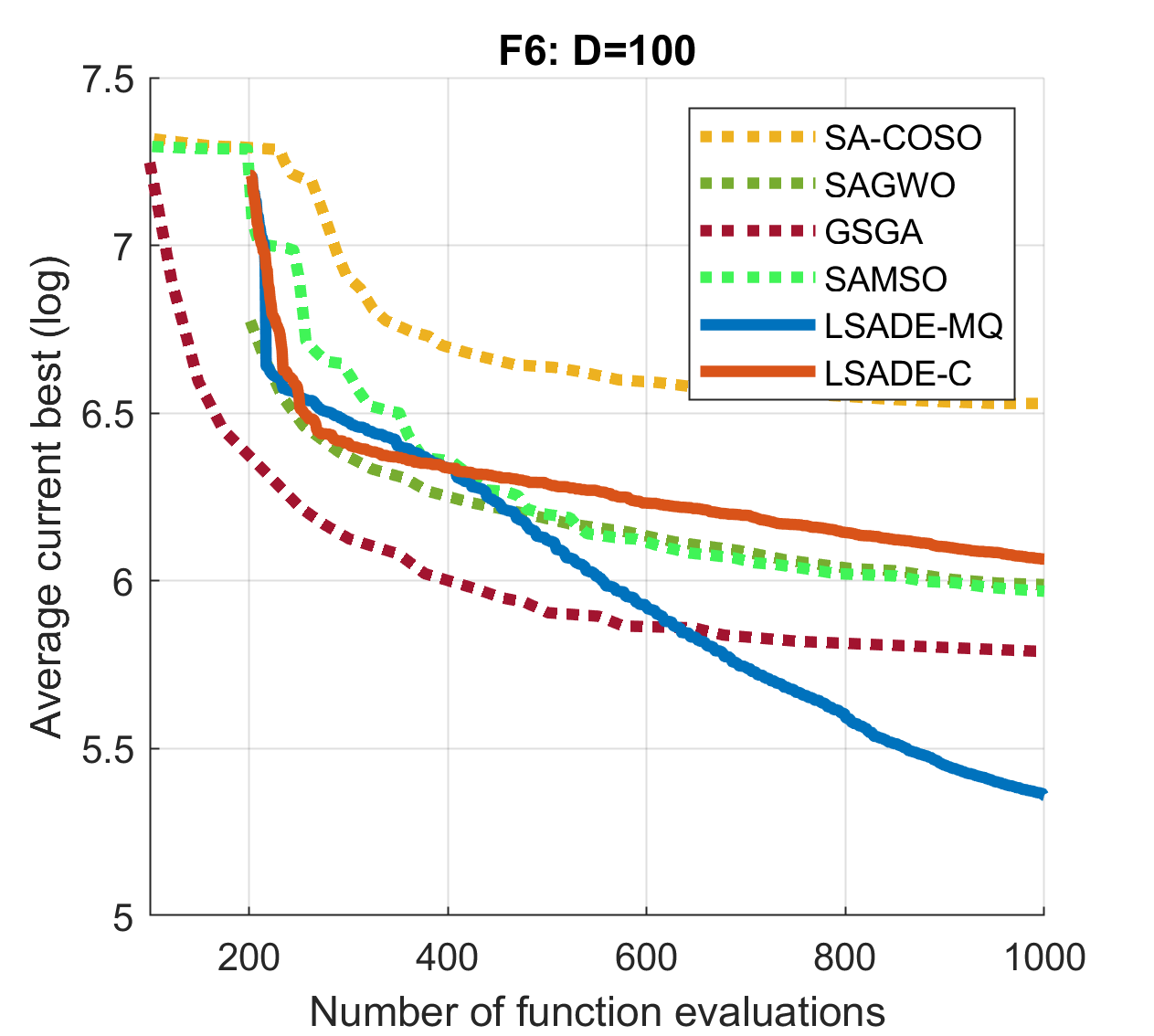} \\
\includegraphics[height = 0.28\textwidth]{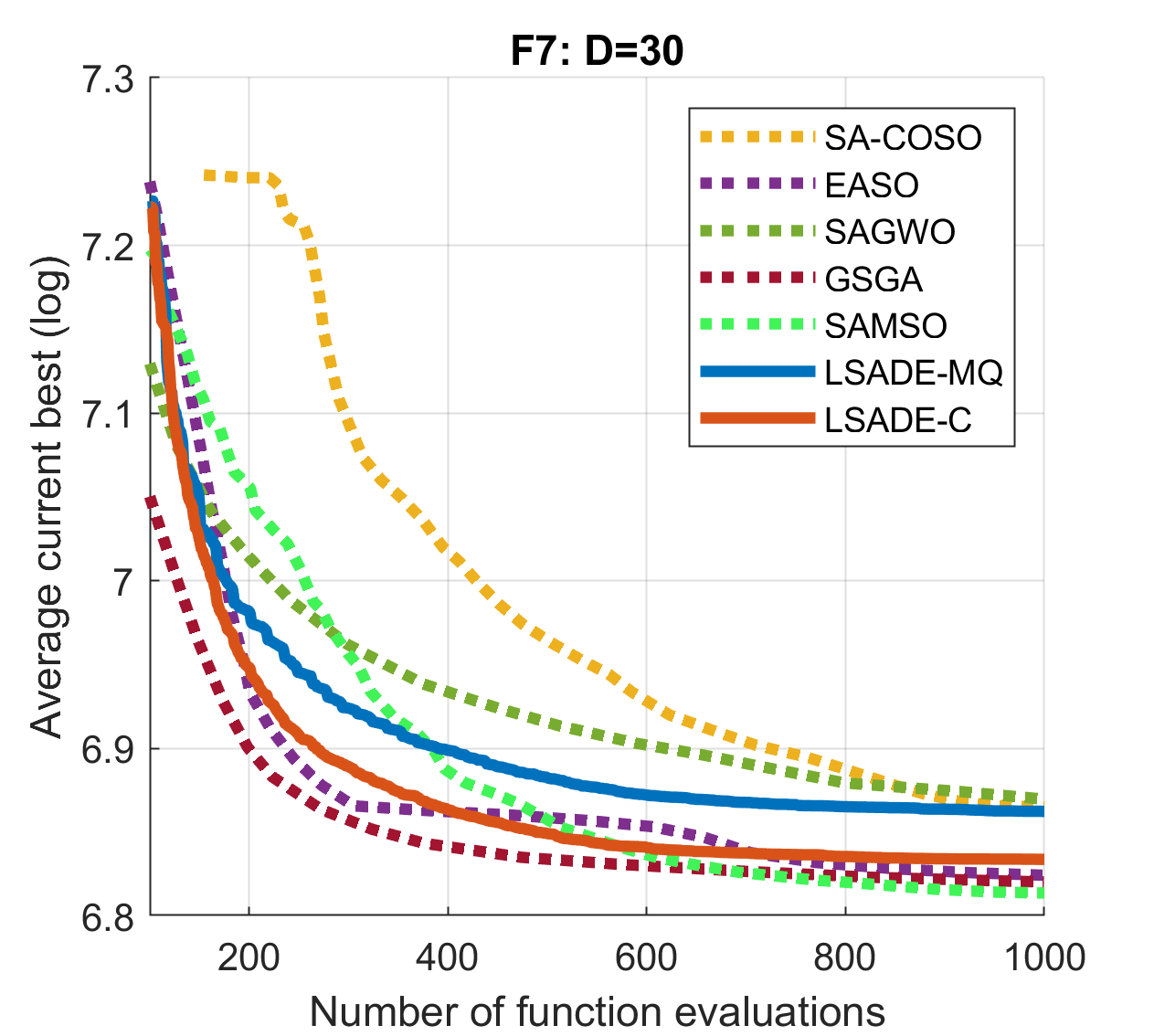} & \includegraphics[height = 0.28\textwidth]{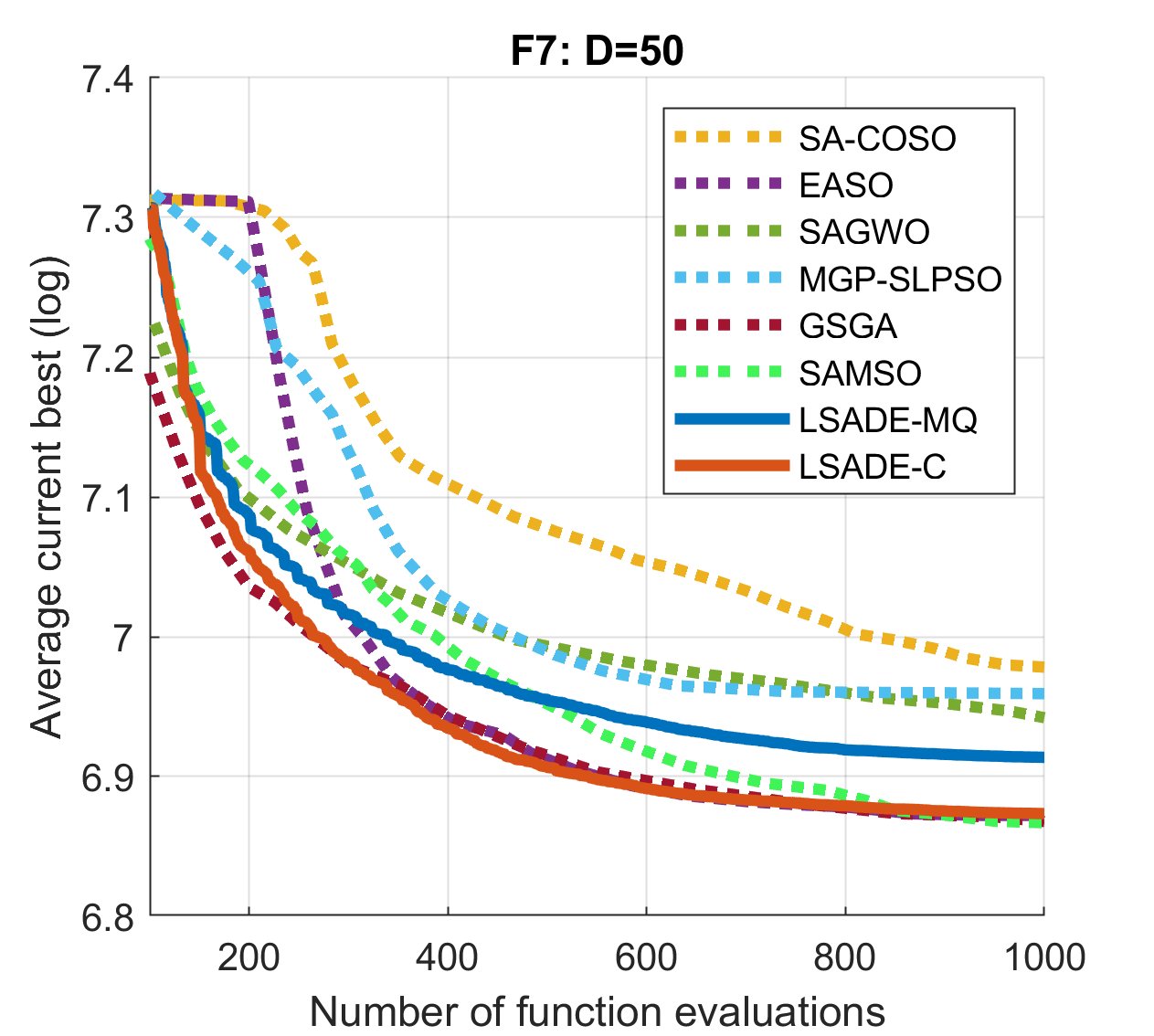} & \includegraphics[height = 0.28\textwidth]{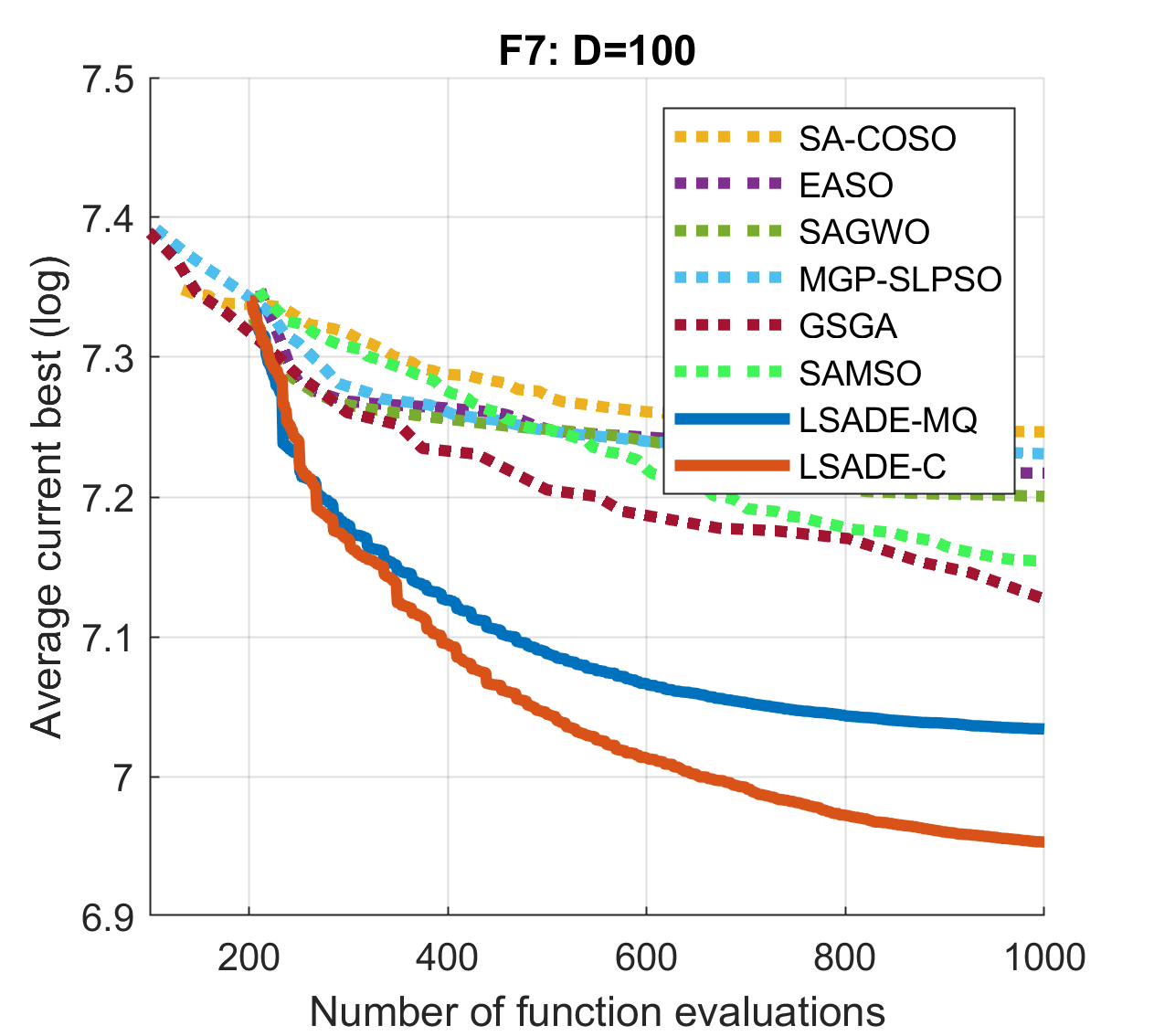} \\

\end{tabular}
\egroup
\caption{Convergence history of the considered algorithms on the complicated benchmark functions F5--F7 in dimensions $D = [30,50,100]$.}
 \label{f:1b}
\end{figure*}

The convergence histories of the considered algorithms for $D=[50,100,200]$ are depicted in Figures \ref{f:1a} and \ref{f:1b}, where on the $y$ axis are not the objective function values, but the difference between the objective function value and the corresponding optimum (otherwise, the $\log$ operator would fail for F5). For $D=200$, the convergence histories of the six compared algorithms were not available, and the convergence history of LSADE can be found in the Appendix.
From these results, it is quite clear that the LSADE algorithm with properly tuned rules for using the newly proposed Lipschitz surrogate model and local optimization procedure compares well to the state-of-the-art SAEAs, especially for the high-dimensional highly complicated benchmark problems.

\subsection{Computational Complexity}
For LSADE the computational complexity mainly consists of five parts: the computation time for initial search, creating and evaluating the local and global RBF surrogate models, creating and evaluating the Lipschitz model, local optimization, and real function evaluations. In the following, we focus on empirical analysis of the computational time for the surrogates and the local optimization procedure, as the time for real function evaluations depends on the problem the algorithm is applied to solve (these evaluations are expected to be costly, otherwise the algorithm should not be used). First, we compare the computational times for the individual components of the LSADE algorithm, using the R0$\,|\,$Li0$\,|\,$Lo1, R0$\,|\,$Li1$\,|\,$Lo0, and R1$\,|\,$Li0$\,|\,$Lo0 variants of the algorithm for the computation of the benchmark problems for $D=[30,50]$. The results of these computations are reported in Table \ref{tab:7}. We observe that the computation of Lipschitz surrogate model is significantly less computationally demanding than the computation of the (multiquadratic) RBF surrogate model. Unsurprisingly, the computational requirements for the local optimization are quite large, as these computations also contain the construction of the local RBF surrogate model.

\begin{table}[h!]
\centering
\caption{Computational time [s] of the individual components of LSADE, $D=[30,50]$.}
\bgroup
\def\arraystretch{1.6}
\setlength{\tabcolsep}{3.7pt}
\begin{tabular}{rl|rrr}
\multicolumn{1}{r}{$D$} & F  & \multicolumn{1}{l}{R0$\,|\,$Li0$\,|\,$Lo1} & \multicolumn{1}{l}{R0$\,|\,$Li1$\,|\,$Lo0} & \multicolumn{1}{l}{R1$\,|\,$Li0$\,|\,$Lo0} \\ \hline \hline
\multirow{7}{*}{30}   & F1 & 76.35                          & 3.44                           & 38.96                          \\
                      & F2 & 43.12                          & 3.36                           & 38.97                          \\
                      & F3 & 46.59                          & 3.19                           & 38.36                          \\
                      & F4 & 30.02                          & 3.03                           & 37.46                          \\
                      & F5 & 70.94                          & 3.40                           & 39.10                          \\
                      & F6 & 55.24                          & 7.01                           & 43.27                          \\
                      & F7 & 74.55                          & 7.03                           & 44.49                          \\ \hline \hline
\multirow{7}{*}{50}   & F1 & 239.17                         & 5.46                           & 51.15                          \\
                      & F2 & 158.25                         & 5.57                           & 51.57                          \\
                      & F3 & 153.27                         & 5.41                           & 53.34                          \\
                      & F4 & 85.77                          & 5.78                           & 53.40                          \\
                      & F5 & 231.53                         & 5.75                           & 52.91                          \\
                      & F6 & 170.06                         & 9.62                           & 56.26                          \\
                      & F7 & 229.90                         & 9.43                           & 55.92    \\ \hline \hline                     
\end{tabular}
\egroup
\label{tab:7}
\end{table}

\begin{table}[]
\centering
\caption{Comparison with other algorithms, computational times [s].} \label{t:8}
\bgroup
\def\arraystretch{1.4}
\setlength{\tabcolsep}{3.7pt}
\begin{tabular}{r|rrr|rr}
$D$   & SA-COSO & MGP-SLPSO & SAGWO & LSADE-MQ & LSADE-C \\ \hline 
30  & N/A     & N/A       & 226   & 33.24      & 54.14         \\
50  & 595     & 666       & 428   & 59.8       & 83.45         \\
100 & 833     & 741       & 1099  & 164.1      & 167.7         \\
200 & N/A     & N/A       & N/A   & 591.3      & 685.6        
\end{tabular}
\egroup
\end{table}

The computational requirements for different variants of the LSADE algorithm will differ based on the number of RBF and Lipschitz surrogate evaluations, and on the number of times the local optimization procedure is used. The number of times these individual components were used for the variant of LSADE that was chosen for numerical comparisons (Li1-4$\,|\,$Lo8-1), as well as for the other variants can be found in the Appendix. The average computational times of LSADE-MQ and LSADE-C for the benchmark problems for $D=[30,50,100,200]$ can be found in Table \ref{t:8}. The computational times for different variants of LSADE as well as for different basis functions can also be found in the Appendix. Also in Table \ref{t:8} are the computational times of SA-COSO, MGP-SLPSO, and SAGWO that were reported in the respective papers. As for the other compared algorithms, GSGA reported a computational time of 3 hours for the function F3 in $D=100$, and EASO and SAMSO did not include an empirical analysis of computational complexity. This comparison gives a clue as to why were the MGP-SLPSO, SAGWO, and GSGA algorithms not used for solving the large $D=200$ problems -- the computational times become a bit prohibitive for a large number of runs on numerous benchmark functions (but not necessarily prohibitive for a real application). On the other hand, the computational requirements for LSADE remain relatively low, with a dependence on the problem dimension that is roughly quadratic (at least for the considered benchmark problems). This is another indication that the LSADE algorithm is well suited for high-dimensional expensive problems.

The complexity of the Lipschitz surrogate itself depends on two main operations: on the estimation of the Lipschitz constant and on the evaluation of the surrogate. Through empirical analysis (performed on F7) shown in Table \ref{t:9} we can see that for the Lipschitz constant estimation there is a linear dependence of the computational time on the problem dimension $D$ and a quadratic dependence on the number of evaluated points $n$. Similarly, in Table \ref{t:10}, we find that the computational time for evaluating the Lipschitz surrogate depends linearly on both $D$ and $n$.

\begin{table}[]
\centering
\caption{Dependence of computational time [s] of computing the Lipschitz constant estimate on dimension $D$ and on the number of points for surrogate construction $n$.} \label{t:9}
\bgroup
\def\arraystretch{1.4}
\setlength{\tabcolsep}{3.7pt}
\begin{tabular}{rr|rrrrrr}
& & \multicolumn{6}{c}{$D$} \\
& & 30       & 50       & 100      & 200      & 500      & 1000     \\ \hline 
\multirow{6}{*}{$n$} & 30                   & 2.13E-04 & 9.45E-05 & 1.34E-04 & 1.81E-04 & 2.76E-04 & 4.84E-04 \\
& 50                   & 3.31E-04 & 3.03E-04 & 2.97E-04 & 4.52E-04 & 7.54E-04 & 1.32E-03 \\
& 100                  & 9.77E-04 & 8.73E-04 & 1.11E-03 & 1.60E-03 & 2.96E-03 & 5.40E-03 \\
& 200                  & 3.02E-03 & 3.41E-03 & 5.14E-03 & 6.27E-03 & 1.20E-02 & 2.13E-02 \\
& 500                  & 1.87E-02 & 2.16E-02 & 2.73E-02 & 3.91E-02 & 7.56E-02 & 1.34E-01 \\
& 1000                 & 6.98E-02 & 8.81E-02 & 1.13E-01 & 1.56E-01 & 3.11E-01 & 5.39E-01
\end{tabular}
\egroup
\end{table}

\begin{table}[]
\centering
\caption{Dependence of computational time [s] for evaluating 10{,}000 points on the Lipschitz surrogate model on dimension $D$ and on the number of points for surrogate construction $n$.} \label{t:10}
\bgroup
\def\arraystretch{1.4}
\setlength{\tabcolsep}{3.7pt}
\begin{tabular}{rr|rrrrrr}
& & \multicolumn{6}{c}{$D$} \\
& & 30       & 50       & 100      & 200      & 500      & 1000     \\ \hline 
\multirow{6}{*}{$n$} & 30                   & 7.09E-02 & 6.81E-02 & 8.57E-02 & 1.19E-01 & 2.15E-01 & 3.92E-01 \\
&50                   & 8.97E-02 & 1.05E-01 & 1.32E-01 & 1.85E-01 & 3.43E-01 & 6.08E-01 \\
 &100                  & 1.60E-01 & 1.88E-01 & 2.47E-01 & 3.53E-01 & 6.73E-01 & 1.19E+00 \\
&200                  & 3.07E-01 & 3.61E-01 & 4.81E-01 & 6.91E-01 & 1.33E+00 & 2.38E+00 \\
&500                  & 7.29E-01 & 8.88E-01 & 1.17E+00 & 1.69E+00 & 3.30E+00 & 5.93E+00 \\
&1000                 & 1.45E+00 & 1.76E+00 & 2.33E+00 & 3.37E+00 & 7.06E+00 & 1.22E+01
\end{tabular}
\egroup
\end{table}

\section{Conclusion} \label{S:5}
In this paper, we proposed a novel Lipschitz-based surrogate model for computationally expensive problems and used it to develop LSADE, a differential evolution-based surrogate-assisted evolutionary algorithm. The LSADE algorithm utilizes the combination of the Lipschitz-based and standard RBF surrogate models and a local optimization procedure to balance the exploration and the exploitation on a limited computational budget. The proposed LSADE algorithm was evaluated and its hyperparameters (such as the choice of the particular RBF surrogate and the frequency of its individual components) were tuned on a testbed of seven widely used 30, 50, 100, and 200 dimensional benchmark problems. The computational results show its effectiveness and competitiveness with other state-of-the-art algorithms, especially for complicated and high-dimensional problems. 

There still remains much room for further improvements. The conditions for including new points based on the Lipschitz surrogate and local optimization could be made in an adaptive manner based on the progress of the algorithm. Similarly, the use of different RBFs or ensembles within the same algorithm, or the use of different evolutionary algorithms could also make the method more effective for certain classes of problems. The method could also be tested on a more diverse set of benchmark functions. Future work will also include the extension of the Lipschitz-based surrogate model to multifidelity and multicriteria optimization problems and its application to real-world problems.

\section*{Acknowledgment}
This work was supported by The Ministry of Education, Youth and Sports of the Czech Republic project No. CZ.02.1.01/0.0/0.0/16\_026/0008392 ``Computer Simulations for Effective Low-Emission Energy'' and by IGA BUT: FSI-S-20-6538.

\newpage

\section*{Appendix A - Detailed Results for the Static Rules}
In Table \ref{tbS:1} are the detailed results for the static rules for $D=50$. It shows, once again, that using both the Lipschitz surrogate model and the local optimization procedure provides substantial benefits. On its own, using the Lipschitz surrogate model was better than using the local optimization procedure for benchmark functions F3, F4, F6 and F7. However, the combinations of these two components are far superior for all considered benchmark functions.

\begin{table}[]
\centering
\caption{Results for the static rule, $D=50$.}
\bgroup
\def\arraystretch{1.2}
\setlength{\tabcolsep}{4.3pt}
\begin{tabular}{c|rrrrrrr}
\,Li$\,|\,$Lo & {F1} & {F2} & {F3} & {F4} & {F5} & {F6} & {F7} \\ \hline 
0$\,|\,$0   & 285.5                & 214.4                & 18.36                 & 79.97                 & 272.9                 & 787.4                 & 1229                 \\
0$\,|\,$1   & 3.728                  & 65.41                 & 17.99                 & 191.9               & 20.03                  & 752.7                 & 1238                 \\
0$\,|\,$2   & 3.523                  & 66.94                 & 17.77                 & 122.0                & 4.710                   & 738.0                 & 1231                   \\
0$\,|\,$4   & 22.32                 & 69.03                 & 17.71                 & 73.31                 & 14.93                  & 703.4                 & 1209                 \\
0$\,|\,$8   & 5.086                  & 65.10                 & 17.72                 & 43.94                 & -6.35                  & 697.0                 & 1181                 \\
1$\,|\,$0   & 69.96                 & 161.8                & 10.58                & 9.118                  & 161.9                 & 567.8                 & 1047                 \\
2$\,|\,$0   & 41.16                 & 112.8                & 13.81                 & 5.002                  & 204.2                 & 597.8                 & 1102                \\
4$\,|\,$0   & 35.79                 & 97.44                 & 16.41             & 6.165                  & 181.6                 & 621.2                 & 1133                 \\
8$\,|\,$0   & 34.67                 & 90.68                 & 16.95                 & 8.976                  & 216.4                 & 654.0                 & 1162                 \\ \hline 
1$\,|\,$1   & 2.352                  & 65.12                 & 15.56                 & 6.464                  & -138.0                & 410.4                 & 1077                 \\
1$\,|\,$2   & 3.861                  & 62.05                 & 13.81                 & 1.628                  & -132.2                & 363.2                 & 1028                 \\
1$\,|\,$4   & 4.645                  & 61.36                 & 9.822                  & 1.082                  & -120.9                & 364.7                 & 1019                 \\
1$\,|\,$8   & 6.003                  & 49.13                 & 6.460                  & 1.045                  & -106.2                & 389.8                 & 1023                 \\

2$\,|\,$1   & 0.817                  & 60.57                 & 15.61                & 12.24                 & -76.34                 & 454.4                 & 1100                 \\
2$\,|\,$2   & 0.687                  & 55.65                 & 15.34                 & 2.408                  & -92.19                 & 423.2                 & 1102                 \\
2$\,|\,$4   & 1.253                  & 51.08                 & 14.16                 & 1.183                  & -90.54                 & 423.9                 & 1061                 \\
2$\,|\,$8   & 1.959                  & 50.49                 & 13.92                 & 1.010                  & -60.46                 & 440.9                 & 1058                 \\

4$\,|\,$1   & 0.575                  & 65.37                 & 16.07                 & 30.49                 & -66.03                 & 558.6                 & 1156                 \\
4$\,|\,$2   & 0.702                  & 56.74                 & 16.21                 & 6.269                  & -64.67                 & 544.1                & 1125                 \\
4$\,|\,$4   & 0.513                  & 54.82                 & 15.78                 & 1.424                  & -45.27                 & 491.7                 & 1095                 \\
4$\,|\,$8   & 0.967                  & 47.45                 & 15.69                 & 1.105                  & -46.09                 & 468.2                 & 1112                 \\

8$\,|\,$1   & 0.629                  & 62.46                 & 16.78                 & 71.78                 & -18.57                 & 615.8                 & 1195                \\
8$\,|\,$2   & 0.623                  & 58.93                 & 16.73                 & 18.43                 & -40.99                 & 570.5                 & 1181                 \\
8$\,|\,$4   & 0.445                  & 59.70                 & 16.72                 & 4.286                  & 0.750                   & 568.5             & 1156                 \\
8$\,|\,$8   & 0.708                  & 48.66                & 16.50                 & 1.587                  & -35.96                 & 533.9                 & 1150                
\end{tabular}
\egroup
\label{tbS:1}     
\end{table}

\section*{Appendix B - Conditions for the Dynamic Rules}
In Table \ref{tbS:2} are the conditions used for the dynamic rules of the LSADE algorithm. The $\text{mod}$ function gives the remainder after division (modulo operation) and $\lceil \cdot \rceil$ is the ceil operation that rounds the value inside to the nearest integer greater than or equal to that value.

\begin{table}[t!]
    \centering
     \caption{Conditions for dynamic rules of the different variants of the LSADE algorithm.}
     \bgroup
\def\arraystretch{1.8}
\setlength{\tabcolsep}{4.1pt}
    \begin{tabular}{c|r|r}
    Li$\,|\,$Lo & \multicolumn{1}{c|}{Lipschitz condition} &\multicolumn{1}{c}{Local Condition} \\ \hline
        {1-4$\,|\,$8-1}  & $\text{mod}(iter,\lceil \frac{8\cdot iter}{1000} \rceil ) = 0 $  & $\text{mod}{}{(iter,\lceil \frac{8000-15\cdot iter}{1000}  \rceil )} = 0 $\\
        {1-6$\,|\,$8-1}  & $\text{mod}{(iter,\lceil \frac{10\cdot iter}{1000} \rceil )} = 0 $  & $\text{mod}{(iter,\lceil \frac{8000-15\cdot iter}{1000}  \rceil )} = 0 $\\
        {1-8$\,|\,$8-1}  & $\text{mod}{(iter,\lceil \frac{14\cdot iter}{1000} \rceil )} = 0 $  & $\text{mod}{(iter,\lceil \frac{8000-15\cdot iter}{1000}  \rceil )} = 0 $\\
        {1-4$\,|\,$6-1}  & $\text{mod}{(iter,\lceil \frac{8\cdot iter}{1000} \rceil )} = 0 $  & $\text{mod}{(iter,\lceil \frac{6000-12\cdot iter}{1000}  \rceil )} = 0 $\\
        {1-6$\,|\,$6-1}  & $\text{mod}{(iter,\lceil \frac{10\cdot iter}{1000} \rceil )} = 0 $  & $\text{mod}{(iter,\lceil \frac{6000-10\cdot iter}{1000}  \rceil )} = 0 $\\
        {1-8$\,|\,$6-1}  & $\text{mod}{(iter,\lceil \frac{14\cdot iter}{1000} \rceil )} = 0 $  & $\text{mod}{(iter,\lceil \frac{6000-10\cdot iter}{1000}  \rceil )} = 0 $\\
        {1-4$\,|\,$4-1}  & $\text{mod}{(iter,\lceil \frac{8\cdot iter}{1000} \rceil )} = 0 $  & $\text{mod}{(iter,\lceil \frac{4000-8\cdot iter}{1000}  \rceil )} = 0 $\\
        {1-6$\,|\,$4-1}  & $\text{mod}{(iter,\lceil \frac{12\cdot iter}{1000} \rceil )} = 0 $  & $\text{mod}{(iter,\lceil \frac{4000-8\cdot iter}{1000}  \rceil )} = 0 $\\
        {1-8$\,|\,$4-1}  & $\text{mod}{(iter,\lceil \frac{15\cdot iter}{1000} \rceil )} = 0 $  & $\text{mod}{(iter,\lceil \frac{4000-8\cdot iter}{1000}  \rceil )} = 0 $\\
    \end{tabular}
    \egroup
    \label{tbS:2}     
\end{table}

\section*{Appendix C - Computational Complexity for Different Dynamic Rules and Basis Functions}
The computational complexity of the different variants of the LSADE algorithm depends on the number of times the algorithm computed the RBF global and local models, the Lipschitz model and the local optimization procedure. Based on the rules described in Table \ref{tbS:2}, the number of evaluation of the individual components of the LSADE algorithm for the different variations of the dynamic rule are shown in Table \ref{tbS:3}.

\begin{table}[h!]
    \centering
    \label{tab:my_label}
     \caption{Number of evaluations of the individual components of LSADE for different dynamic rules for $D=[30,50]$}
     \bgroup
\def\arraystretch{1.4}
\setlength{\tabcolsep}{3.7pt}
    \begin{tabular}{c|ccc}
    Li$\,|\,$Lo & global RBF surrogate & Lipschitz surrogate & local optimization\\ 
    & & & \textbf (+local RBF)  \\  \hline
        {1-4$\,|\,$8-1}  & 495 & 260 & 145 \\ 
        {1-6$\,|\,$8-1}  & 510 & 231 & 159 \\
        {1-8$\,|\,$8-1}  & 531 & 189 & 180 \\
        {1-4$\,|\,$6-1}  & 471 & 254 & 175 \\
        {1-6$\,|\,$6-1}  & 512 & 231 & 157\\
        {1-8$\,|\,$6-1}  & 533 & 189 & 178\\
        {1-4$\,|\,$4-1}  &  445 & 248 & 207\\
        {1-6$\,|\,$4-1}  &  469 & 200 & 231\\
        {1-8$\,|\,$4-1}  &  483 & 172 & 245\\
    \end{tabular}
    \egroup
    \label{tbS:3}     
\end{table}
\begin{table*}[ht!]
    \centering
    \label{tab:my_label}
     \caption{Computational time [s] for the different dynamic rules for $D=[30,50,100]$}
     \bgroup
\def\arraystretch{1.6}
\setlength{\tabcolsep}{3.7pt}
\begin{tabular}{rl|rrr|rrr|rrr}
$D$ & F & 1-4$\,|\,$8-1            & 1-6$\,|\,$8-1            & 1-8$\,|\,$8-1            & 1-4$\,|\,$6-1            & 1-6$\,|\,$6-1            & 1-8$\,|\,$6-1            & 1-4$\,|\,$4-1           & 1-6$\,|\,$4-1            & 1-8$\,|\,$4-1            \\ \hline \hline 
\multirow{7}{*}{30}  & F1 & 34.81  & 35.30  & 38.17  & 43.48  & 36.11  & 35.77  & 36.47  & 39.48  & 39.76  \\
                     & F2 & 36.44  & 38.48  & 42.56  & 53.36  & 37.84  & 37.81  & 40.02  & 42.55  & 43.94  \\
                     & F3 & 28.24  & 33.15  & 35.92  & 39.94  & 31.07  & 30.67  & 30.01  & 33.20  & 33.78  \\
                     & F4 & 29.62  & 33.24  & 36.32  & 40.36  & 32.01  & 32.38  & 31.52  & 32.34  & 34.22  \\
                     & F5 & 36.27  & 40.97  & 45.66  & 41.19  & 38.88  & 43.65  & 39.14  & 40.21  & 41.59  \\
                     & F6 & 36.95  & 39.75  & 48.67  & 43.02  & 40.63  & 47.03  & 39.86  & 40.17  & 42.91  \\
                     & F7 & 30.39  & 34.61  & 42.91  & 35.24  & 32.88  & 40.35  & 33.63  & 34.65  & 36.38  \\ \hline \hline
\multirow{7}{*}{50}  & F1 & 66.73  & 69.96  & 75.20  & 84.41  & 73.73  & 90.86  & 82.39  & 89.50  & 88.31  \\
                     & F2 & 70.27  & 74.56  & 81.22  & 97.81  & 81.04  & 94.28  & 93.03  & 97.69  & 97.13  \\
                     & F3 & 44.13  & 46.28  & 49.45  & 59.08  & 51.52  & 61.87  & 58.04  & 58.47  & 58.86  \\
                     & F4 & 49.63  & 52.70  & 58.52  & 64.85  & 59.76  & 75.67  & 63.64  & 70.84  & 74.52  \\
                     & F5 & 70.52  & 74.81  & 80.54  & 97.25  & 77.47  & 94.39  & 91.03  & 91.95  & 94.57  \\
                     & F6 & 64.46  & 69.44  & 73.05  & 80.62  & 69.74  & 77.45  & 82.91  & 85.41  & 86.28  \\
                     & F7 & 52.90  & 55.64  & 61.47  & 58.95  & 58.13  & 66.13  & 66.27  & 69.47  & 71.75  \\ \hline \hline 
\multirow{7}{*}{100} & F1 & 194.43 & 209.00 & 228.94 & 237.74 & 229.36 & 234.38 & 270.51 & 301.35 & 312.44 \\
                     & F2 & 200.79 & 225.49 & 244.92 & 256.21 & 239.94 & 248.38 & 287.85 & 327.42 & 331.41 \\
                     & F3 & 124.03 & 146.51 & 137.47 & 150.47 & 144.99 & 147.01 & 174.52 & 203.86 & 182.72 \\
                     & F4 & 136.74 & 161.40 & 185.65 & 163.92 & 168.58 & 185.41 & 199.25 & 238.07 & 238.36 \\
                     & F5 & 202.60 & 211.31 & 246.00 & 239.35 & 231.15 & 242.88 & 280.10 & 309.55 & 360.38 \\
                     & F6 & 163.63 & 170.32 & 199.43 & 187.03 & 176.06 & 199.93 & 219.83 & 246.08 & 276.70 \\
                     & F7 & 126.78 & 138.73 & 170.19 & 151.88 & 150.29 & 168.68 & 173.01 & 207.39 & 234.45 \\ \hline \hline
\end{tabular}
\egroup
\label{tbS:4}     
\end{table*}

In Table \ref{tbS:4} are the computational times for the different variation of the dynamic rule for $D=[30,50,100]$. We can see that the computational effort is tied most directly to the number of times the local optimization procedure was used -- the variants that use it more often needed more computational time, especially when the dimension of the problems increased. Another interesting observation can be made regarding the difference in computational complexity for the different benchmark functions -- F1, F2, and F5 seem to require significantly more computational effort for the dynamic rules, especially in higher dimensions. We can compare this observation with the computational times for the individual components of LSADE that is reported in the paper. There, we can see that the computational times for local optimization procedure were quite high for problems F1, F5, and F7, while the other two components had only small dependence of computational time on the benchmark function. 

However, when we look at the computational complexity for different basis functions, that is reported in Table \ref{tbS:5}, we see that this dependence on the benchmark function is not shared among them. What we see instead is that for each choice of a basis function there are benchmark functions for which the computations seem to be more ``difficult'', regardless of dimension. For instance, F3 and F4 need more computational time for the linear basis function, while being among the ``easiest'' for the multiquadratic basis function. This could be explained by the different nature (and, thus, different ``difficulty'') of the local RBF models for the sequential quadratic programming optimizer that is used as the local optimization procedure.

\section*{Appendix D - Convergence Histories for Different Basis Functions}
The convergence histories for different basis functions are depicted in Figure \ref{fS:1a}. We can see that, most of the time, the best variant (i.e., the best choice of the basis function) of LSADE for a particular problem instance did not plateau around the 1000 real function evaluation limit. Also, the best performing variant for the particular problem instance (i.e, the one that had be best result after 1000 real function evaluations) is not necessarily the one that was the best when the number of real function evaluations was smaller. This phenomenon can be clearly observed for the $D=200$ benchmark problems, where the convergence histories for cubic and multiquadratic basis functions cross one another for the majority of the considered benchmark functions. This suggests that it may be beneficial to consider several basis functions in an ensemble at the same time and find a rule for using one of them based on the properties of the particular problem.

\begin{table}[h!]
    \centering
     \caption{Computational time [s] for the different basis functions, $D=[30,50,100,200]$.}
     \bgroup
\def\arraystretch{1.6}
\setlength{\tabcolsep}{3.7pt}
\begin{tabular}{rl|rrrrr}
$D$ &  F  & {MQ} &{Cubic} & {TPS} & {Linear} & {Gaussian} \\ \hline \hline
\multirow{7}{*}{30}  & F1 & 34.81                  & 54.53                     & 43.40                   & 33.71                      & 45.88                        \\
                     & F2 & 36.44                  & 58.33                     & 49.46                   & 32.42                      & 50.65                        \\
                     & F3 & 28.24                  & 51.36                     & 41.60                   & 48.91                      & 51.99                        \\
                     & F4 & 29.62                  & 62.10                     & 54.11                   & 54.30                      & 51.92                        \\
                     & F5 & 36.27                  & 54.06                     & 44.84                   & 33.63                      & 38.53                        \\
                     & F6 & 36.95                  & 53.23                     & 51.09                   & 37.53                      & 43.71                        \\
                     & F7 & 30.39                  & 45.41                     & 49.36                   & 38.11                      & 40.61                        \\ \hline \hline
\multirow{7}{*}{50}  & F1 & 66.73                  & 102.76                    & 92.01                   & 44.63                      & 86.73                        \\
                     & F2 & 70.27                  & 108.29                    & 85.52                   & 48.05                      & 85.45                        \\
                     & F3 & 44.13                  & 64.07                     & 62.10                   & 77.25                      & 76.60                        \\
                     & F4 & 49.63                  & 93.61                     & 84.86                   & 76.13                      & 77.39                        \\
                     & F5 & 70.52                  & 76.13                     & 70.61                   & 41.16                      & 47.43                        \\
                     & F6 & 64.46                  & 72.62                     & 73.43                   & 44.33                      & 53.72                        \\
                     & F7 & 52.90                  & 66.70                     & 59.00                   & 43.56                      & 52.37                        \\ \hline \hline 
\multirow{7}{*}{100} & F1 & 194.43                 & 243.85                    & 201.95                  & 93.25                      & 214.97                       \\
                     & F2 & 200.79                 & 257.15                    & 209.09                  & 104.64                     & 189.06                       \\
                     & F3 & 124.03                 & 171.90                    & 191.66                  & 168.00                     & 175.51                       \\
                     & F4 & 136.74                 & 179.88                    & 179.44                  & 167.52                     & 131.28                       \\
                     & F5 & 202.60                 & 107.27                    & 90.91                   & 70.36                      & 84.71                        \\
                     & F6 & 163.63                 & 104.77                    & 88.37                   & 80.79                      & 88.68                        \\
                     & F7 & 126.78                 & 109.52                    & 93.51                   & 90.94                      & 96.84              \\ \hline \hline 
                     \multirow{7}{*}{200} & F1 & 883.57                 & 1078.10     & -- & -- & --                \\
                     & F2 & 847.72                 & 1114.60           & -- & -- & --        \\
                     & F3 & 446.99                 & 801.77          & -- & -- & --            \\
                     & F4 & 420.28                 & 501.90          & -- & -- & --            \\
                     & F5 & 640.02                 & 546.68        & -- & -- & --              \\
                     & F6 & 467.05                 & 327.74     & -- & -- & --                 \\
                     & F7 & 433.72                 & 428.97     & -- & -- & --                \\ \hline \hline 
\end{tabular}
\egroup
\label{tbS:5}
\end{table}

\section*{Appendix E - Detailed Results for the Algorithms Considered for the Comparison}
In Tables \ref{tbS:6} and \ref{tbS:7} are detailed results of the computations of the six algorithms considered for comparison and two two LSADE variants (LSADE-MQ and LSADE-C). These detailed results were obtained from the respective publications, with the expection of the results for EASO and SA-COSO that were obtained from the SAMSO paper, and contain the best value, mean, the worst value, and standard deviation from the corresponding computational experiments (for some algorithms, some of these statistics were not available, and not all of the algorithms were tried on all of the benchmark functions).

From these results, we can see that although the LSADE variants are mediocre for the benchmark problems in smaller dimensions, they are performing very well in the dimensions $D=[100,200]$, especially for the benchmark functions F5-F7 with a more complicated multimodal landscape. For instance, the worst solution obtained by LSADE-MQ in $D=100$ for benchmark functions F5-F7 was better than the mean of the solutions of all other compared algorithms (except for LSADE-C).

\begin{figure*}[ht!]
\centering
\setlength{\tabcolsep}{5pt}
\bgroup
\def\arraystretch{5}
\begin{tabular}{cccc}
\includegraphics[height = 0.2\textwidth]{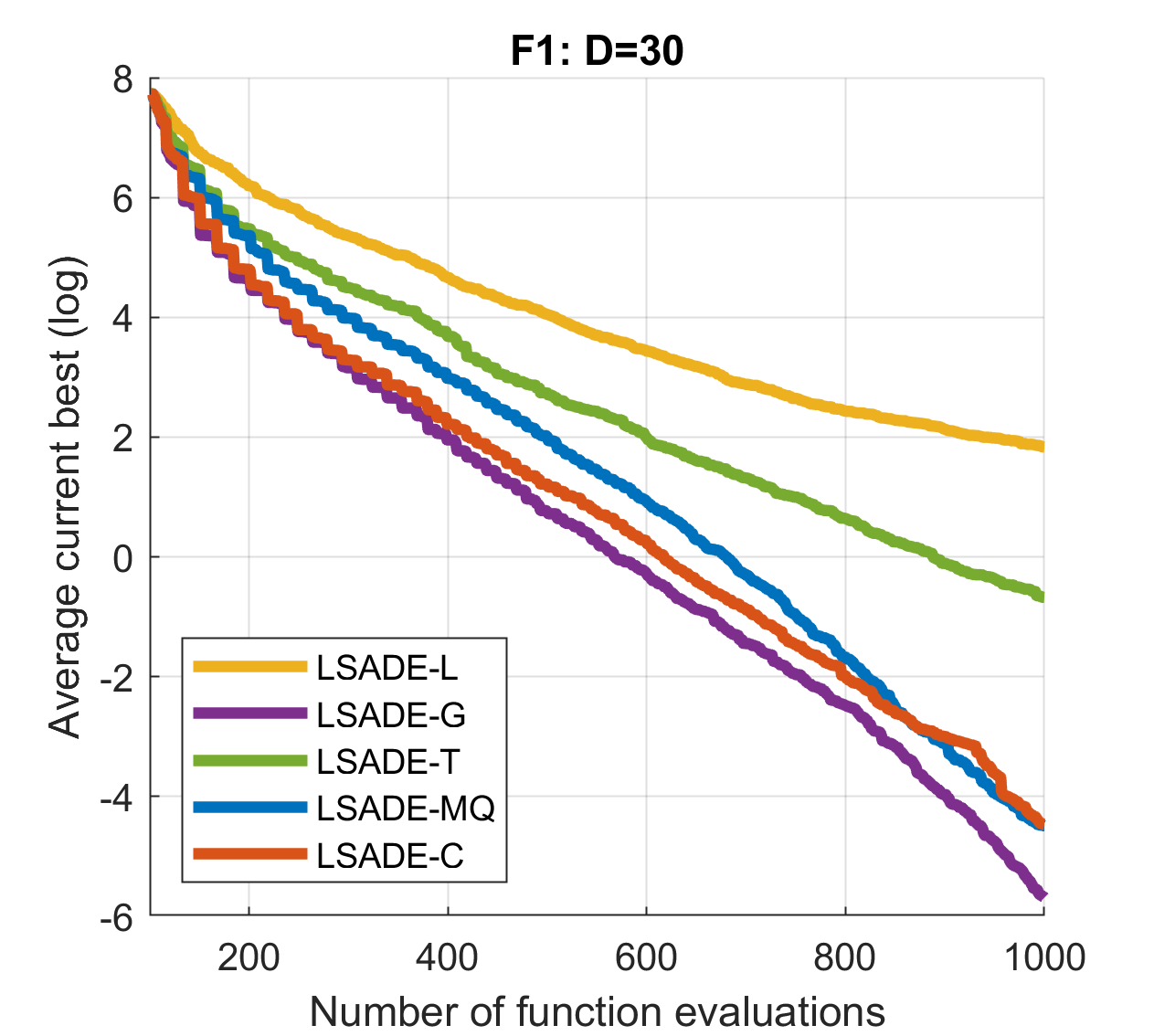} & \includegraphics[height = 0.2\textwidth]{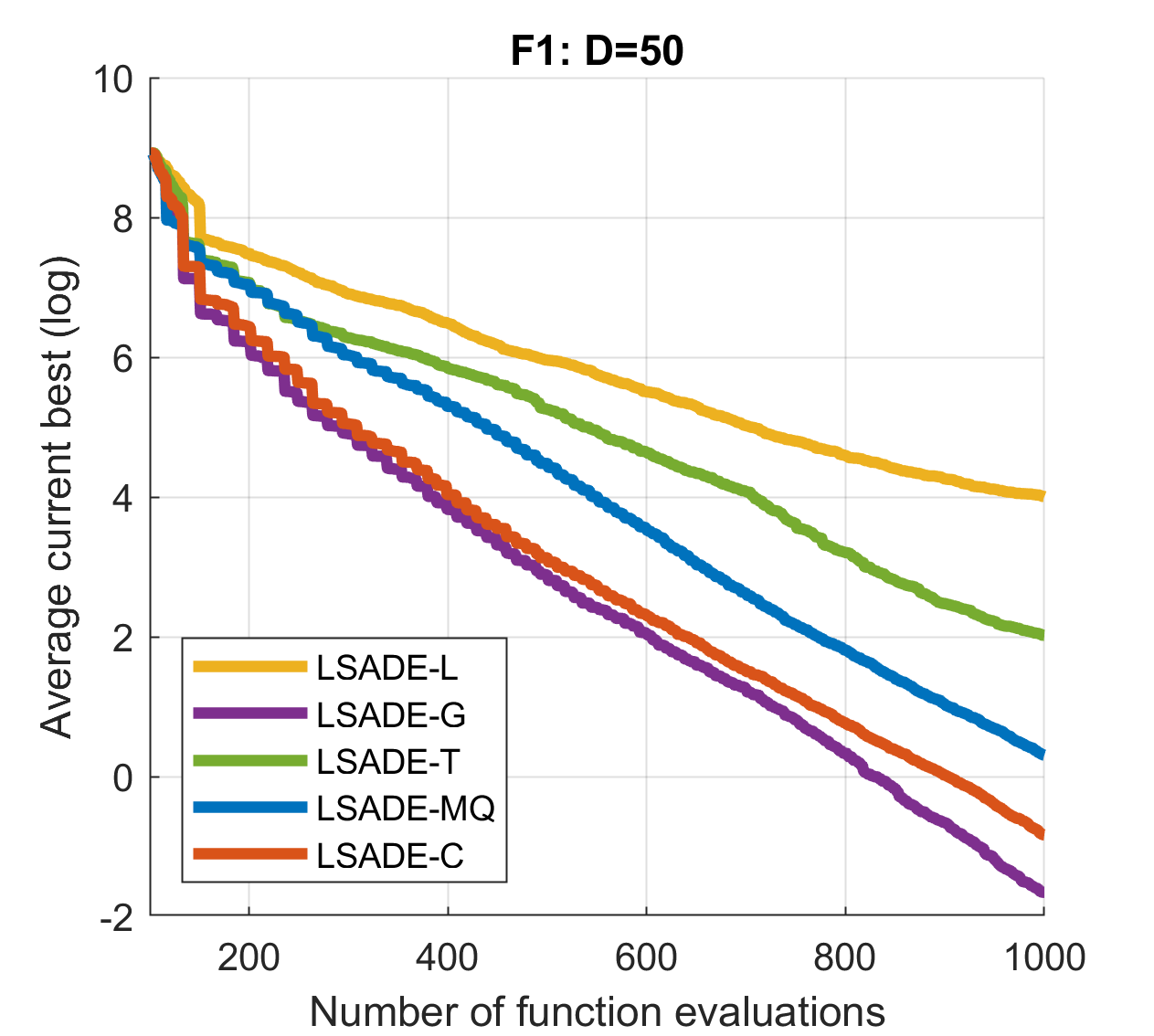} & \includegraphics[height = 0.2\textwidth]{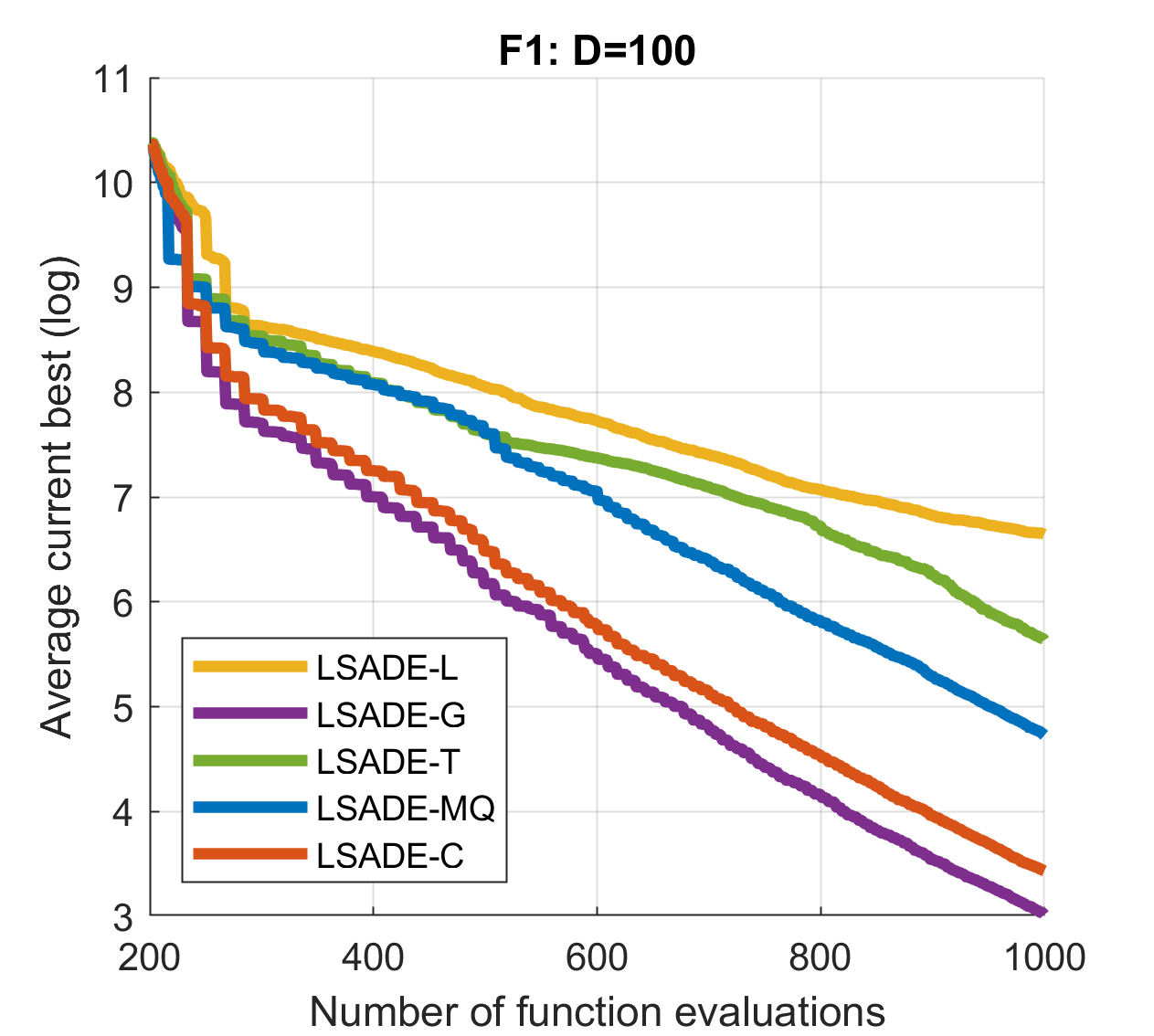} & \includegraphics[height = 0.2\textwidth]{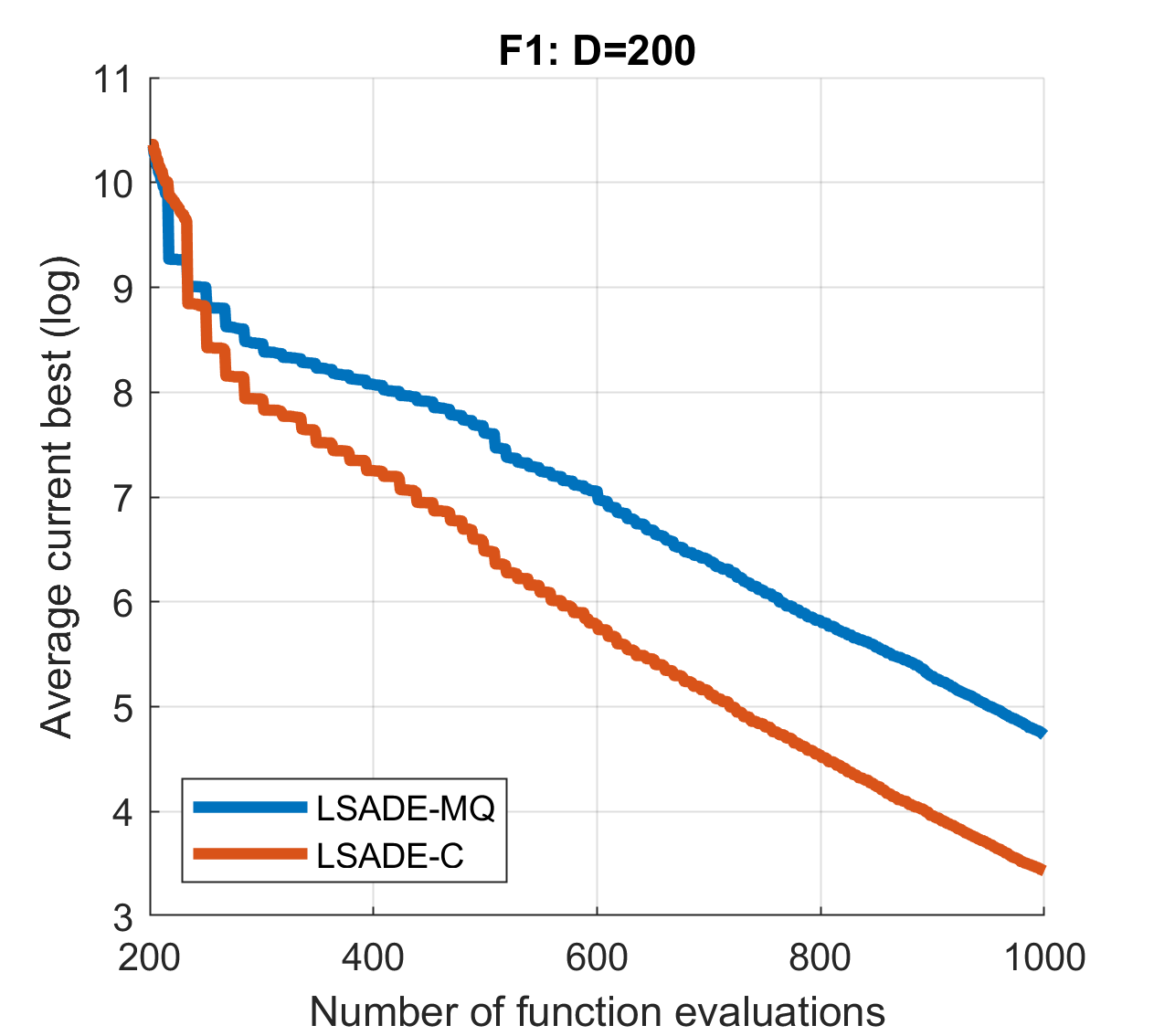}\\
\includegraphics[height = 0.2\textwidth]{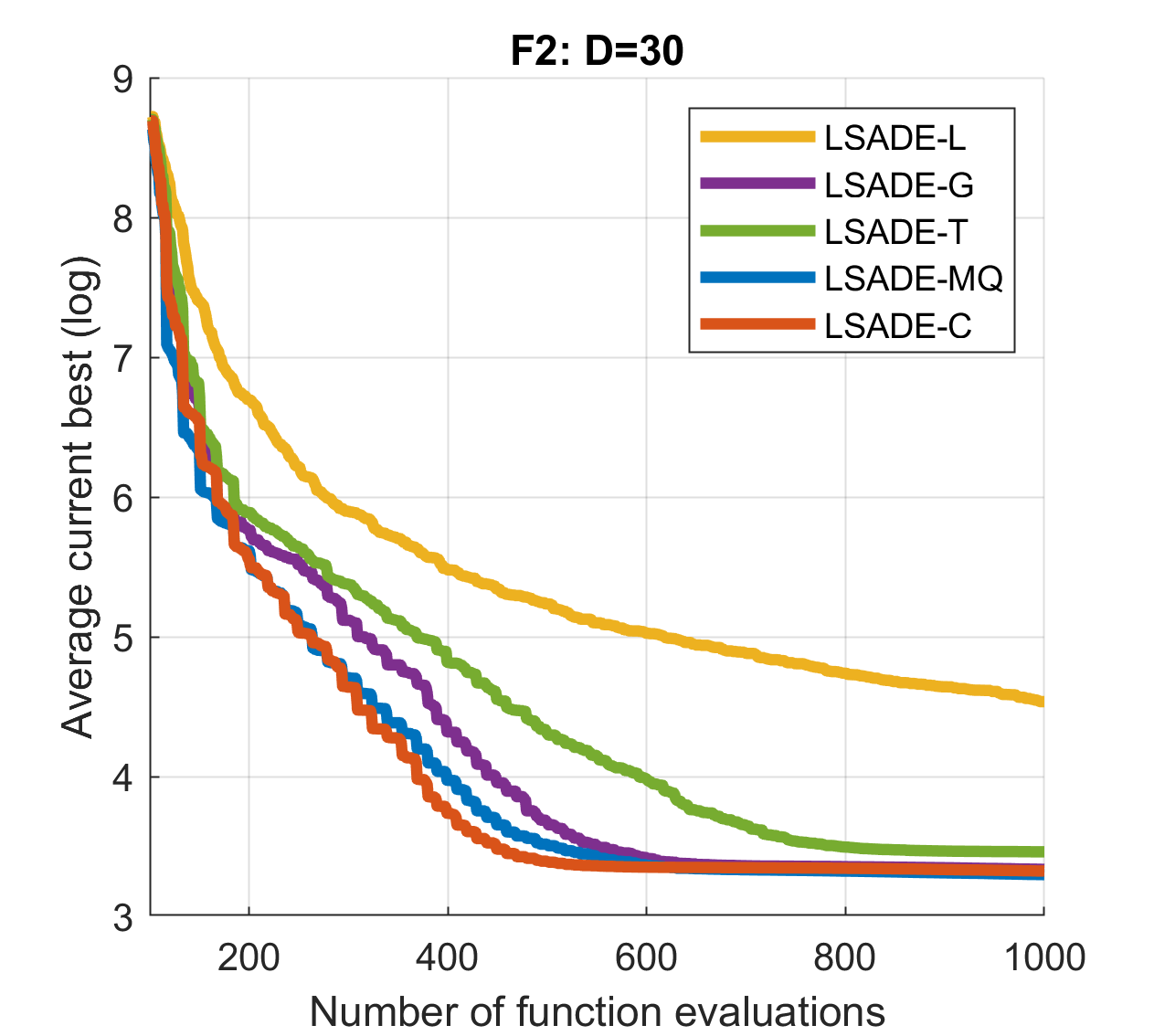} & \includegraphics[height = 0.2\textwidth]{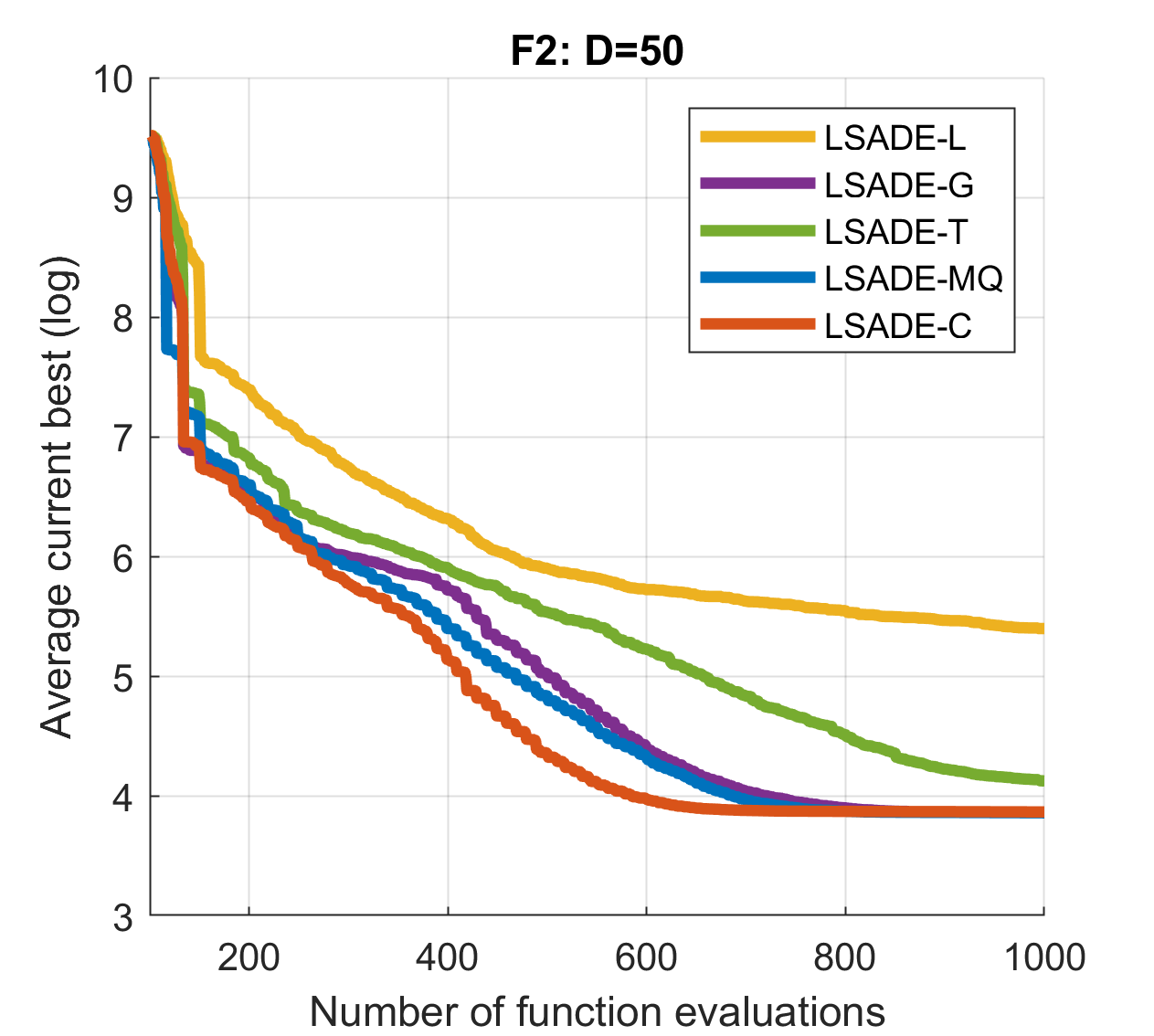} & \includegraphics[height = 0.2\textwidth]{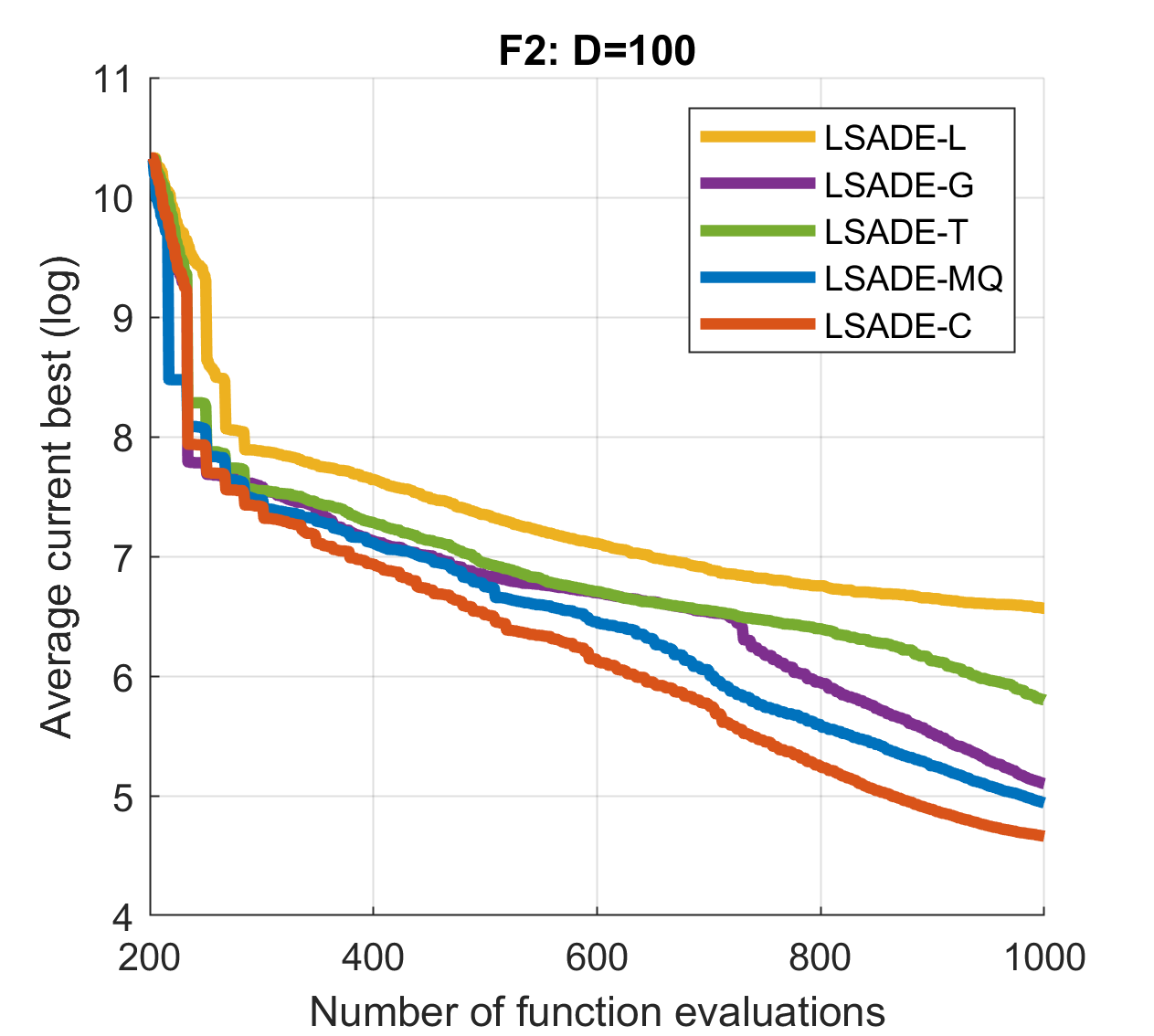} & \includegraphics[height = 0.2\textwidth]{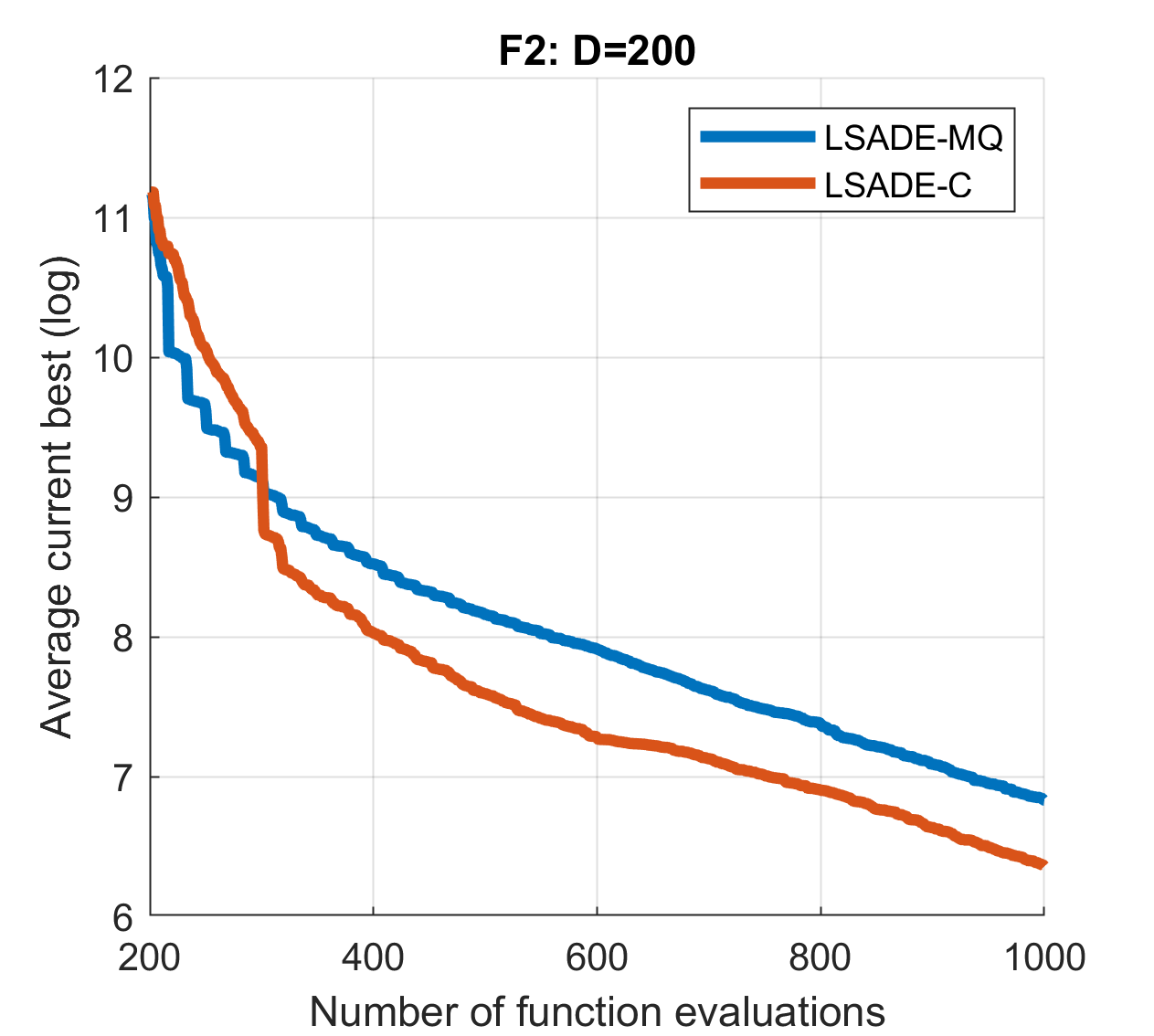}\\
\includegraphics[height = 0.2\textwidth]{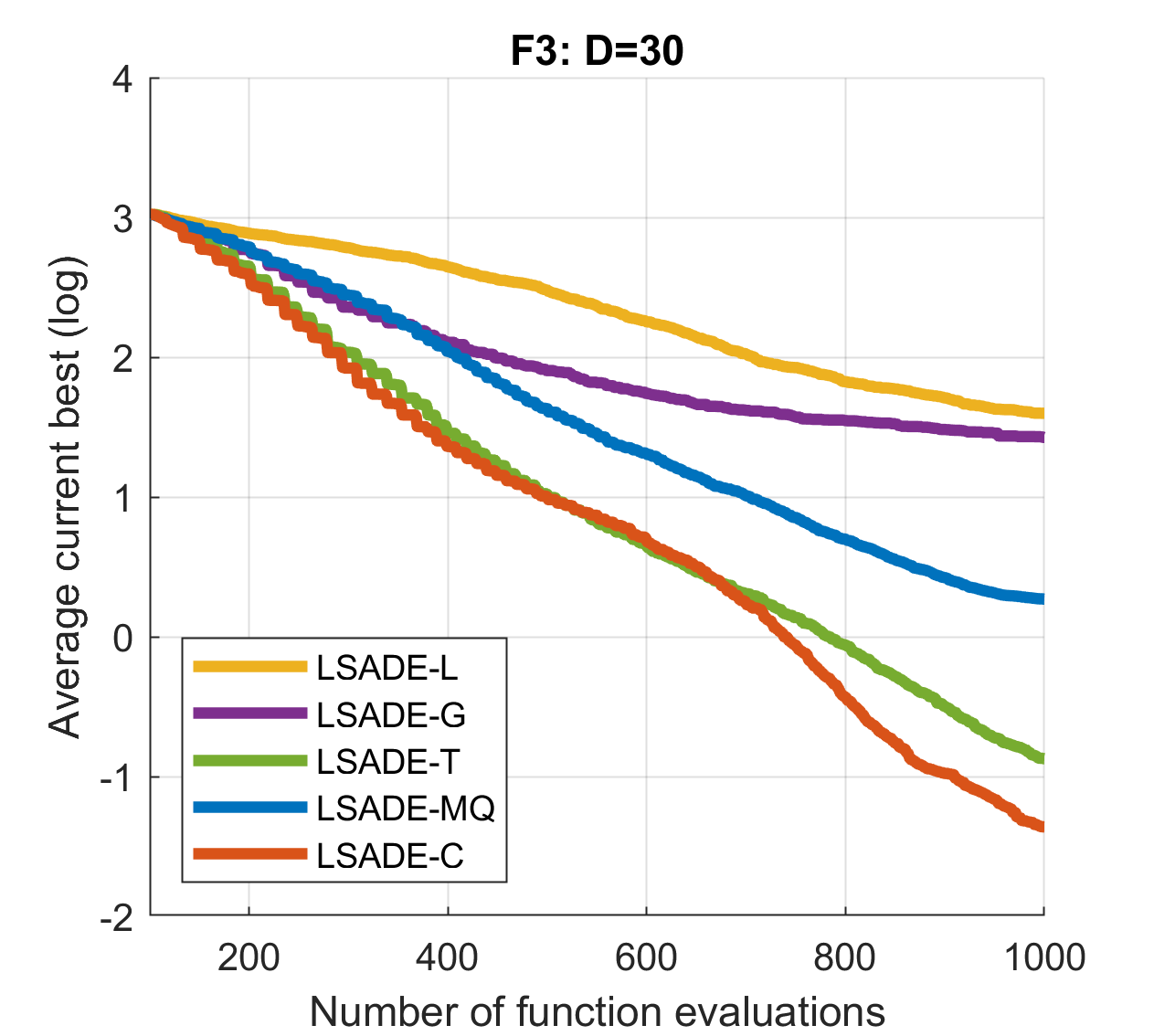} & \includegraphics[height = 0.2\textwidth]{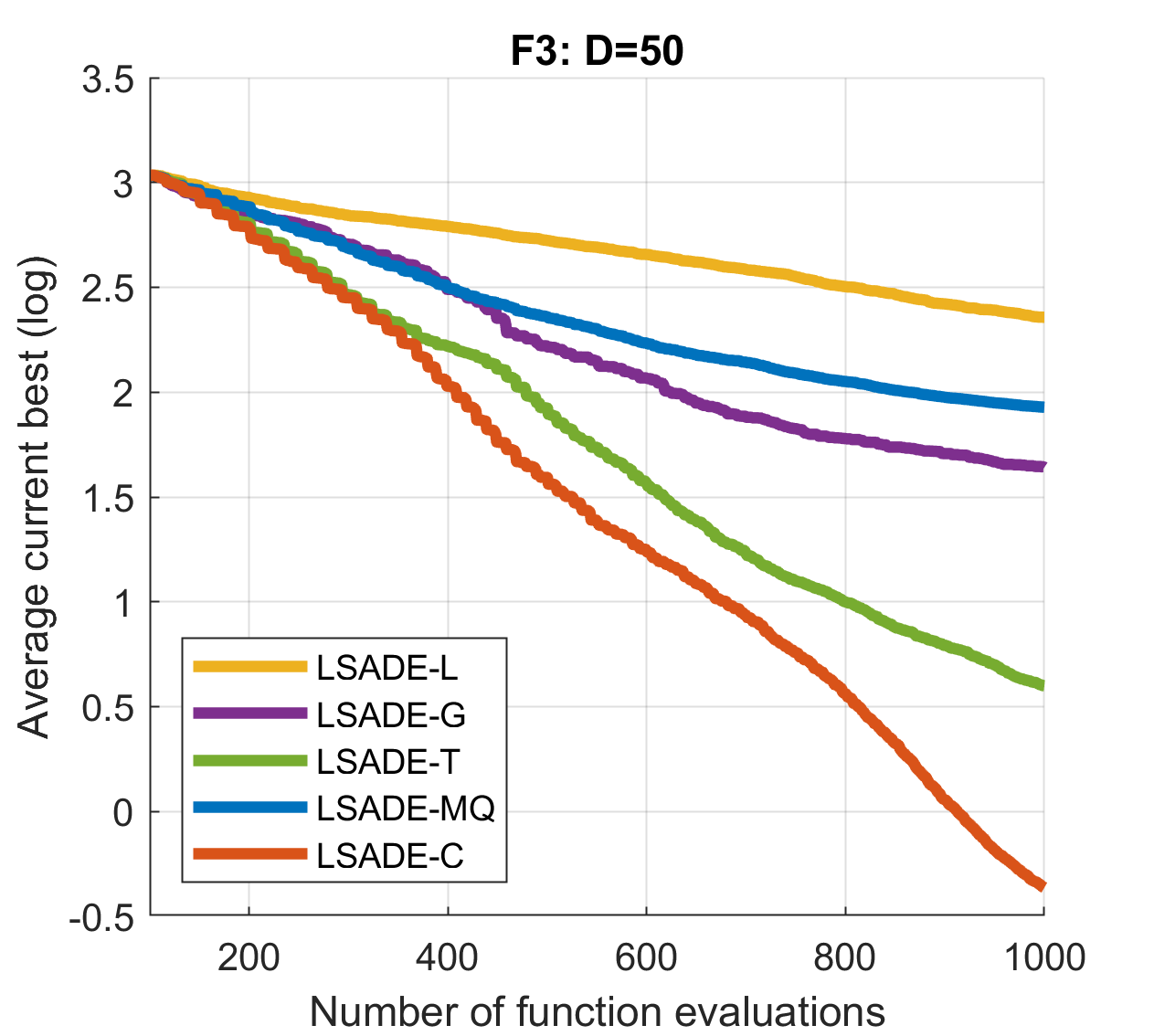} & \includegraphics[height = 0.2\textwidth]{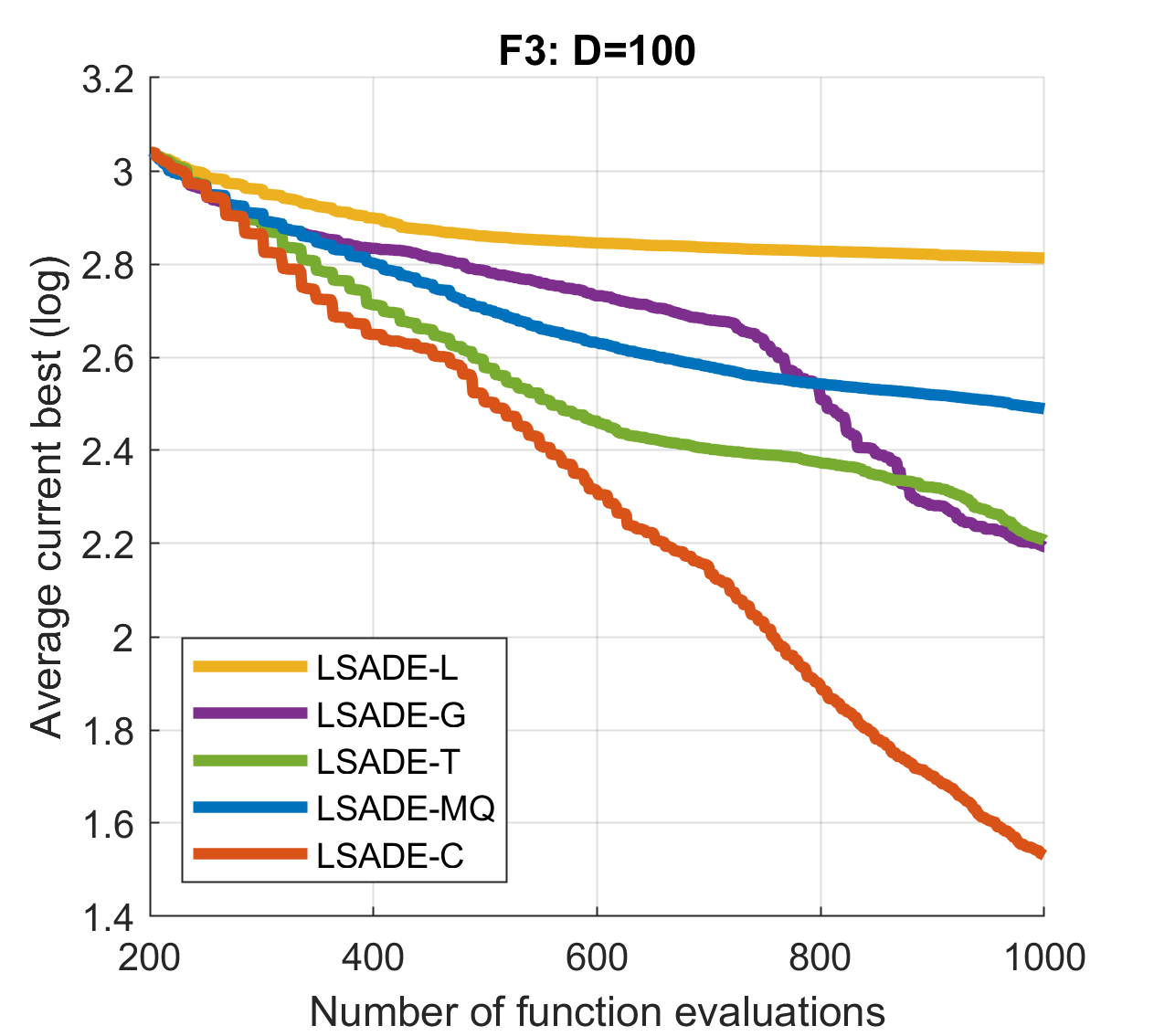}& \includegraphics[height = 0.2\textwidth]{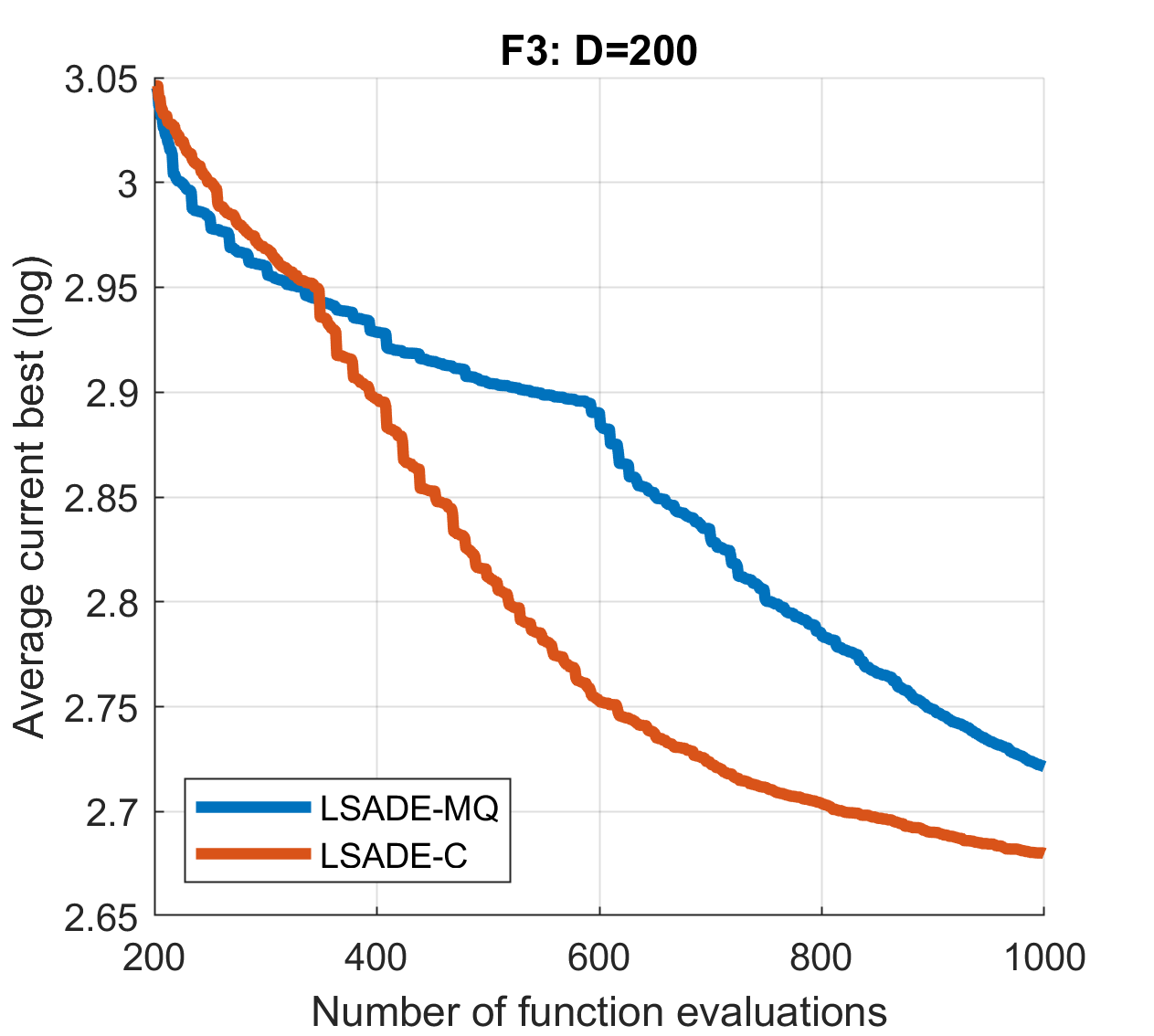} \\
\includegraphics[height = 0.2\textwidth]{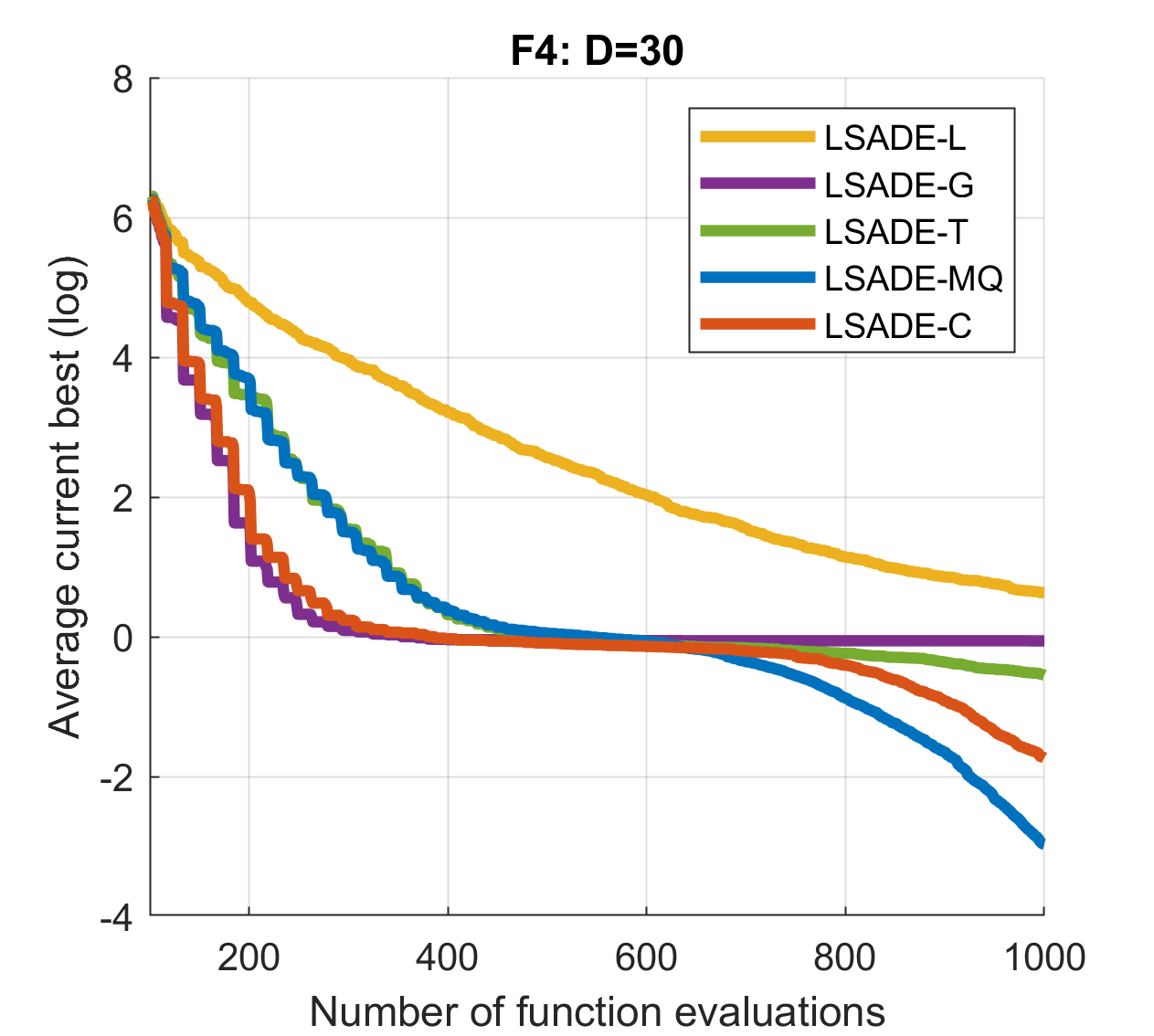} & \includegraphics[height = 0.2\textwidth]{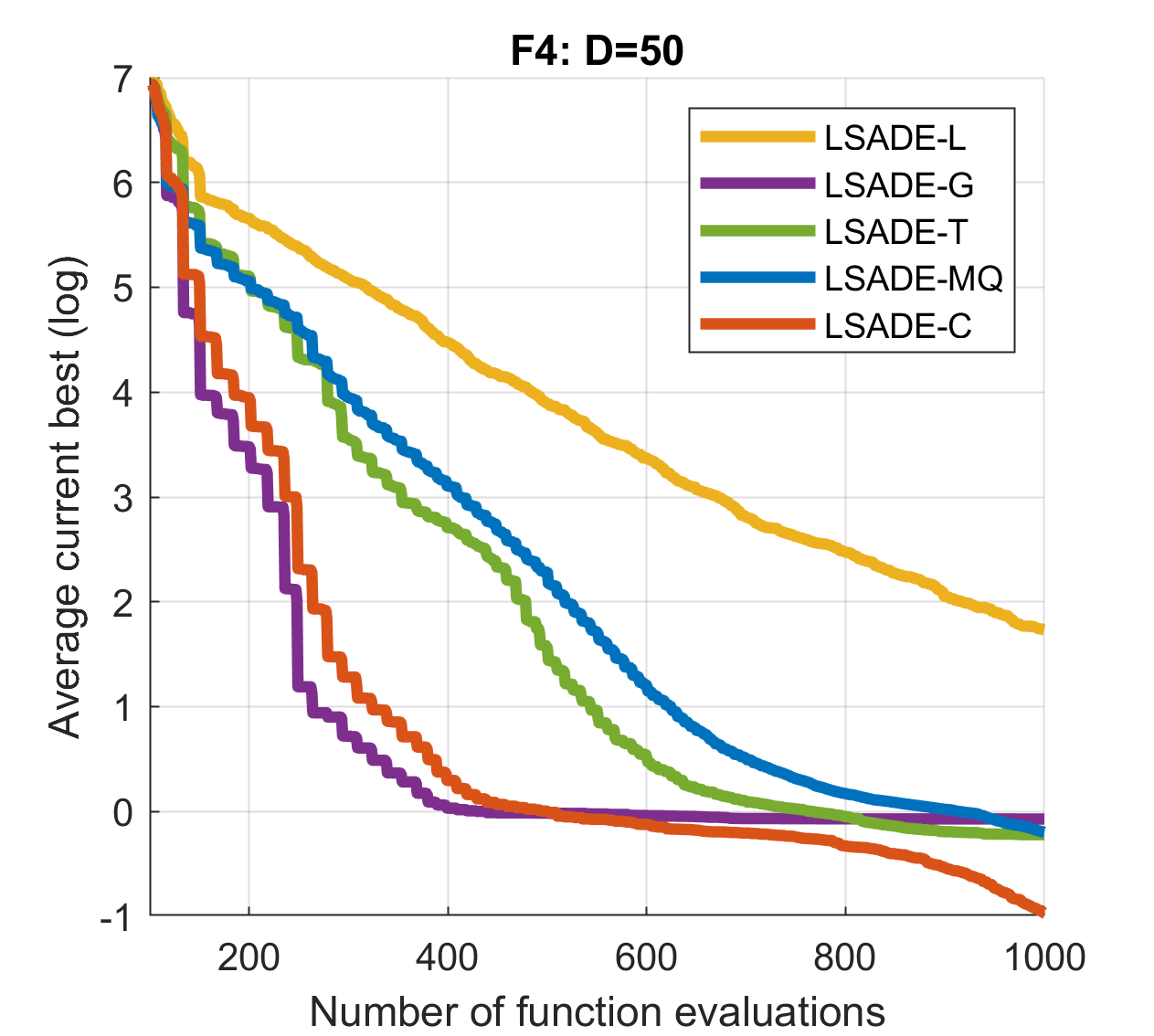} & \includegraphics[height = 0.2\textwidth]{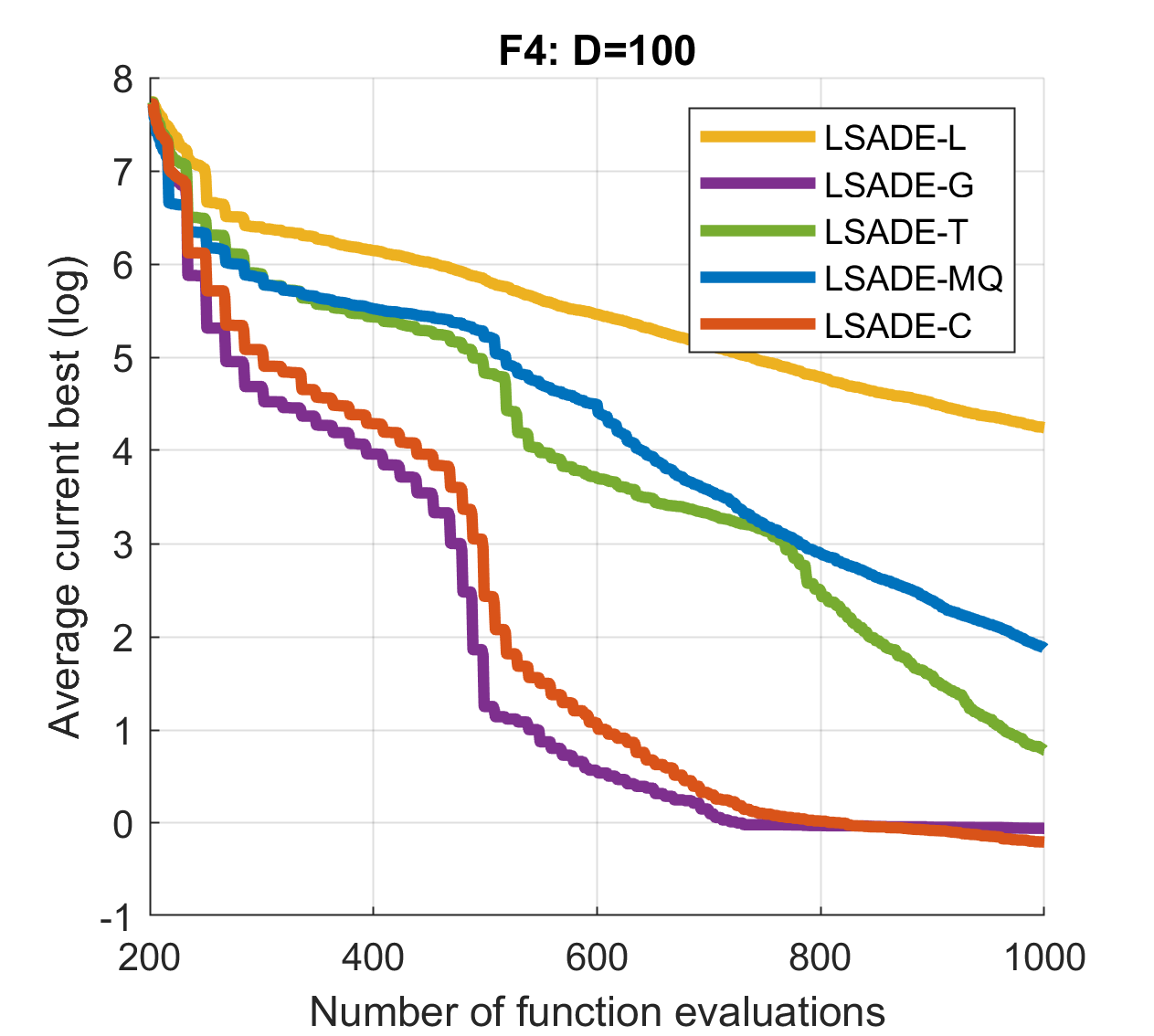} & \includegraphics[height = 0.2\textwidth]{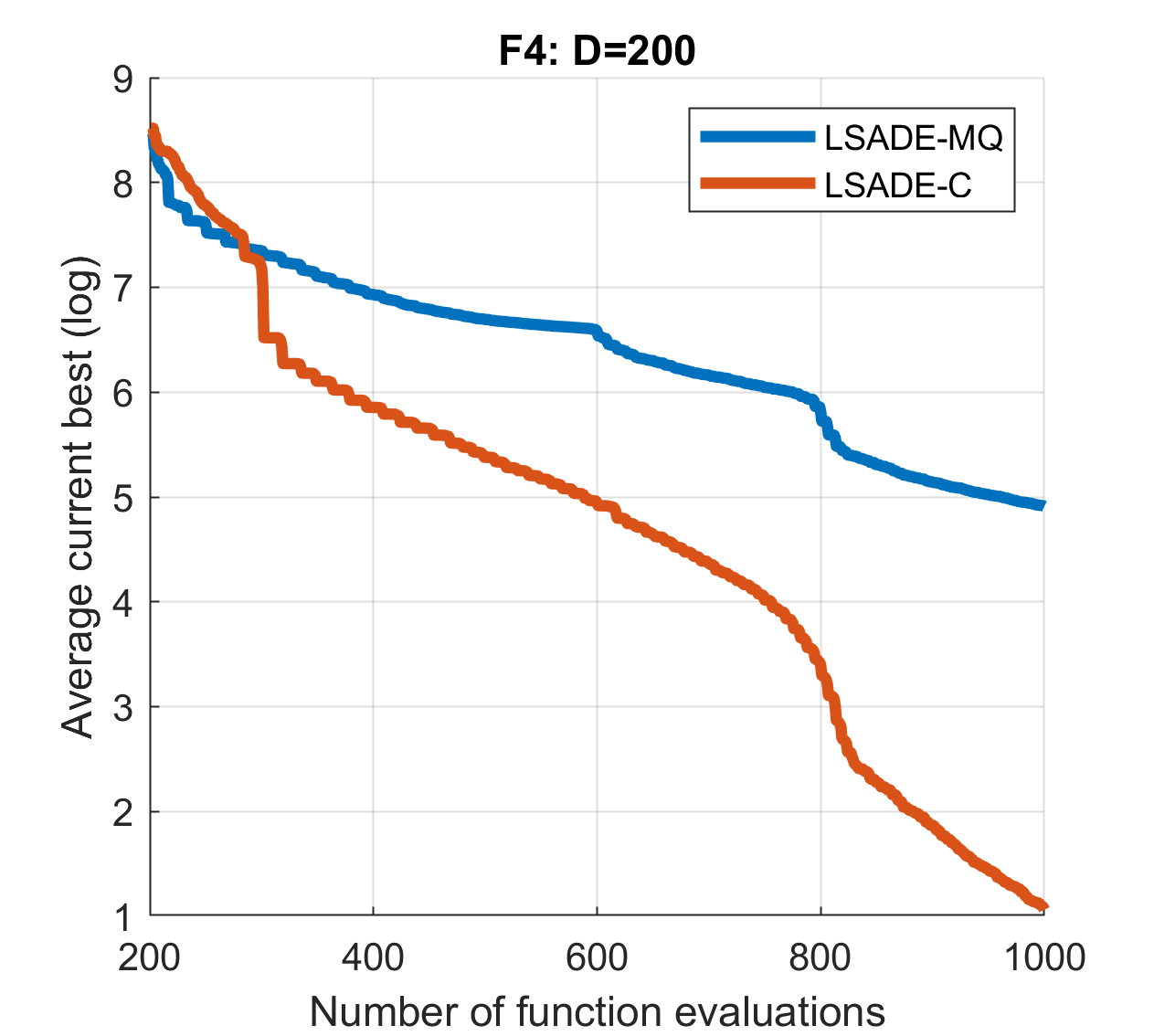}\\
\includegraphics[height = 0.2\textwidth]{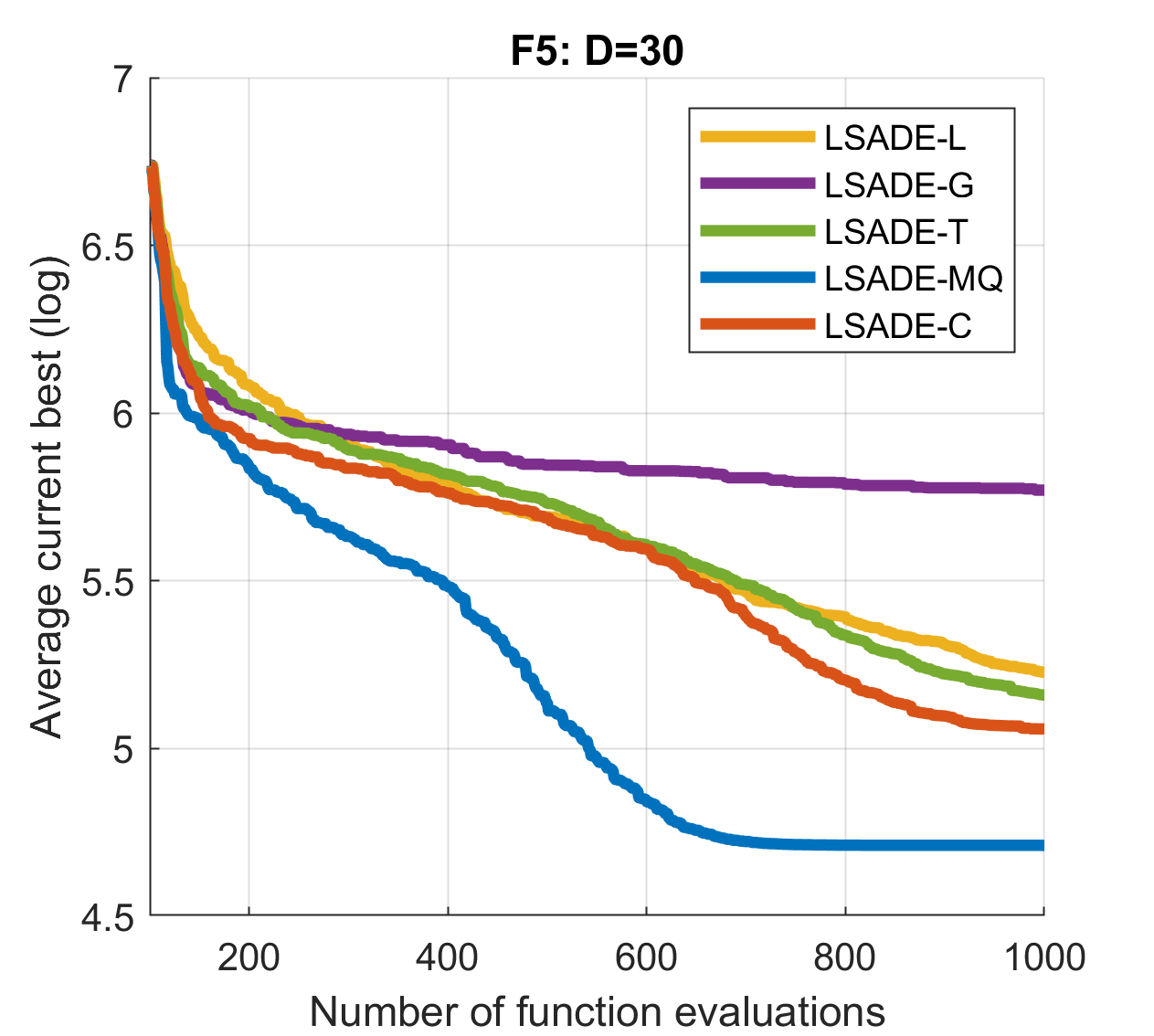} & \includegraphics[height = 0.2\textwidth]{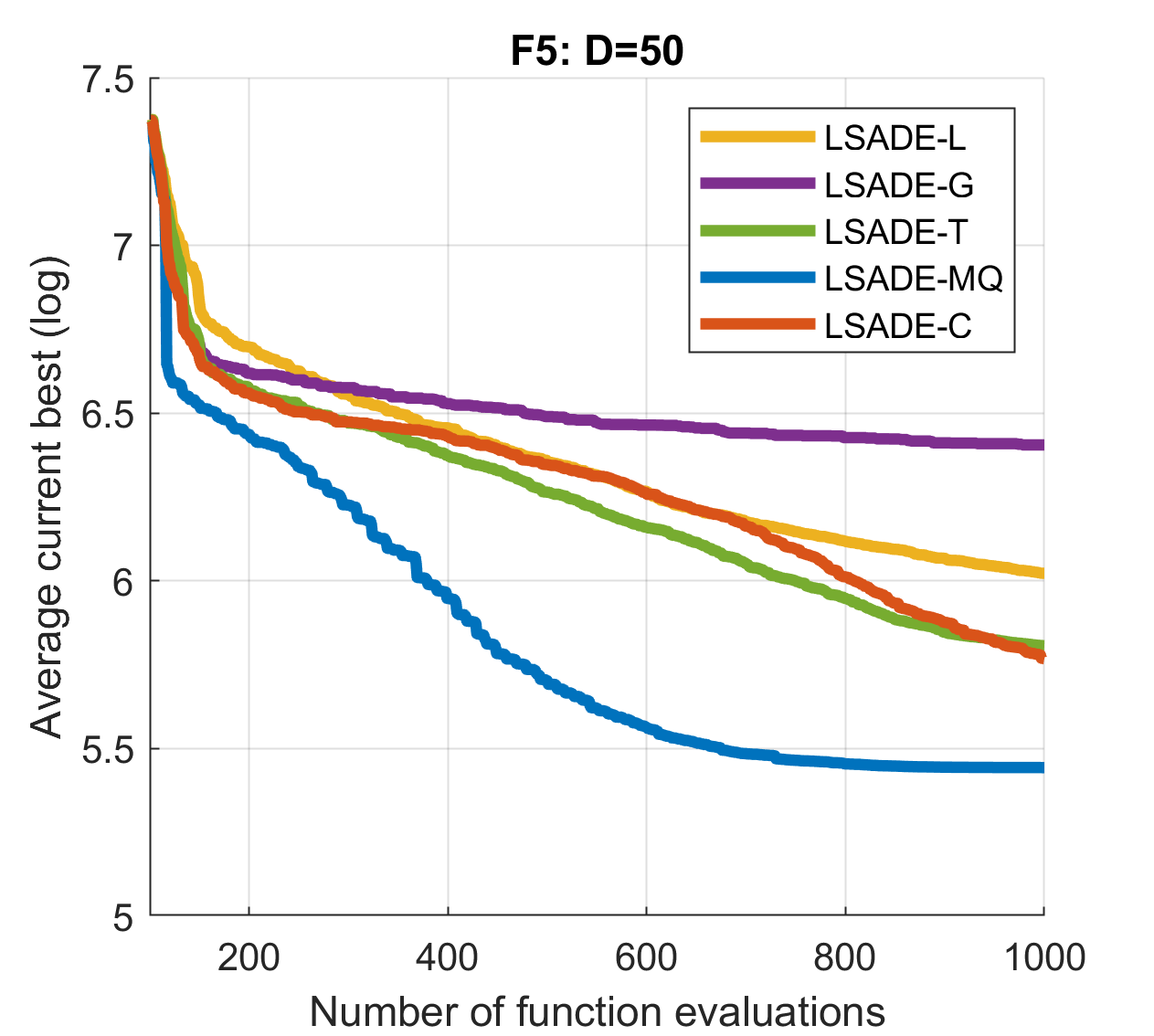} & \includegraphics[height = 0.2\textwidth]{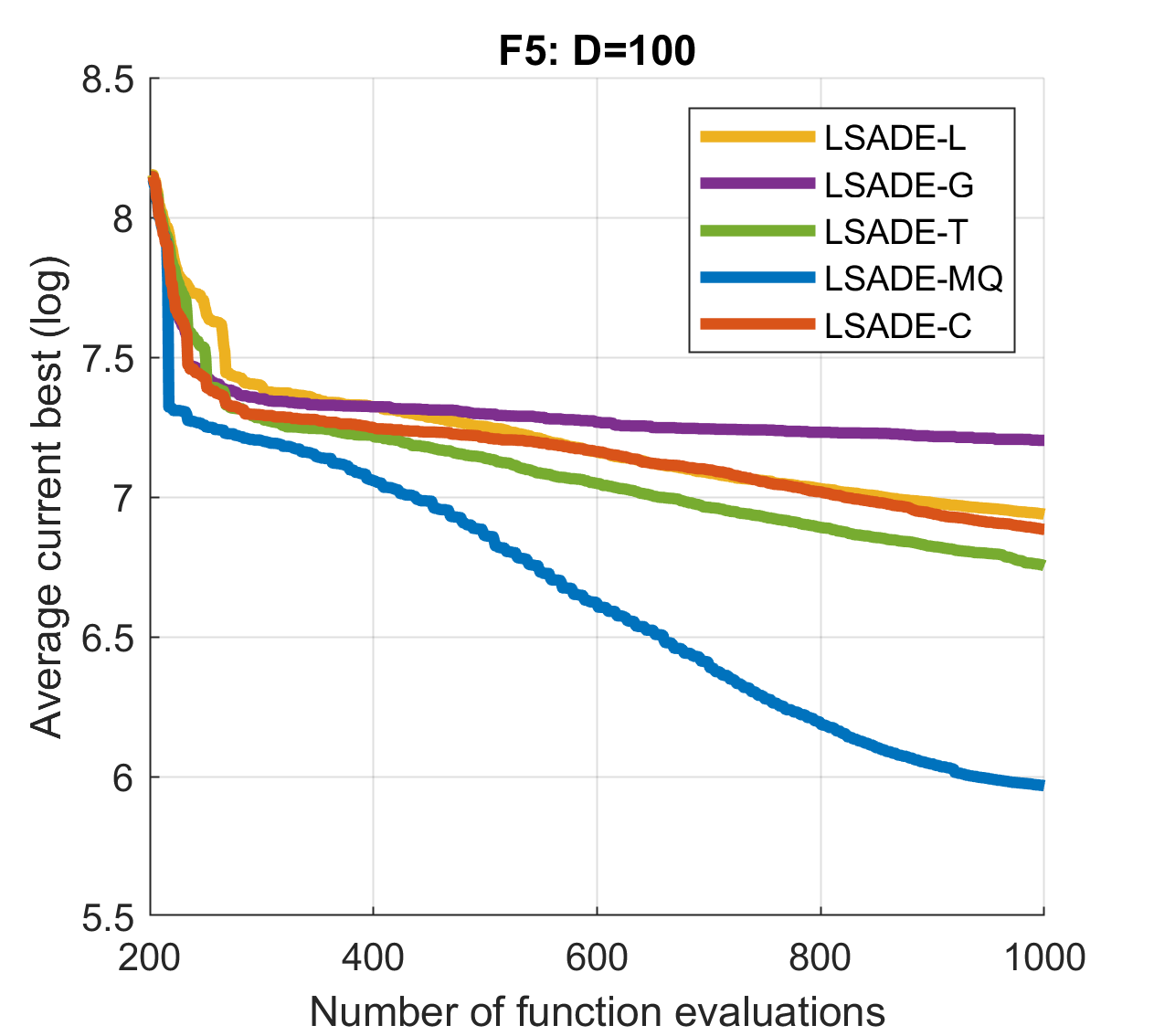} & \includegraphics[height = 0.2\textwidth]{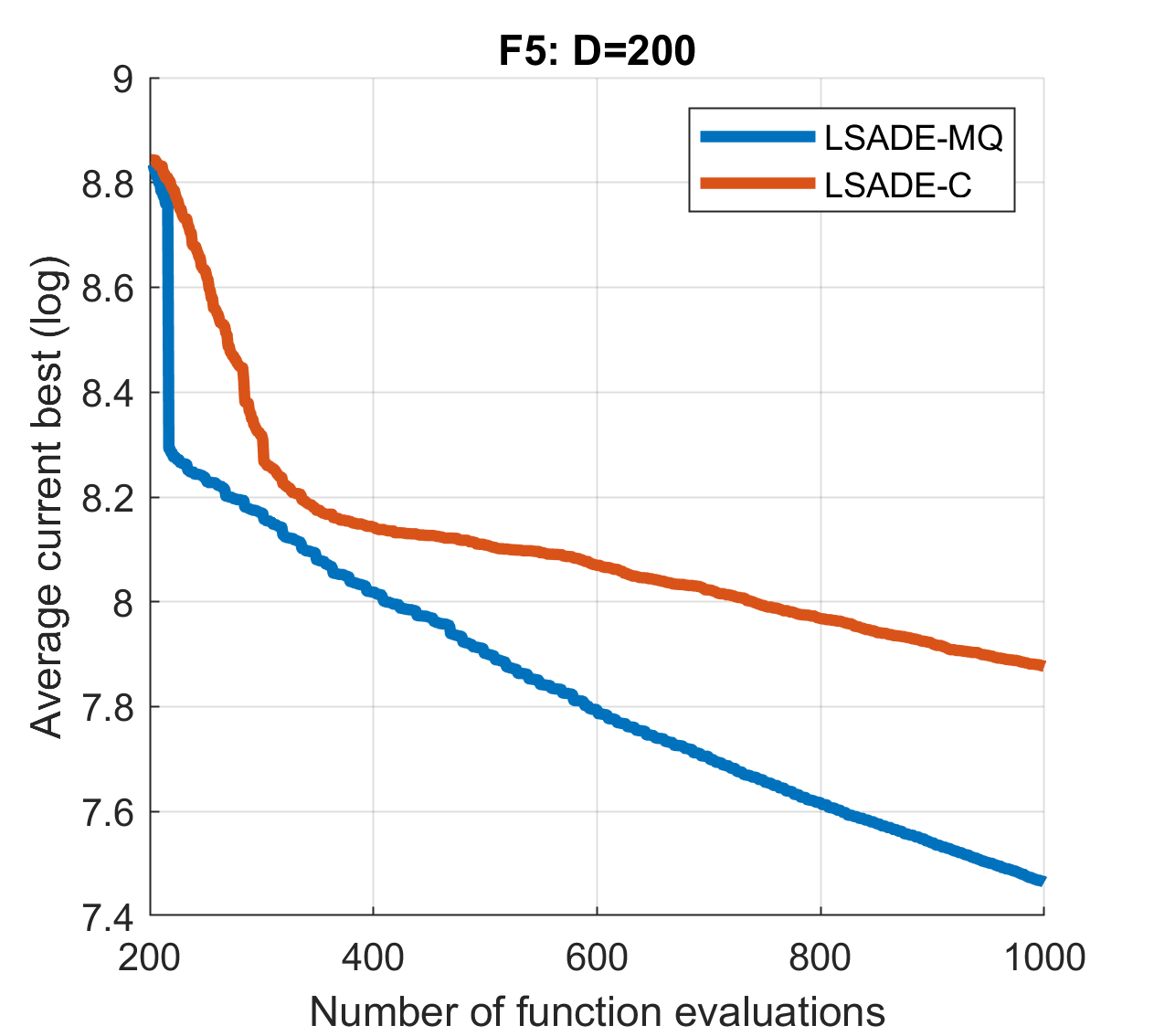}\\
\includegraphics[height = 0.2\textwidth]{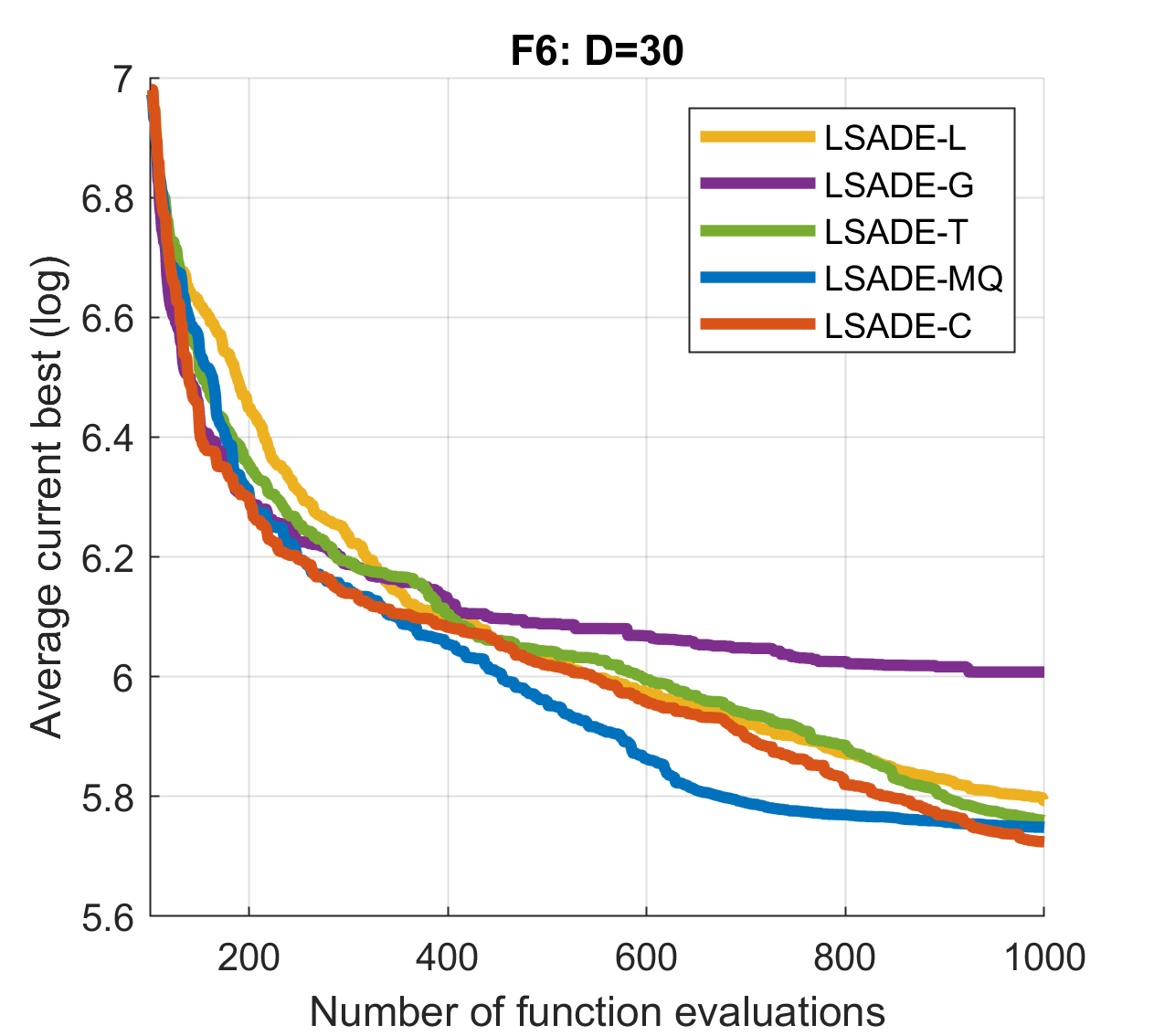} & \includegraphics[height = 0.2\textwidth]{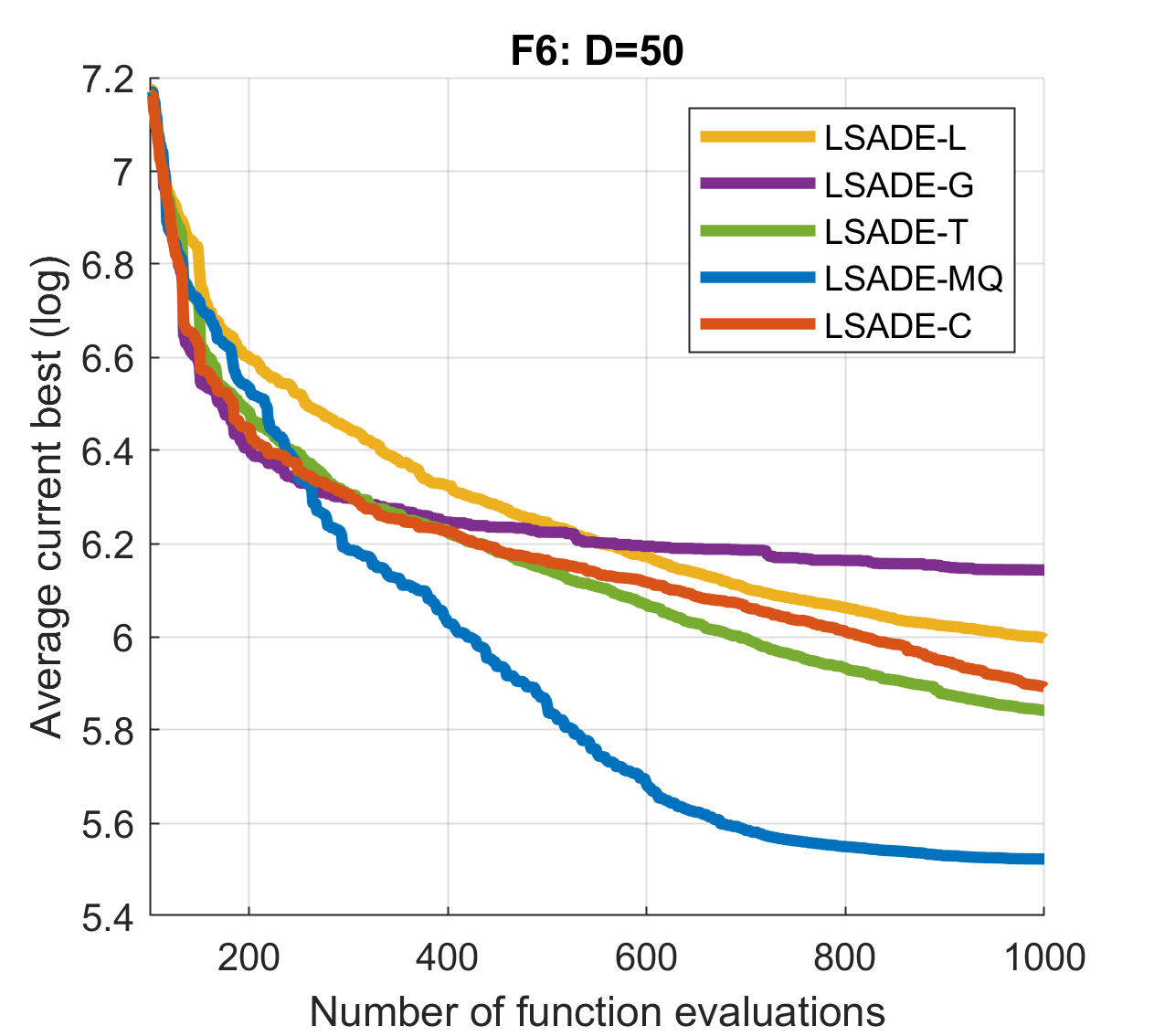} & \includegraphics[height = 0.2\textwidth]{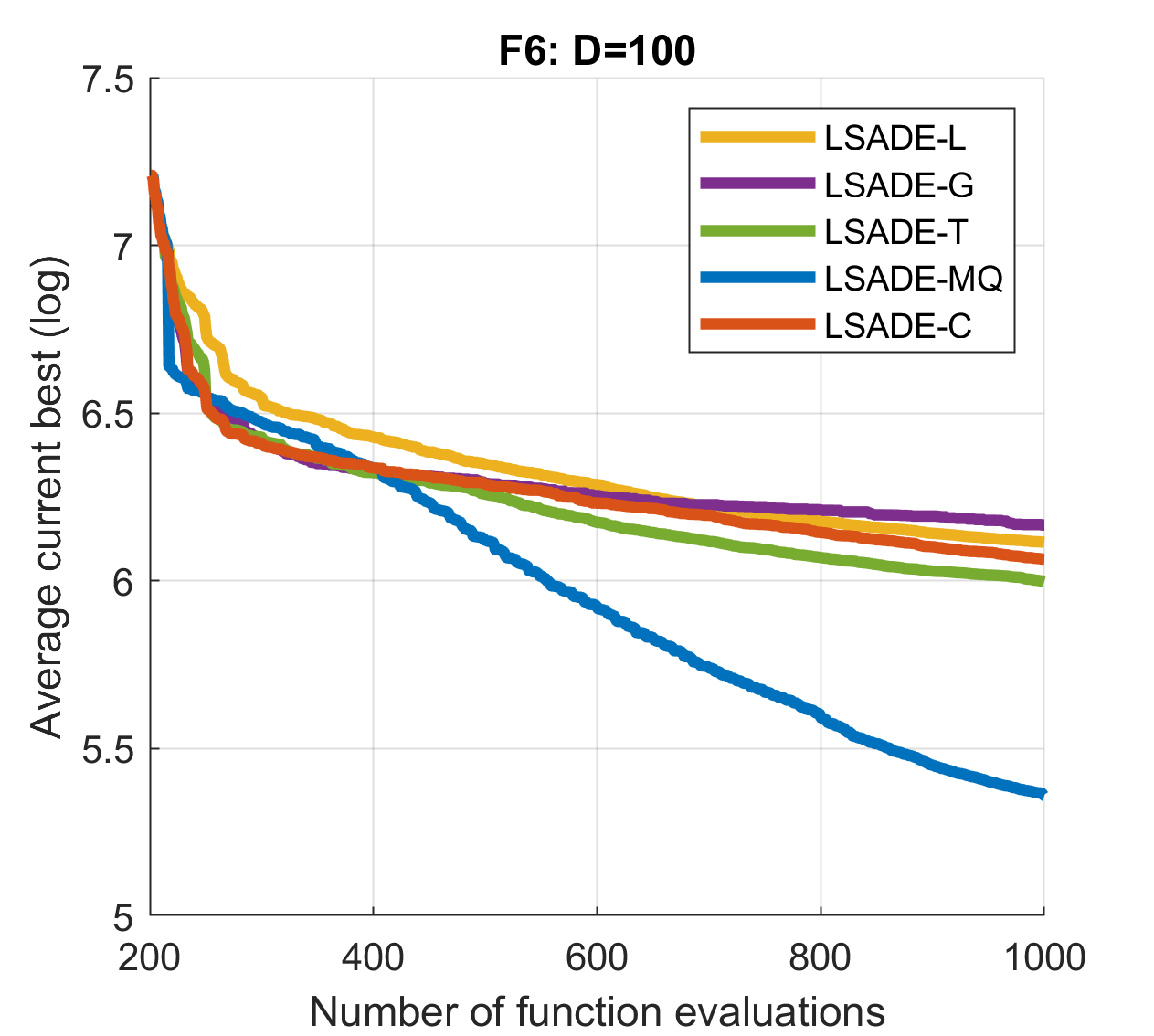} & \includegraphics[height = 0.2\textwidth]{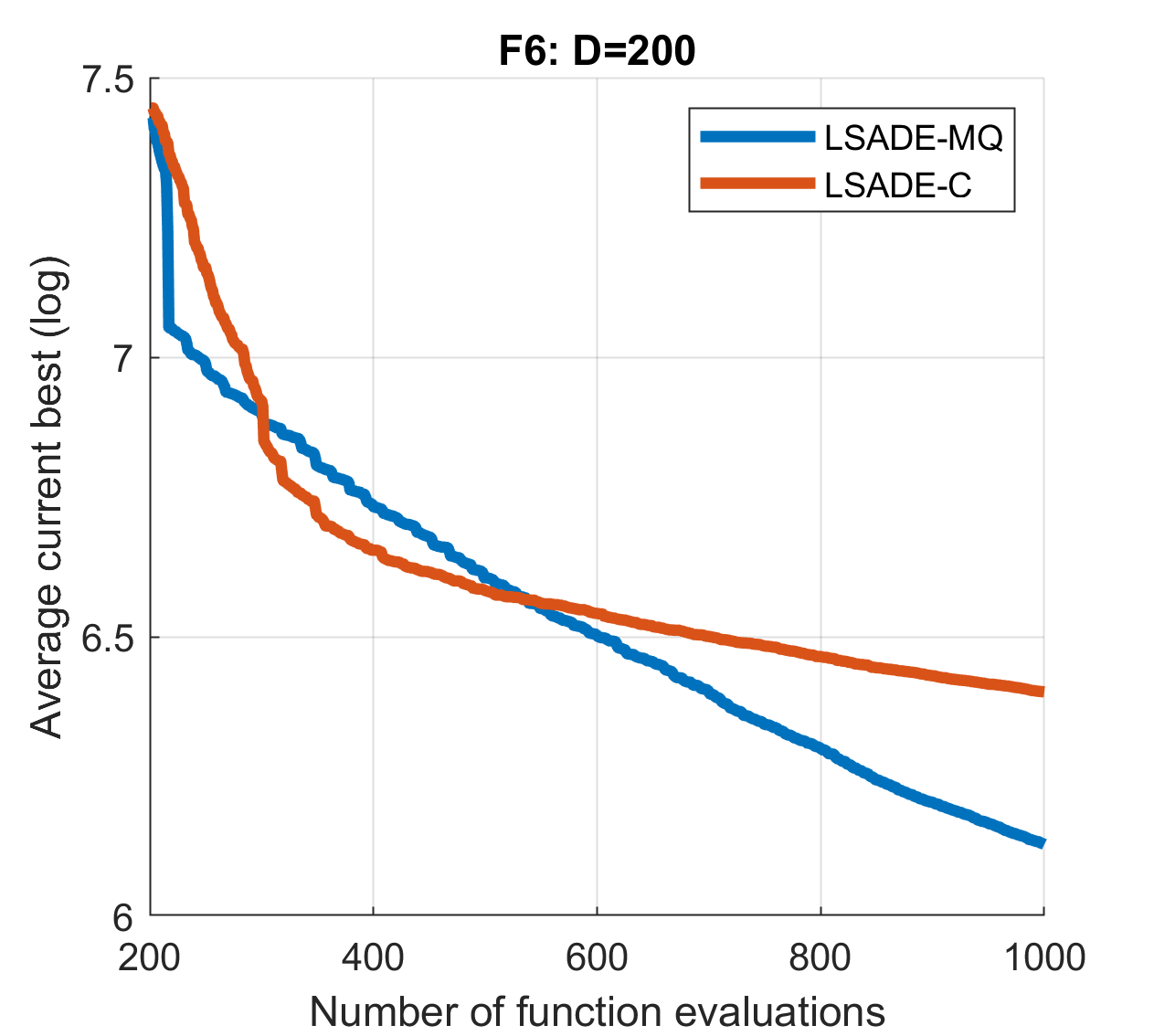}\\
\includegraphics[height = 0.2\textwidth]{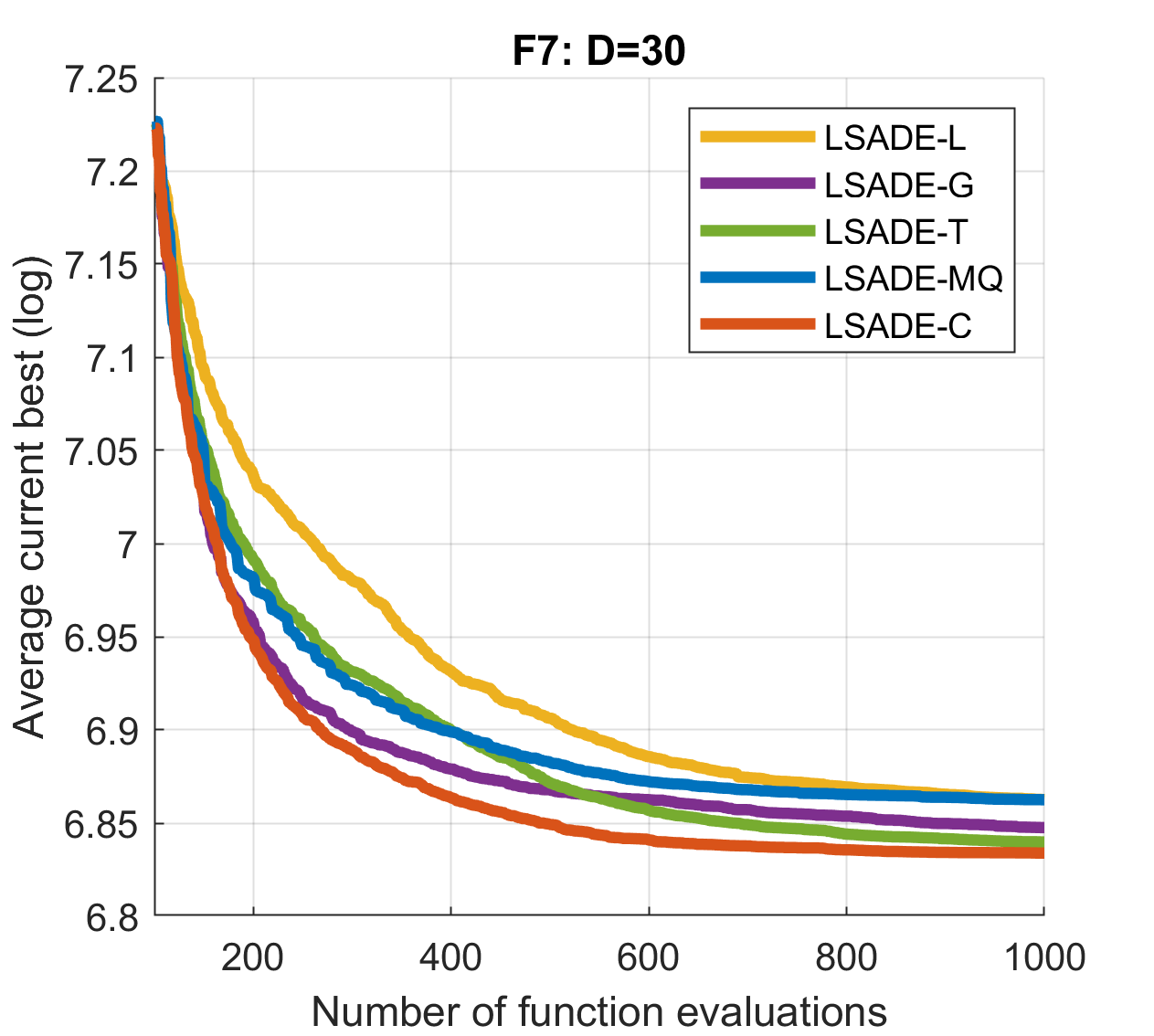} & \includegraphics[height = 0.2\textwidth]{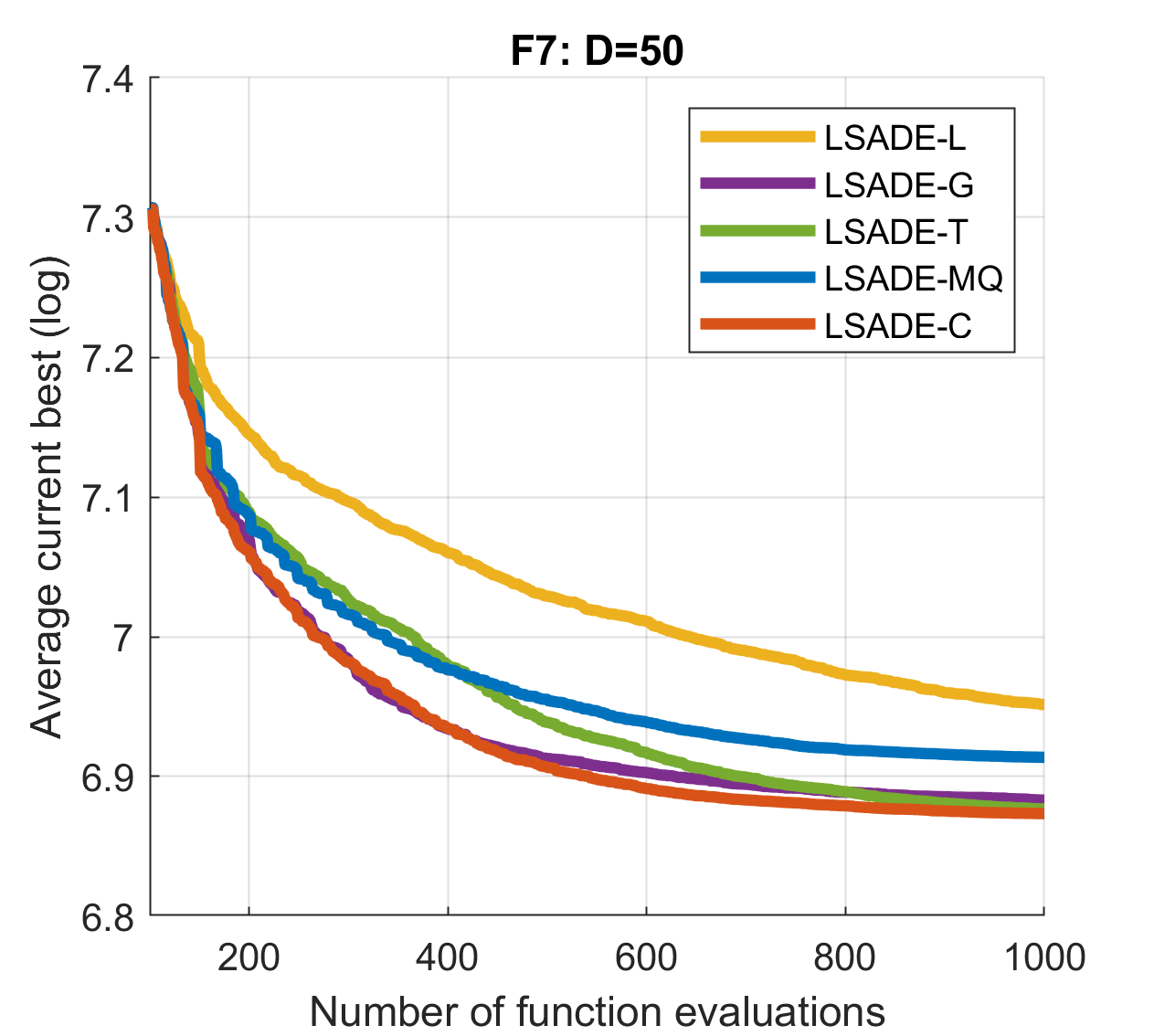} & \includegraphics[height = 0.2\textwidth]{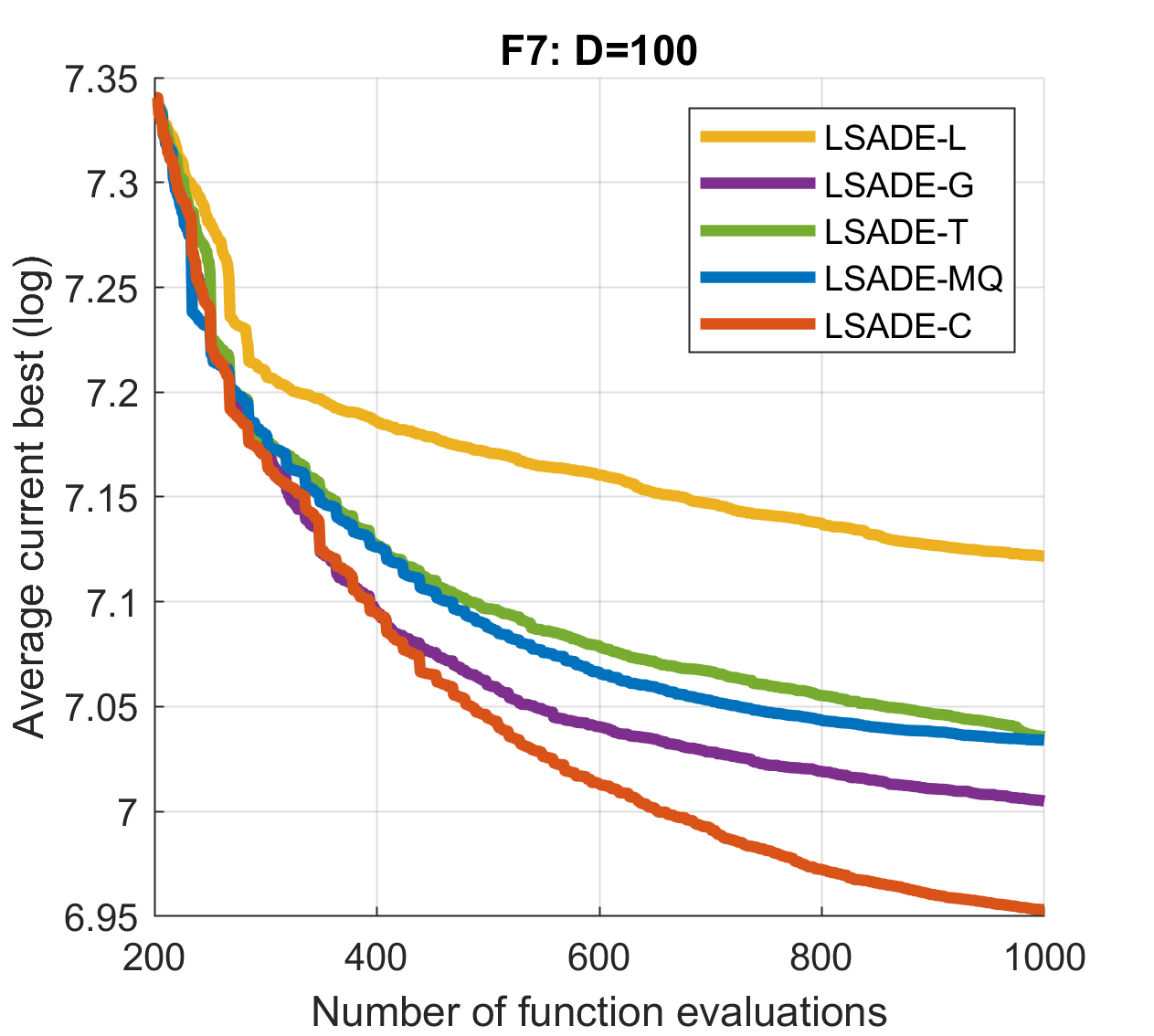}& \includegraphics[height = 0.2\textwidth]{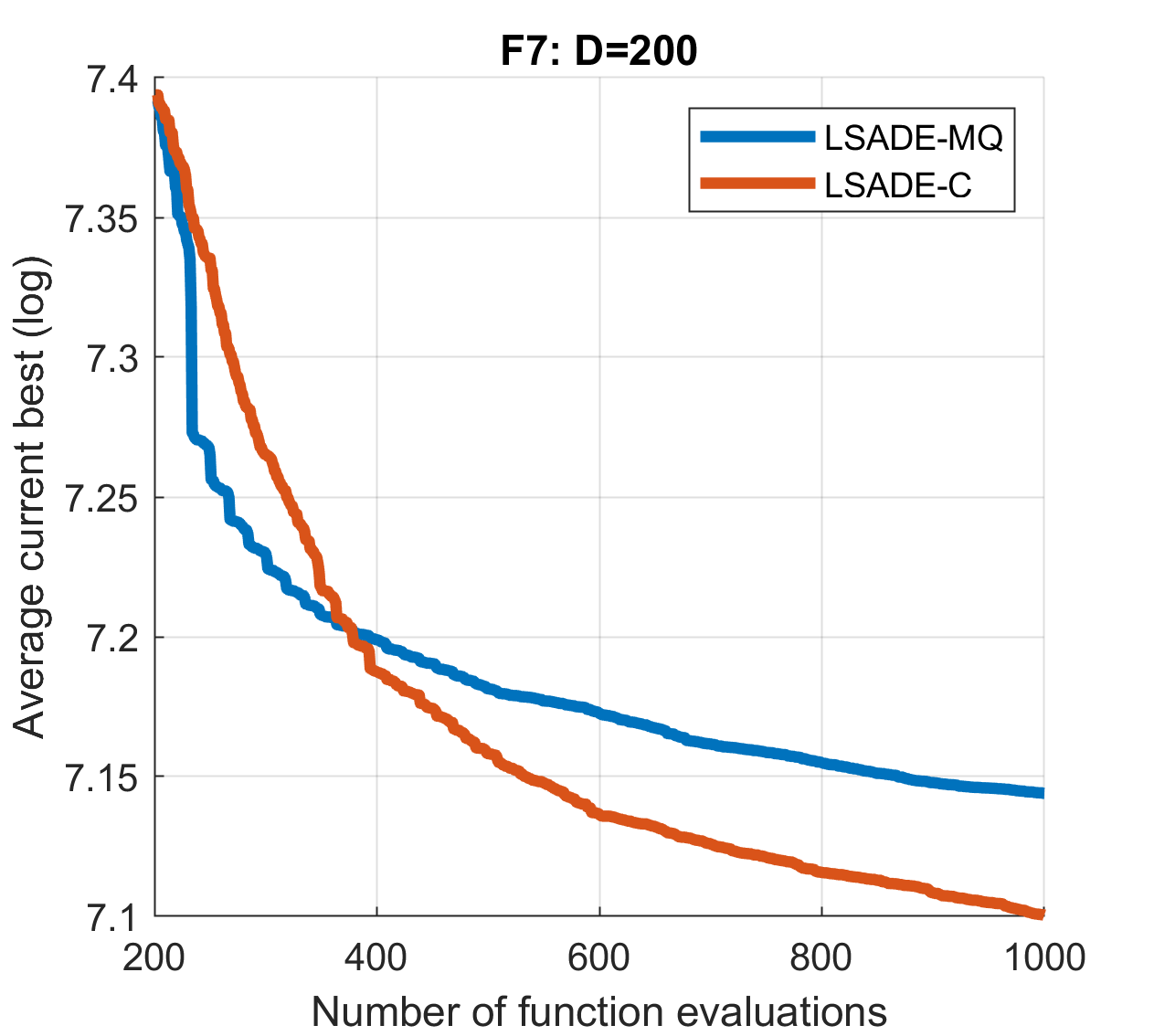} \\
\end{tabular}
\egroup
\caption{Convergence history of LSADE with different basis functions on the benchmark functions F1--F7 in different dimensions.}
 \label{fS:1a}
\end{figure*}

\begin{table}[]
\centering
\caption{Detailed statistics of the results for SAMSO, MGP-SLPSO, GSGA, and SAGWO algorithms on all considered benchmark functions.}
\bgroup
\def\arraystretch{1.2}
\setlength{\tabcolsep}{1.2pt}
\begin{tabular}{ll|rr|rrrr|rrrr|rrrr}
                        & \multicolumn{1}{l}{} & \multicolumn{2}{c}{SAMSO} & \multicolumn{4}{c}{MGP-SLPSO}       & \multicolumn{4}{c}{GSGA}                   & \multicolumn{4}{c}{SAGWO}                 \\
\multicolumn{1}{r}{$D$}   & F                    & mean        & std         & best   & mean    & worst   & std    & best      & mean     & worst    & std      & best     & mean       & worst  & std      \\ \hline \hline 
\multicolumn{1}{r}{}    & F1                   & 0.0053      & 0.0057      & N/A    & 0       & N/A     & 0      & 0.0051 & 0.072 & 0.326  & 0.093 & 0.00001  & 0.000065 & 0.0003 & 0.000075 \\
                        & F2                   & 28.3        & 0.854       & N/A    & 100     & N/A     & 22.3   & 25.69    & 27.59   & 29.04   & 1.295   & 26.79    & 28.29     & 28.83  & 0.517    \\
                        & F3                   & 0.628       & 0.542       & N/A    & 6.58    & N/A     & 2.6    & 0.0065 & 0.023 & 0.057 & 0.023 & 0        & 0          & 0      & 0        \\
\multicolumn{1}{r}{30}  & F4                   & 0.538       & 0.144       & N/A    & 0.013   & N/A     & 0.005  & 0.095  & 0.228    & 0.383  & 0.222   & 0.000001 & 0.015      & 0.134 & 0.032   \\
\multicolumn{1}{r}{}    & F5                   & -239        & 24.3        & N/A    & -220    & N/A     & 19.6   & -245.2    & -203     & -159.0  & 24.87   & -176     & -128.8     & -58.71 & 30.82    \\
                        & F6                   & 372         & 14.7        & N/A    & N/A     & N/A     & N/A    & 275.5    & 424.7    & 563.1   & 106.2   & 348.4    & 489.8      & 675.8  & 128.8    \\
                        & F7                   & 922         & 3.66        & N/A    & 952     & N/A     & 19     & 918.8    & 927.2    & 938.8   & 6.043   & 942.5    & 973.2      & 1016   & 18.47    \\ \hline \hline 
                        & F1                   & 0.513       & 0.285       & 0      & 0       & 0       & 0      & 0.203   & 0.621  & 1.868   & 0.484  & 0.0007   & 0.004      & 0.015  & 0.0036   \\
\multicolumn{1}{r}{}    & F2                   & 50.1        & 0.768       & 88.4   & 120     & 165     & 18.7   & 46.84    & 48.21   & 49.10   & 0.766   & 48.35    & 49.06      & 49.94  & 0.449    \\
                        & F3                   & 1.53        & 0.436       & 7.77   & 9.31    & 12.1    & 1.13   & 0.0060 & 0.021 & 0.076 & 0.023 & 0        & 0          & 0      & 0        \\
\multicolumn{1}{r}{50}  & F4                   & 0.666       & 0.107       & 0.037 & 0.154   & 0.614   & 0.13   & 0.272   & 0.346    & 0.442  & 0.071  & 0.000035 & 0.025      & 0.230  & 0.058    \\
                        & F5                   & -169        & 31.7        & -43.4  & 33      & 88.4    & 36.1   & -139.6   & -75.82   & 12.09   & 49.99   & -16.63   & 98.39      & 161.5  & 46.90    \\
\multicolumn{1}{r}{}    & F6                   & 326         & 98.6        &    N/A    & N/A     &     N/A    &  N/A      & 271.8    & 403.3    & 524.8   & 87.59  & 430.2    & 502        & 564.2  & 45.25    \\ 
                        & F7                   & 970         & 29.2        & 1030   & 1060    & 1110    & 21.4   & 943.7    & 970.7    & 1002   & 18.18   & 910      & 1044       & 1132   & 40.83    \\ \hline \hline 
                        & F1                   & 72.1        & 17.8        & 0      & 0.00005 & 0.001 & 0.0002 & 2.603     & 12.32   & 27.15   & 9.394   & 0.017    & 0.139      & 0.371  & 0.097    \\
                        & F2                   & 286         & 52.5        & 455    & 612     & 733     & 67.9   & 99.743    & 109.0   & 139.3   & 11.76   & 104.9    & 123.4      & 144.8  & 11.02    \\
\multicolumn{1}{r}{}    & F3                   & 6.12        & 0.409       & 13.4   & 14.3    & 15.7    & 0.621  & 0.156   & 1.31     & 2.807   & 0.968  & 0        & 0          & 0      & 0        \\
\multicolumn{1}{r}{100} & F4                   & 1.06        & 0.026      & 0.478  & 0.715   & 0.847   & 0.724  & 0.580   & 0.706    & 0.804  & 0.070 & 0.00021  & 0.023      & 0.229  & 0.052    \\
                        & F5                   & 737         & 42          & 877    & 885     & 1160    & 117    & 620.4     & 672.5    & 705.2   & 29.79   & 676.7    & 800.1      & 919.0  & 79.27    \\
                        & F6                   & 513         & 18.5        & N/A    & N/A     & N/A     & N/A    & 422.4    & 447.2    & 472.6   & 14.25   & 482.0    & 518.6      & 555.3  & 20.54    \\
\multicolumn{1}{r}{}    & F7                   & 1290        & 33.4        & 1330   & 1390    & 1490    & 47.7   & 1220    & 1256     & 1287   & 24.56   & 910.2    & 1350       & 1437   & 107.5    \\ \hline \hline 
                        & F1                   & 1520        & 21.2        & N/A    & N/A     & N/A     & N/A    & N/A       & N/A      & N/A      & N/A      & N/A      & N/A        & N/A    & N/A      \\
                        & F2                   & 1150        & 11.6        & N/A    & N/A     & N/A     & N/A    & N/A       & N/A      & N/A      & N/A      & N/A      & N/A        & N/A    & N/A      \\
                        & F3                   & 12          & 0.4         & N/A    & N/A     & N/A     & N/A    & N/A       & N/A      & N/A      & N/A      & N/A      & N/A        & N/A    & N/A      \\
\multicolumn{1}{r}{200} & F4                   & 9.03        & 1.33        & N/A    & N/A     & N/A     & N/A    & N/A       & N/A      & N/A      & N/A      & N/A      & N/A        & N/A    & N/A      \\
                        & F5                   & 4960        & 138         & N/A    & N/A     & N/A     & N/A    & N/A       & N/A      & N/A      & N/A      & N/A      & N/A        & N/A    & N/A      \\
                        & F6                   & 684         & 37.2        & N/A    & N/A     & N/A     & N/A    & N/A       & N/A      & N/A      & N/A      & N/A      & N/A        & N/A    & N/A      \\
                        & F7                   & 1340        & 24.3        & N/A    & N/A     & N/A     & N/A    & N/A       & N/A      & N/A      & N/A      & N/A      & N/A        & N/A    & N/A     \\ \hline \hline 
\end{tabular}
\egroup
\label{tbS:6}
\end{table}

\begin{table*}[]
\centering
\caption{Detailed statistics of the results for EASO, SA-COSO, LSADE-MQ, and LSADE-C algorithms on all considered benchmark functions.}
\bgroup
\def\arraystretch{1.2}
\setlength{\tabcolsep}{4.2pt}
\begin{tabular}{ll|rr|rr|rrrr|rrrr}
                        & \multicolumn{1}{l}{} & \multicolumn{2}{|c}{EASO} & \multicolumn{2}{|c}{SA-COSO} & \multicolumn{4}{|c}{LSADE-MQ}         & \multicolumn{4}{|c}{LSADE-C}           \\
\multicolumn{1}{r}{$D$}   & F                    & mean        & std        & mean         & std          & best   & mean     & worst    & std     & best       & mean     & worst    & std      \\ \hline \hline
\multicolumn{1}{r}{}    & F1                   & 0.027       & 0.070      & 3.85         & 1.19         & 0.0039 & 0.011   & 0.021    & 0.005   & 0.0008    & 0.011   & 0.047    & 0.012   \\
                        & F2                   & 25.04       & 1.57       & 59.9         & 24.3         & 24.31  & 27.06    & 29.35   & 1.243   & 27.20   & 27.77    & 29.36   & 0.546   \\
                        & F3                   & 2.521       & 0.84       & 5.01         & 1.22         & 0.026  & 1.308    & 3.028    & 1.011   & 0.0025    & 0.256    & 1.186    & 0.441   \\
\multicolumn{1}{r}{30}  & F4                   & 0.953       & 0.05       & 1.44         & 0.18         & 0.0098 & 0.051    & 0.107    & 0.027   & 0.046    & 0.176    & 0.673    & 0.172   \\
\multicolumn{1}{r}{}    & F5                   & 6.325       & 26.5       & -57.4        & 17.5         & -278.2 & -218.7   & -136.0 & 35.68  & -256.2 & -172.6   & -81.31  & 39.83  \\
                        & F6                   & N/A         & N/A        & 528          & 94.8         & 229.2  & 433.7    & 664.1  & 149.3 & 233.5  & 426.2    & 674.3  & 148.1 \\
                        & F7                   & 931.6       & 8.94       & 969          & 24.3         & 922.2  & 965.7    & 1097 & 51.86  & 916.1  & 938.8    & 1004.3 & 26.37  \\ \hline \hline
                        & F1                   & 0.740       & 0.555      & 46.6         & 17.4         & 0.265  & 1.358   & 3.500    & 0.860   & 0.047      & 0.433   & 1.304    & 0.299    \\
\multicolumn{1}{r}{}    & F2                   & 47.39       & 1.71       & 253          & 56.7         & 43.92  & 47.65    & 49.17   & 1.332   & 45.53     & 47.98    & 49.19   & 0.864    \\
                        & F3                   & 1.431       & 0.249      & 8.86         & 1.1          & 2.615  & 6.876    & 15.39   & 3.456   & 0.029      & 0.695    & 2.264    & 0.600    \\
\multicolumn{1}{r}{50}  & F4                   & 0.94        & 0.042      & 5.63         & 0.892        & 0.560  & 0.819    & 1.051    & 0.132   & 0.198      & 0.38     & 0.662    & 0.129    \\
                        & F5                   & 198.6       & 45.8       & 235          & 40.9         & -194.6 & -98.78   & -5.288   & 52.92  & -183.0   & -10.03   & 151.2  & 93.88   \\
\multicolumn{1}{r}{}    & F6                   & N/A         & N/A        & 613          & 37.4         & 259.0  & 370.3    & 579.8  & 109.5 & 339.2    & 481.6    & 657.1  & 80.89   \\
                        & F7                   & 975.3       & 37.1       & 1080         & 36.6         & 954.3  & 1016     & 1134 & 53.369  & 936.1    & 976.3    & 1103 & 38.52   \\ \hline \hline
                        & F1                   & 1283        & 134        & 985          & 214          & 58.02  & 112.8 & 171.2  & 33.61  & 12.93     & 30.94  & 61.73   & 12.46   \\
                        & F2                   & 578.8       & 44.8       & 2500         & 97.4         & 108.3  & 140.6    & 194.3  & 24.10  & 97.62     & 106.4    & 120.8  & 6.631    \\
\multicolumn{1}{r}{}    & F3                   & 10.36       & 0.211      & 15.9         & 0.514        & 9.431  & 12.05    & 16.54   & 2.203   & 3.540      & 4.622    & 6.157    & 0.619    \\
\multicolumn{1}{r}{100} & F4                   & 57.34       & 5.84       & 63.5         & 14.9         & 3.344  & 6.517    & 10.04   & 1.974   & 0.694      & 0.816    & 0.923    & 0.059    \\
                        & F5                   & 713.4       & 26.5       & 1420         & 123          & -71.63 & 60.28    & 426.2  & 121.0 & 503.0    & 646.8    & 768.0  & 64.37   \\
                        & F6                   & N/A         & N/A        & 807          & 65.7         & 267.3  & 332.7    & 419.4  & 37.77  & 486.8    & 550.4    & 688.2  & 43.30   \\
\multicolumn{1}{r}{}    & F7                   & 1372        & 27.5       & 1410         & 22.8         & 1076   & 1144     & 1232 & 44.45  & 1002  & 1056     & 1146 & 34.27   \\ \hline \hline
                        & F1                   & 17616       & 1170       & 16382        & 2980         & 2473   & 3959   & 5192  & 705.2 & 587.6    & 793.5 & 1137 & 154.3  \\
                        & F2                   & 4318        & 284        & 16411        & 4100         & 683.0  & 927.2    & 1087  & 112.4 & 507.3    & 576.3    & 662.5  & 46.35   \\
                        & F3                   & 14.69       & 0.219      & 17.86        & 0.022        & 13.97  & 15.2     & 16.08    & 0.509   & 11.59     & 14.58    & 17.30   & 1.400    \\
\multicolumn{1}{r}{200} & F4                   & 572.9       & 36         & 577.7        & 101          & 93.89  & 135.6    & 188.2   & 22.05  & 2.149      & 2.892    & 3.456    & 0.394    \\
                        & F5                   & 5389        & 157        & 3927         & 27.3         & 1114   & 1416     & 2034  & 287.1 & 2042   & 2305     & 2625 & 156.8  \\
                        & F6                   & N/A         & N/A        & N/A          & N/A          & 474.5  & 578.7    & 883.3   & 88.76  & 541.2   & 722.7    & 818.7  & 60.65   \\
                        & F7                   & 1456        & 20.4       & 1347         & 24.7         & 1226   & 1276     & 1352  & 28.15  & 1140   & 1222     & 1274 & 33.38  \\ \hline \hline
\end{tabular}
\egroup
\label{tbS:7}
\end{table*}

\end{document}